\documentclass[journal,hidelinks]{IEEEtran}

\usepackage{silence}
\usepackage{orcidlink}
\usepackage{cite}
\usepackage{amsmath}
\usepackage{amssymb}
\usepackage{bm}
\usepackage[nolist,nohyperlinks]{acronym}
\usepackage[inline,shortlabels]{enumitem}
\usepackage{subcaption}
\usepackage{multirow}
\usepackage{booktabs}

\usepackage{siunitx}
\usepackage{tikz}
\usepackage{pgfplots}

\usepackage{soul}
\sethlcolor{yellow}
\renewcommand{\hl}[1]{#1}
\robustify\ac
\robustify\acs
\robustify\acp
\robustify\acsp
\robustify\acf
\robustify\acfp
\robustify\qtyproduct
\robustify\num

\usepackage{hyperref}

\DeclareSIUnit{\pixel}{pixel}
\DeclareSIUnit{\pixels}{pixels}
\DeclareSIUnit{\band}{bands}
\DeclareSIUnit{\sband}{spectral \ bands}
\DeclareSIUnit{\patch}{patch}
\DeclareSIUnit{\patches}{patches}
\DeclareSIUnit{\tiles}{tiles}
\DeclareSIUnit{\epoch}{epochs}
\DeclareSIUnit{\megapixel}{MP}
\DeclareSIUnit{\bit}{bit}
\DeclareSIUnit{\day}{days}
\DeclareSIUnit{\flops}{FLOPs}
\DeclareSIUnit{\nothing}{\relax}

\usetikzlibrary{arrows}
\usetikzlibrary{positioning}
\usetikzlibrary{calc}
\usetikzlibrary{shapes.geometric}
\usetikzlibrary{patterns}
\usetikzlibrary{patterns.meta}
\usetikzlibrary{spy}
\usetikzlibrary{fit}
\usetikzlibrary{trees}

\tikzset{
    align=center,
    node distance=4mm,
}
\tikzstyle{every node} = [very thick]
\tikzstyle{arrow} = [thick,draw,->,>=stealth]
\tikzstyle{arrow-reverse} = [thick,draw,<-,>=stealth]
\tikzstyle{add} = [circle,draw,inner sep=1.5pt]
\tikzstyle{block} = [rectangle,rounded corners=1pt,draw,minimum width=7mm,minimum height=7mm,fill=teal!20]
\tikzstyle{conv1x1} = [trapezium,draw,shape border rotate=270,trapezium angle=78,fill=blue!20]
\tikzstyle{convkxk} = [trapezium,draw,shape border rotate=270,trapezium angle=78,fill=red!20]
\tikzstyle{act} = [rectangle,rounded corners=1pt,draw,minimum width=4mm,minimum height=15mm,fill=gray!20]
\tikzstyle{dim} = [ellipse,draw,densely dashed,minimum width=2cm,minimum height=2mm]
\tikzstyle{square} = [rectangle,rounded corners=1pt,minimum width=1mm,minimum height=1mm]
\tikzstyle{token} = [rectangle,rounded corners=2mm,minimum width=8mm,minimum height=5mm,draw,anchor=west,fill=magenta]
\tikzstyle{postoken} = [rectangle,rounded corners=2mm,minimum width=6mm,minimum height=1mm,draw,anchor=west,fill=cyan]

\usepgfplotslibrary{
    patchplots,
    statistics,
}
\pgfplotsset{
    compat=newest,
    grid=both,
    grid style={line width=.1pt, draw=gray!10},
    major grid style={line width=.2pt,draw=gray!50},
    every axis plot/.append style={line width=0.8pt},
}

\newcommand{\vect}[1]{\mathbf{#1}}

\begin{document}

% acronyms
\begin{acronym}
    % \acro{ours}[OURS]{Our Model}
    \acro{ours}[HyVIC]{spatio-spectral \textbf{v}ariational \textbf{hy}perspectral \textbf{i}mage \textbf{c}ompression architecture}
    
    \acro{rsim}[RSiM]{Remote Sensing Image Analysis}
    \acro{bifold}[BIFOLD]{Berlin Institute for the Foundations of Learning and Data}
    \acro{enmap}[EnMAP]{Environmental Mapping and Analysis Program}
    
    \acro{l2a}[L2A]{level 2A}
    \acro{gpu}[GPU]{graphics processing unit}
    \acro{gt}[GT]{ground truth}
    \acro{vram}[VRAM]{video random access memory}
    \acro{uav}[UAV]{unmanned aerial vehicle}
    \acro{ae}[AE]{arithmetic encoder}
    \acro{ad}[AD]{arithmetic decoder}
    \acro{q}[Q]{quantizer}
    \acro{ans}[ANS]{asymmetric numeral systems}
    \acro{llm}[LLM]{large language model}
    \acro{lulc}[LULC]{land use / land cover}

    \acro{1d}[1-D]{one-dimensional}
    \acro{2d}[2-D]{two-dimensional}
    \acro{3d}[3-D]{three-dimensional}

    \acro{rs}[RS]{remote sensing}
    \acro{eo}[EO]{earth obervation}
    \acro{cv}[CV]{computer vision}
    \acro{rgb}[RGB]{red, green and blue}
    \acro{hsi}[HSI]{hyperspectral image}
    \acro{aux}[aux]{auxiliary}
    \acro{swir}[SWIR]{short-wavelength infrared}
    \acro{vnir}[VNIR]{visible and near-infrared}

    \acro{spiht}[SPIHT]{set partitioning in hierarchical trees}
    \acro{speck}[SPECK]{set partitioning embedded block}
    \acro{pca}[PCA]{principle component analysis}
    \acro{dpcm}[DPCM]{differential pulse code modulation}
    \acro{dct}[DCT]{discrete cosine transform}
    \acro{klt}[KLT]{Karhunen–Loève transform}
    
    \acro{cnn}[CNN]{convolutional neural network}
    \acro{ann}[ANN]{artificial neural network}
    \acro{vae}[VAE]{variational autoencoder}
    \acro{cae}[CAE]{convolutional autoencoder}
    \acro{gan}[GAN]{generative adversarial network}
    \acro{inr}[INR]{implicit neural representation}
    \acro{fm}[FM]{foundation model}
    \acro{gsm}[GSM]{gaussian scale mixture}
    \acro{gmm}[GMM]{gaussian mixture model}
    
    \acro{se}[SE]{squeeze and excitation}
    \acro{swin}[Swin]{shifted windows}
    \acro{rstb}[RSTB]{residual swin transformer block}
    \acro{ntu}[NTU]{neural transformation unit}
    \acro{fe}[FE]{feature embedding}
    \acro{fu}[FU]{feature unembedding}
    \acro{stl}[STL]{swin transformer layer}
    \acro{wa}[WA]{window attention}
    \acro{swa}[SWA]{shifted window attention}

    \acro{snr}[SNR]{signal-to-noise ratio}
    \acro{psnr}[PSNR]{peak signal-to-noise ratio}
    \acro{sa}[SA]{spectral angle}
    \acro{mse}[MSE]{mean squared error}
    \acro{ssim}[SSIM]{structural similarity index measure}
    \acro{cr}[CR]{compression ratio}
    \acro{bpppc}[bpppc]{bits per pixel per channel}
    \acro{decibel}[\si{\decibel}]{decibels}
    \acro{gsd}[GSD]{ground sample distance}
    \acro{flops}[FLOPs]{floating-point operations}
    \acro{bdpsnr}[BD-PSNR]{Bjøntegaard delta PSNR}
    \acro{miou}[mIoU]{mean intersection over union}

    \acro{leakyrelu}[LeakyReLU]{leaky rectified linear unit}
    \acro{prelu}[PReLU]{parametric rectified linear unit}
    \acro{mlp}[MLP]{multilayer perceptron}
    \acro{ln}[LN]{layer normalization}
    \acro{gdn}[GDN]{generalized divisive normalization}
    \acro{igdn}[IGDN]{inverse generalized divisive normalization}
    \acro{lr}[LR]{learning rate}
    \acro{bs}[BS]{batch size}
    \acro{rd}[RD]{rate-distortion}

    \acro{1dcae}[1D-CAE]{1D-convolutional autoencoder}
    \acro{a1dcae}[A1D-CAE]{adaptive 1D-convolutional autoencoder}
    \acro{sscnet}[SSCNet]{spectral signals compressor network}
    \acro{3dcae}[3D-CAE]{3D convolutional auto-encoder}
    \acro{linerwkv}[LineRWKV]{line receptance weighted key value}
    \acro{hycot}[HyCoT]{hyperspectral compression transformer}
    \acro{hycass}[HyCASS]{adjustable spatio-spectral hyperspectral image compression network}
    \acro{tic}[TIC]{transformer-based image compression}
    \acro{s2cnet}[S2C-Net]{spatio-spectral compression network}
    \acro{hific}[HiFiC]{high fidelity compression}
    \acro{fiof}[FIO-F]{frame-index-only framework}
    \acro{jscifif}[JSCIFI-F]{joint spatial coordinate index with frame index framework}
    \acro{msahific}[MSAHiFiC]{multi-scale attention spatial-spectral high-fidelity compression network}
    \acro{btcnet}[BTC-Net]{bit-level tensor data compression network}
    \acro{tcm}[TCM]{transformer-CNN mixture}
    \acro{hhcnet}[HCCNet]{hyperspectral compression network via contrastive learning}
    \acro{suncnn}[SUnCNN]{sparse unmixing using unsupervised convolutional neural network}
\end{acronym}

\title{HyVIC: A Metric-Driven Spatio-Spectral Hyperspectral Image Compression Architecture Based on Variational Autoencoders}

\author{%
    Martin Hermann Paul Fuchs\,\orcidlink{0009-0003-7800-284X},
    Behnood Rasti\,\orcidlink{0000-0002-1091-9841},~\IEEEmembership{Senior Member,~IEEE},
    and~Begüm Demir\,\orcidlink{0000-0003-2175-7072},~\IEEEmembership{Senior Member,~IEEE}%
    \thanks{Martin Hermann Paul Fuchs and Begüm Demir are with the Faculty of Electrical Engineering and Computer Science, Technische Universität Berlin, 10587 Berlin, Germany and also with the \acf{bifold}, 10587 Berlin, Germany (e-mail: m.fuchs@tu-berlin.de; demir@tu-berlin.de).}%
    \thanks{Behnood Rasti is with the National Technical University of Athens, 10682 Athens, Greece (e-mail: behnood.rasti@gmail.com).}
}

% The paper headers
\markboth{Journal of \LaTeX\ Class Files,~Vol.~14, No.~8, August~2021}%
{Shell \MakeLowercase{\textit{et al.}}: A Sample Article Using IEEEtran.cls for IEEE Journals}

\maketitle

\begin{abstract}
The rapid growth of hyperspectral data archives in \ac{rs} necessitates effective compression methods for storage and transmission.
Recent advances in learning-based \ac{hsi} compression have significantly enhanced both reconstruction fidelity and compression efficiency.
However, existing methods typically adapt variational image compression models designed for natural images, without adequately accounting for the distinct spatio-spectral redundancies inherent in \acp{hsi}.
In particular, they lack explicit architectural designs to balance spatial and spectral feature learning, limiting their ability to effectively leverage the unique characteristics of hyperspectral data in {\ac{rs}}.
To address this issue, in this paper, we aim to study the effects of spatio-spectral feature learning on the \ac{rd} performance of variational {\ac{hsi}} compression as a first time in \ac{rs}.
To this end, we propose to use configurable spatial and spectral feature learning blocks within variational {\ac{hsi}} compression.
To achieve this, we introduce \ac{ours}, a configurable \ac{vae} for \ac{hsi} compression.
\hl{{\ac{ours}} is specifically designed for ground-based {\ac{hsi}} compression,} enabling independent control of spatial and spectral feature learning to facilitate hyperspectral-specific variational image compression.
Extensive experiments on two benchmark datasets demonstrate that the trade-off between spatial and spectral feature learning is crucial for the reconstruction fidelity.
Motivated by this, we also present a metric-driven strategy to systematically select the hyperparameters of the proposed model.
In detail, {\ac{ours}} achieves high spatial and spectral reconstruction fidelity across a wide range of \acp{cr} and improves the state of the art by up to \SI[round-mode=places,round-precision=2]{4.6585}{\decibel} in terms of \acs{bdpsnr}.
Based on our results, we offer insights and derive practical guidelines to guide future research directions in learning-based variational \ac{hsi} compression in \ac{rs}.
Our code and pre-trained model weights are publicly available at \url{https://git.tu-berlin.de/rsim/hyvic}.
\end{abstract}

\acresetall

\begin{IEEEkeywords}
Hyperspectral image compression, spatio-spectral feature learning, rate–distortion optimization, variational autoencoder, deep learning, remote sensing.
\end{IEEEkeywords}

\section{Introduction}
\label{sec:introduction}
\IEEEPARstart{H}{yperspectral} sensors deployed on satellites, aircraft, and \acp{uav} provide rich spectral information, enabling accurate identification and discrimination of materials on the Earth’s surface.
This supports a wide range of \ac{rs} applications, such as climate \& environmental monitoring \cite{rajabi2024hyperspectral}, urban mapping \& planning \cite{nisha2022current}, defense \& security surveillance \cite{gross2022multi}, disaster monitoring \cite{ye2022remote}, and precision agriculture \cite{pande2023application}.
Ongoing advancements in hyperspectral sensor technology have enabled the acquisition of increasingly detailed spectral and spatial information, which is critical for performing advanced analytical tasks.
However, the substantial data volumes generated by these sensors pose significant challenges with respect to data storage and transmission.
Consequently, the efficient and effective compression of \acp{hsi} has emerged as an essential research topic in \ac{rs} \cite{gomes2025lossy}.

Numerous \ac{hsi} compression methods have been proposed in the literature, which can generally be grouped into three categories:
\begin{enumerate*}[i)]
    \item lossless, e.g. \cite{hernandez2021ccsds, valsesia2024onboard};
    \item near-lossless, e.g. \cite{qian2006near, zheng2022recursive}; and
    \item lossy, e.g. \cite{ryan1997lossy, kuester2023adaptive}
\end{enumerate*}
\ac{hsi} compression.
Each category corresponds to a distinct trade-off between reconstruction fidelity and compression efficiency.
While lossless \ac{hsi} compression enables a perfect reconstruction, the achievable \acp{cr} are typically limited to a factor of \SIrange{2}{4}{} \cite{altamimi2024lossless}.
Near-lossless \ac{hsi} compression can achieve higher \acp{cr} by allowing a small, controllable, and upper-bounded distortion, but the achievable \acp{cr} remain limited.
In contrast, lossy \ac{hsi} compression enables substantially higher \acp{cr} by selectively discarding perceptually irrelevant and less important information, making it well suited to meet the stringent bandwidth and storage constraints in \ac{rs} \cite{gomes2025lossy}.

Traditional methods for lossy \ac{hsi} compression have been extensively investigated.
Early works predominantly rely on conventional signal processing techniques, among which transform coding \cite{goyal2002theoretical} emerged as one of the most widely considered approaches.
They derive compact latent representations of hyperspectral data by applying predefined mathematical transformations, followed by coefficient quantization and subsequent entropy coding, e.g. \cite{du2007hyperspectral, dragotti2000compression, tang2006three}.
Despite their demonstrated effectiveness, traditional transform coding methods rely on linear, hand-crafted transformations that are not specifically optimized for the underlying data distribution, thereby limiting their generalization capability across heterogeneous hyperspectral data.

To address the limitations of traditional transform coding methods, lossy learning-based \ac{hsi} compression, which can model complex non-linear relationships, has recently attracted great attention.
Learning-based models exploit hierarchical and data-adaptive representations learned from large-scale training datasets and have demonstrated improved \ac{rd} performance as well as enhanced generalization across diverse scenes \cite{gomes2025lossy}.
Various learning-based models have been introduced, building upon \acp{gan} \cite{fuchs2024generative, guo2025spectral,wan2025msahific}, \acp{inr} \cite{rezasoltani2024hyperspectral, zhao2025paradigm}, \acp{cae} \cite{kuester20211d, la2022hyperspectral, chong2021end}, and transformer-based autoencoders \cite{fuchs2024hycot}.
Among them, autoencoders have shown particularly strong reconstruction fidelity.
They are trained end-to-end and employ \ac{1d} \cite{kuester20211d,fuchs2024hycot}, \ac{2d} \cite{la2022hyperspectral}, or \ac{3d} \cite{chong2021end} neural network architectures to achieve dimensionality reduction and obtain a compact latent representation.
However, due to their fixed architectural designs, these autoencoders lack mechanisms to dynamically exploit spectral and spatial redundancies under specific compression requirements and sensor characteristics.
Flexible models such as \ac{s2cnet} \cite{sprengel2024learning} and \ac{hycass} \cite{fuchs2025adjustable} address this issue by combining spatial and spectral compression blocks to enable adjustable spatio-spectral \ac{hsi} compression across different \acp{cr} and \acp{gsd}.
Although adjustable spatio-spectral compression exhibits promising results, existing models are solely based on dimensionality reduction.
Consequently, the latent representations may contain redundant information, which could be leveraged to achieve more efficient compression.

To jointly optimize rate and distortion, several learning-based \ac{hsi} compression models \cite{guo2023hyperspectral,mijares2023reduced,park2025hyperspectral} have been introduced that are based on \acp{vae} \cite{kingma2013auto}.
These models employ the hyperprior framework \cite{balle2018variational, minnen2018joint} from natural image compression to jointly reduce the distortion of the reconstruction and the entropy of the quantized latent representation, given a learned prior probability model, which enables lossless entropy coding of the latent.
Despite their advances, existing \ac{vae}-based \ac{hsi} compression models lack proper architectural design to exploit the complex spatio-spectral redundancies inherent in \acp{hsi} as they mainly rely on spatial feature learning.
As a result, current models are constrained in their ability to fully leverage the unique characteristics of \acp{hsi}, limiting both the reconstruction fidelity and the achievable \acp{cr}.

To overcome the above-mentioned critical issues, as a first time in {\ac{rs}}, we aim to study the effects of spatio-spectral feature learning on the {\ac{rd}} performance of variational {\ac{hsi}} compression.
To this end, we propose the use of simple but efficient configurable spatial and spectral feature learning blocks for the {\ac{vae}}-based compression of {\acp{hsi}}.
To achieve this, we introduce \ac{ours}, a configurable {\ac{vae}} based on the mean \& scale hyperprior \cite{minnen2018joint} which enables independent control of spatial and spectral feature learning for effective hyperspectral-specific variational image compression.
The proposed model is \hl{specifically designed for ground-based {\ac{hsi}} compression and} made up of:
\begin{enumerate*}[1)]
    \item a configurable spatio-spectral encoder;
    \item a spatio-spectral hyperencoder;
    \item a spatio-spectral hyperdecoder; and
    \item a configurable spatio-spectral decoder.
\end{enumerate*}
In contrast to existing models, \ac{ours} specifically leverages the spatio-spectral redundancies of \acp{hsi} by incorporating configurable spatial and spectral model components, whose hyperparameters are fixed through a metric-driven selection strategy.
The proposed model is evaluated on the HySpecNet-11k \cite{fuchs2023hyspecnet} and the MLRetSet \cite{omruuzun2024novel} benchmark datasets in order to demonstrate its effectiveness along the \ac{rd} curve.
Our experimental results indicate that variational \ac{hsi} compression is highly dependent on a balanced trade-off between spatial and spectral feature learning.
The main contributions of this paper are summarized as follows:
\begin{itemize}
    \item To the best of our knowledge, we are the first to study the effects of spatio-spectral feature learning on the {\ac{rd}} performance of variational {\ac{hsi}} compression in {\ac{rs}}. In particular, we propose the use of configurable spatial and spectral feature learning blocks within variational image compression \cite{minnen2018joint} to more effectively leverage the unique spatio-spectral redundancies present in {\acp{hsi}}.
    \item Our experimental results demonstrate that the selection of the hyperparameters of the proposed model, which control spatial and spectral feature learning, has a substantial impact on compression performance. Motivated by this, we employ a metric-driven strategy to systematically adapt the model's hyperparameters to the characteristics of \acp{hsi}.
    \item We conduct extensive experiments on two benchmark datasets, including an ablation study, comparisons with other approaches, and analyses of the reconstruction results and errors. The results show the effectiveness and generalization capability of the proposed model, as well as the robustness of the metric-driven hyperparameter selection strategy.
    \item Based on our findings, we provide insights and derive practical guidelines that help shape future research in learning-based variational \ac{hsi} compression in {\ac{rs}}.
\end{itemize}

The remaining part of this paper is organized as follows.
\autoref{sec:related-work} presents the related work.
\autoref{sec:proposed-model} introduces the proposed model.
\autoref{sec:dataset-setup} describes the considered datasets and provides the experimental setup, while the experimental results are carried out in \autoref{sec:experiments}.
\autoref{sec:conclusion} concludes our paper.

\section{Related Work}
\label{sec:related-work}
In this section, we first review recent advances in lossy learning-based \ac{hsi} compression and subsequently discuss variational image compression methods in both \ac{cv} and \ac{rs}.

\subsection{Lossy Learning-Based HSI Compression}
Lossy learning-based \ac{hsi} compression has attracted great attention, as it enables high \acp{cr} at the cost of a small degree of information loss.
Most of the state-of-the-art methods employ the autoencoder architecture, where an encoder performs dimensionality reduction by transforming the input into a lower-dimensional latent space, and a decoder reconstructs the data from this compressed representation.
Existing autoencoder architectures mainly differ in their way of processing the spectral and spatial dimensions.
For example, Küster et al. \cite{kuester20211d,kuester2023adaptive} introduce \ac{1dcae} that leverages \ac{1d} convolutions to learn spectral features and \ac{1d} pooling for spectral compression.
To increase the achievable \acp{cr} and exploit spatial redundancies, La Grassa et al. \cite{la2022hyperspectral} propose \ac{sscnet}, which achieves spatial feature learning and spatial compression through \ac{2d} convolutions and \ac{2d} pooling, respectively.
For a joint feature learning and compression of spatio-spectral redundancies, Chong et al. \cite{chong2021end} present \ac{3dcae} that stacks multiple \ac{3d} convolutions with stride.
% However, the limited receptive field of these convolutional architectures restricts their ability to fully capture spatial and spectral redundancies.
Motivated by the effectiveness of spectral feature learning at low \acp{cr} \cite{fuchs2023hyspecnet}, \ac{hycot} \cite{fuchs2024hycot} employs a transformer-based autoencoder for pixelwise compression of \ac{hsi} that exploits long-range spectral redundancies.
% However, due to their fixed architectural designs, these autoencoders lack mechanisms to dynamically exploit spectral and spatial redundancies under specific compression requirements and sensor characteristics.
Flexible models such as \ac{s2cnet} \cite{sprengel2024learning} and \ac{hycass} \cite{fuchs2025adjustable} combine spatial and spectral compression blocks, enabling adjustable spatio-spectral \ac{hsi} compression that adapts to both compression requirements and sensor characteristics.

In addition to autoencoders, Fuchs et al. \cite{fuchs2024generative} introduce \acs{hific}\textsubscript{\acs{se}}, a \ac{gan} designed for spatio-spectral compression of \acp{hsi} that extends the \ac{hific} framework \cite{mentzer2020high} by incorporating \ac{se} blocks to effectively leverage spectral redundancies alongside spatial redundancies.
Another \ac{gan}-based model, named \ac{msahific}, is proposed by Wan et al. \cite{wan2025msahific}, which employs a superprior network to guide \ac{rd} optimization, enhancing performance under constrained bitrate conditions.
To strengthen spectral feature modeling, the network employs a multiscale spectral attention module, and they integrate the spectral reconstruction fidelity term into the loss function.
Rezasoltani et al. \cite{rezasoltani2024hyperspectral} propose conducting lossy \ac{hsi} compression by using \acp{inr}, where a \ac{mlp} network learns to map pixel locations to their spectral signatures, with the network weights serving as a compressed representation of the \ac{hsi}.
Zhao et al. \cite{zhao2025paradigm} extend this approach by interpreting hyperspectral data as a video sequence, where each spectral band is treated as an individual frame of information.
Variations across spectral bands are modeled as transformations between frames, and neural video representation is used to encode the data within the network parameters.
Zhou et al. \cite{zhou2023btc} introduce \ac{btcnet} for onboard {\ac{hsi}} compression, which leverages a data-driven lightweight quantized neural encoder with two-stage bit compression.
Building upon this, the authors subsequently propose {\ac{btcnet}} V2 \cite{zhou2026btc}, incorporating a large-kernel convolutional encoder and a spatial-priority hierarchical decoder that prioritizes spatial upsampling to improve reconstruction performance.

\subsection{Variational Image Compression in CV and RS}
% Despite the demonstrated effectiveness of the above-mentioned \ac{hsi} compression works,
Recent advances in \ac{cv} are largely driven by variational image compression frameworks \cite{balle2018variational, minnen2018joint}, which have subsequently inspired similar compression works in \ac{rs}.
% These methods learn a probabilistic latent representation together with an associated prior probability model, enabling end-to-end \ac{rd} optimization.
% Over the past decade, learning-based natural image compression has advanced significantly within the \ac{cv} community, resulting in substantial \ac{rd} performance improvements.
Variational image compression is introduced by Ballé et al. \cite{balle2017endtoend} as a non-linear transform coding approach that can be trained end-to-end.
The objective is to reduce the entropy of the latent representation under a prior probability model, referred to as the entropy model, which is shared between the encoder and decoder.
This enables end-to-end optimization of both compression efficiency and reconstruction fidelity using a \ac{rd} loss.
In this context, they propose the fully-factorized prior, a \ac{cnn}-based \ac{vae} with a fixed, non-parametric, and fully-factorized entropy model that is learned during training.
Building upon this work, Ballé et al. \cite{balle2018variational} propose the scale-only hyperprior, which employs a conditional \ac{gsm} as an image-dependent entropy model.
The scale parameters of the \ac{gsm} are inferred from additionally transmitted side information, denoted as the hyperlatent, which effectively models the spatial redundancies inside the latent representation.
This model is subsequently extended by Minnen et al. \cite{minnen2018joint} to the mean \& scale hyperprior, replacing the conditional \ac{gsm} with a conditional \ac{gmm} and jointly deriving mean and scale parameters from the hyperlatent.
Furthermore, they introduce the context + hyperprior model, which extends the mean \& scale hyperprior with a complementary autoregressive context model, where both components are jointly optimized.
Apart from \ac{cnn}-based models, hybrid models have been proposed that combine transformer layers with convolutional layers to jointly capture long-range and short-range redundancies \cite{lu2021transformer, liu2023learned}.

In recent years, several learning-based \ac{hsi} compression works based on variational image compression have been proposed in \ac{rs}.
For example, Verdú et al. \cite{mijares2023scalable} introduce a scalable \ac{hsi} compression model that leverages channel clustering to lower computational demand while enabling adaptation to hyperspectral sensors with varying numbers of spectral bands.
This work is further extended in \cite{mijares2023reduced} to support multirate compression of panchromatic, multispectral, and hyperspectral \ac{rs} images.
In \cite{guo2023hyperspectral}, Guo et al. propose \ac{hhcnet} to tackle attribute collapse inside the latent representation, which is caused by the inherent properties of \acp{hsi} and the effects of quantization during compression.
To this end, the network learns discriminative representations instead of relying solely on \ac{rd} optimization.
In \cite{park2025hyperspectral}, Park et al. propose hyperspectral-\acs{vae} for joint spatio-spectral compression of \acp{hsi} while preserving key atmospheric information.
However, existing variational \ac{hsi} compression models are constrained in effectively leveraging the spatio-spectral redundancies of \acp{hsi}, as their architectures do not explicitly facilitate a balanced spatial and spectral feature learning.

\section{Proposed Spatio-Spectral Variational Hyperspectral Image Compression Architecture (H\textnormal{y}VIC)}
\label{sec:proposed-model}
Let $\vect{X} \in \mathbb{R}^{H \times W \times C}$ denote an \ac{hsi}, where $H$, $W$ and $C$ represent image height, image width and the number of spectral bands, respectively.
To compress $\vect{X}$, in this paper we focus our attention on variational \ac{hsi} compression.
In particular, we present \ac{ours} to facilitate effective spatio-spectral \ac{hsi} compression \hl{at the ground segment}.
The proposed model builds upon the mean \& scale hyperprior \cite{minnen2018joint} by introducing simple but efficient configurable spatial and spectral feature learning blocks that enable to more effectively leverage the unique spatio-spectral redundancies present in \acp{hsi}.
Variational \ac{hsi} compression aims at transforming $\vect{X}$ into a compact latent representation $\vect{Y} \in \mathbb{R}^{\Sigma \times \Omega \times M}$ with latent height $\Sigma$, latent width $\Omega$ and $M$ latent channels.
$\vect{Y}$ can be quantized to $\vect{\hat{Y}} \in \mathbb{Z}^{\Sigma \times \Omega \times M}$ and encoded into a bitstream $\vect{b} \in ( \num{0}, \num{1} )^L$ of length $L$ using entropy coding (i.e. arithmetic coding \cite{rissanen1979arithmetic}) based on an entropy model. 
Simultaneously, $\vect{\hat{Y}}$ should preserve sufficient information to enable the lossy reconstruction of $\vect{\hat{X}} \in \mathbb{R}^{H \times W \times C}$ with minimal distortion.
The training objective is to jointly minimize the expected length of $\vect{b}$ and the expected distortion between $\vect{X}$ and $\vect{\hat{X}}$, resulting in a \ac{rd} optimization problem.

\begin{figure*}
    \centering
    \begin{tikzpicture}[node distance=4mm and 2mm]
        % E
        \node[inner sep=0pt] (enc-in) {\includegraphics[width=.118\linewidth]{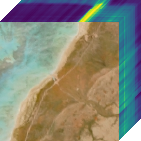}};
        \node[conv1x1,right=of enc-in] (enc-conv-0) {Conv2D \\ $C \rightarrow N$ \\ \qtyproduct{1x1}{}};
        \node[act,right=of enc-conv-0] (enc-act-0) {\rotatebox{90}{\acs{gdn}}};
        \node[right=1.5mm of enc-act-0,inner sep=0pt,outer sep=0pt] (enc-tmp-0) {$\bullet$};
        \node[convkxk,right=1.5mm of enc-tmp-0] (enc-conv-1) {Conv2D \\ $N \rightarrow N$ \\ $k \times k$ \\ $\downarrow^2$};
        \node[act,right=of enc-conv-1] (enc-act-1) {\rotatebox{90}{\acs{gdn}}};
        \node[right=1.5mm of enc-act-1,inner sep=0pt,outer sep=0pt] (enc-tmp-1) {$\bullet$};
        \node[conv1x1,right=1.5mm of enc-tmp-1] (enc-conv-2) {Conv2D \\ $N \rightarrow M$ \\ \qtyproduct{1x1}{}};

        % arithmetic coder main
        \node[block,below right=2mm of enc-conv-2] (q-0) {\acs{q}};
        \node[block,below=of q-0] (ae-0) {\acs{ae}};
        \node[block,below=of ae-0,pattern=checkerboard,minimum width=9mm,minimum height=2.5mm] (bit-0) {};
        \node[block,below=of bit-0] (ad-0) {\acs{ad}};
        \node[below=of ad-0,inner sep=0pt,outer sep=0pt] (dec-tmp-2) {\phantom{$\bullet$}};

        % E_H
        \node[conv1x1,above right=2mm of q-0] (enc-conv-3) {Conv2D \\ $M \rightarrow N$ \\ \qtyproduct{1x1}{}};
        \node[act,right=of enc-conv-3] (enc-act-3) {\rotatebox{90}{\acs{leakyrelu}}};
        \node[convkxk,right=of enc-act-3] (enc-conv-4) {Conv2D \\ $N \rightarrow N$ \\ $k \times k$ \\ $\downarrow^2$};
        \node[act,right=of enc-conv-4] (enc-act-4) {\rotatebox{90}{\acs{leakyrelu}}};
        \node[convkxk,right=of enc-act-4] (enc-conv-5) {Conv2D \\ $N \rightarrow N$ \\ $k \times k$ \\ $\downarrow^2$};

        % arithmetic coder side
        \node[block,below right=of enc-conv-5,yshift=2.3mm] (q-1) {$\acs{q}^\mathcal{H}$};
        \node[block,below=of q-1] (ae-1) {$\acs{ae}^\mathcal{H}$};
        \node[block,below=of ae-1,pattern=checkerboard,minimum width=9mm,minimum height=2.5mm] (bit-1) {};
        \node[block,below=of bit-1] (ad-1) {$\acs{ad}^\mathcal{H}$};

        % D_H
        \node[convkxk,below left=of ad-1] (dec-conv-5) {Conv2D \\ $N \rightarrow M$ \\ $k \times k$ \\ $\uparrow^2$};
        \node[act,left=of dec-conv-5] (dec-act-5) {\rotatebox{90}{\acs{leakyrelu}}};
        \node[convkxk,left=of dec-act-5] (dec-conv-4) {Conv2D \\ $M \rightarrow \frac{3}{2}M$ \\ $k \times k$ \\ $\uparrow^2$};
        \node[act,left=of dec-conv-4] (dec-act-4) {\rotatebox{90}{\acs{leakyrelu}}};
        \node[conv1x1,right=8mm of bit-0] (dec-conv-3) {Conv2D \\ $\frac{3}{2}M \rightarrow 2M$ \\ \qtyproduct{1x1}{}};
        \node[left=3mm of dec-conv-3,inner sep=0pt,outer sep=0pt] (dec-tmp-9) {$\bullet$};
        
        % D
        \node[conv1x1,below left=of ad-0] (dec-conv-2) {Conv2D \\ $M \rightarrow N$ \\ \qtyproduct{1x1}{}};
        \node[act,left=of dec-conv-2] (dec-act-2) {\rotatebox{90}{\acs{igdn}}};
        \node[left=1.5mm of dec-act-2,inner sep=0pt,outer sep=0pt] (dec-tmp-1) {$\bullet$};
        \node[convkxk,left=1.5mm of dec-tmp-1] (dec-conv-1) {Conv2D \\ $N \rightarrow N$ \\ $k \times k$ \\ $\uparrow^2$};
        \node[act,left=of dec-conv-1] (dec-act-1) {\rotatebox{90}{\acs{igdn}}};
        \node[left=1.5mm of dec-act-1,inner sep=0pt,outer sep=0pt] (dec-tmp-0) {$\bullet$};
        \node[conv1x1,left=of dec-tmp-0] (dec-conv-0) {Conv2D \\ $N \rightarrow C$ \\ \qtyproduct{1x1}{}};
        \node[left=of dec-conv-0, inner sep=0pt] (dec-out) {\includegraphics[width=.118\linewidth]{img/hsi.pdf}};

        % MISC
        \node (enc-tmp-2) at ($(enc-conv-2.east)!0.5!(enc-conv-3.west)+(-0.03,0)$) {$\bullet$};
        
        % ARROWS
        \draw[arrow] (enc-in) -- (enc-conv-0);
        \draw[arrow] (enc-conv-0) -- (enc-act-0);
        \draw[arrow] (enc-act-0) -- (enc-conv-1);
        \draw[arrow] (enc-conv-1) -- (enc-act-1);
        \draw[arrow] (enc-act-1) -- (enc-conv-2);

        \draw[arrow] (enc-tmp-2.center) -- (q-0);
        \draw[arrow] (q-0) -- node[left] {$\vect{\hat{Y}}$} (ae-0);
        \draw[arrow] (ae-0) -- (bit-0);
        \draw[arrow] (bit-0) -- (ad-0);
        \draw[arrow] (ad-0) |- node[below] {$\vect{\hat{Y}}$} (dec-conv-2);

        \draw[arrow] (enc-conv-2) -- node[above,yshift=0.5mm] {$\vect{Y}$} (enc-conv-3);
        \draw[arrow] (enc-conv-3) -- (enc-act-3);
        \draw[arrow] (enc-act-3) -- (enc-conv-4);
        \draw[arrow] (enc-conv-4) -- (enc-act-4);
        \draw[arrow] (enc-act-4) -- (enc-conv-5);

        \draw[arrow] (enc-conv-5) -| node[above] {$\vect{Z}$} (q-1);
        \draw[arrow] (q-1) -- node[right] {$\vect{\hat{Z}}$} (ae-1);
        \draw[arrow] (ae-1) -- (bit-1);
        \draw[arrow] (bit-1) -- (ad-1);
        \draw[arrow] (ad-1) |- node[below] {$\vect{\hat{Z}}$} (dec-conv-5);

        \draw[arrow] (dec-conv-5) -- (dec-act-5);
        \draw[arrow] (dec-act-5) -- (dec-conv-4);
        \draw[arrow] (dec-conv-4) -- (dec-act-4);
        \draw[arrow] (dec-act-4) -| ++(-5mm,12mm) -- ++(20mm,4.5mm) |- (dec-conv-3);

        \draw[arrow] (dec-conv-3) -- (dec-tmp-9) |- ++(0,9mm) -- node[near start,above] {$\bm{\hat{\mu}}, \bm{\hat{\sigma}}$} (ae-0);
        \draw[arrow] (dec-tmp-9) |- ++(0,-9mm) -- node[near start,below] {$\bm{\hat{\mu}}, \bm{\hat{\sigma}}$} (ad-0);

        \draw[arrow] (dec-conv-2) -- (dec-act-2);
        \draw[arrow] (dec-act-2) -- (dec-conv-1);
        \draw[arrow] (dec-conv-1) -- (dec-act-1);
        \draw[arrow] (dec-act-1) -- (dec-conv-0);
        \draw[arrow] (dec-conv-0) -- (dec-out);

        \draw[arrow] (enc-tmp-1.center) -- ++(0,1.4) -| node[pos=0.25, above] (enc-N) {$S \times$} (enc-tmp-0);
        \draw[arrow] (dec-tmp-0.center) -- ++(0,-1.4) -| node[pos=0.25, below] (dec-N) {$S \times$} (dec-tmp-1);

        % FITS
        \node[draw=cyan,dotted,fit=(enc-conv-0)(enc-conv-1)(enc-N)(enc-conv-2),inner sep=0.6mm,yshift=-1.5mm,label=below:{\textcolor{cyan}{$E_\Phi$}}] {};
        \node[draw=magenta,dotted,fit=(dec-conv-0)(dec-conv-1)(dec-N)(dec-conv-2),inner sep=0.6mm,yshift=1.6mm,label=above:{\textcolor{magenta}{$D_{\Phi'}$}}] {};

        \node[draw=cyan,dotted,fit=(enc-conv-3)(enc-conv-5),inner xsep=0.6mm,inner ysep=2.2mm,label=above:{\textcolor{cyan}{$E^\mathcal{H}_\Psi$}}] {};
        \node[draw=magenta,dotted,fit=(dec-conv-3)(dec-conv-5),inner xsep=0.6mm,inner ysep=2.8mm,label=below:{\textcolor{magenta}{$D^\mathcal{H}_\Gamma$}}] {};

        % ANNOTATIONS
        \node[below=0.2mm of enc-in] {$\vect{X}$};
        \node[below=0.2mm of dec-out] {$\vect{\hat{X}}$};
    \end{tikzpicture}
    \caption{Block diagram of the proposed \ac{ours} model.
    Initially, the encoder $E_\Phi$ transforms the input \ac{hsi} $\vect{X}$ into its latent representation $\vect{Y}$, which is subsequently transformed by the hyperencoder $E^\mathcal{H}_\Psi$ to the hyperlatent $\vect{Z}$.    
    $\vect{Y}$ and $\vect{Z}$ are quantized to $\vect{\hat{Y}}$ and $\vect{\hat{Z}}$, compressed into a bitstream and subsequently reconstructed using arithmetic coding.
    $\vect{\hat{Z}}$ serves as side information to estimate both mean $\boldsymbol{\hat{\mu}}$ and scale $\boldsymbol{\hat{\sigma}}$ parameters via the hyperdecoder $D^\mathcal{H}_\Gamma$, which are used inside the conditional \ac{gmm}-based entropy model to encode $\vect{\hat{Y}}$.
    In conrast, $\vect{\hat{Z}}$ is encoded using a fully-factorized entropy model.
    Finally, the decoder $D_{\Phi'}$ reconstructs $\vect{\hat{X}}$ based on $\vect{\hat{Y}}$.
    }
    \label{fig:blockdiagram}
\end{figure*}

As shown in \autoref{fig:blockdiagram}, \ac{ours} is composed of four main components:
\begin{enumerate*}[i)]
    \item a configurable encoder $E_\Phi$ that transforms the input \ac{hsi} into its latent representation by exploiting both short-range spatial and long-range spectral redundancies;
    \item a hyperencoder $E^\mathcal{H}_\Psi$ that leverages spatio-spectral redundancies present in the latent to form the hyperlatent;
    \item a hyperdecoder $D^\mathcal{H}_\Gamma$ that estimates the parameters used for entropy coding from the quantized hyperlatent; and
    \item a configurable decoder $D_{\Phi'}$ that reconstructs the \ac{hsi} based on the short-range spatial and long-range spectral information present inside the quantized latent.
\end{enumerate*}
During our experiments, we demonstrate that the configurable trade-off between spatial and spectral feature learning inside \ac{ours} has a substantial impact on reconstruction fidelity.
To appropriately select these respective hyperparameters, we employ a metric-driven selection strategy that is guided by the \ac{bdpsnr} \cite{bjontegaard2001calculation}, a quantitative evaluation metric commonly employed in video compression.
This ensures a well-balanced trade-off between spatial and spectral feature learning, which optimizes reconstruction quality across the entire \ac{rd} curve.

In detail, the operational flow of \ac{ours} can be described as follows:
First, the encoder $E_\Phi$ transforms the original \ac{hsi} $\vect{X}$ into the latent representation $\vect{Y}$, which is subsequently processed by the hyperencoder $E^\mathcal{H}_\Psi$ forming the hyperlatent representation $\vect{Z}$.
$\vect{Z}$ is quantized to $\vect{\hat{Z}}$ and entropy-coded by the \acl{ae} $\text{\ac{ae}}^\mathcal{H}$ into a bitstream, which serves as side information.
The \acl{ad} $\text{\ac{ad}}^\mathcal{H}$ is used to losslessly reconstruct $\vect{\hat{Z}}$ from the bitstream.
Both $\text{\ac{ae}}^\mathcal{H}$ and $\ac{ad}^\mathcal{H}$ use the same fixed, non-parametric and fully-factorized entropy model that is learned during training.
The hyperdecoder $D^\mathcal{H}_\Gamma$ is responsible to transform the hyperlatent $\vect{\hat{Z}}$ into both the mean $\boldsymbol{\hat{\mu}}$ and scale $\boldsymbol{\hat{\sigma}}$ parameters of the conditional \ac{gmm} used to estimate the probability distribution for the \acf{ae} and \acf{ad}.
The quantized latent $\vect{\hat{Y}}$ is then entropy-coded using the estimated entropy model.
Finally, the decoder $D_{\Phi'}$ reconstructs $\vect{\hat{X}}$ based on $\vect{\hat{Y}}$.
A comprehensive description of the proposed model is presented in the following subsections.

\subsection{HyVIC Encoder}
Initially, the encoder $E_\Phi: \mathbb{R}^{H \times W \times C} \rightarrow \mathbb{R}^{\Sigma \times \Omega \times M}$ of the proposed \ac{ours} model transforms the original \ac{hsi} $\vect{X}$ into its latent representation $\vect{Y}$, where $\vect{Y} = E_\Phi ( \vect{X} )$.
$E_\Phi$ leverages spatio–spectral redundancies by sequentially stacking configurable spectral and spatial processing blocks.
First, a pixelwise convolution is applied, which is realized via a \ac{2d} convolutional layer using a \qtyproduct{1x1}{} kernel.
While \ac{2d} convolutions are generally employed to model spatial relationships, using a \qtyproduct{1x1}{} kernel guarantees that the convolution operates on each pixel location independently, without combining information from neighboring spatial locations.
The pixelwise convolution exploits spectral redundancies across the entire spectral signature of each pixel by mapping the high-dimensional number of spectral bands $C$ into a feature space using $N$ filters.
Afterwards, a non-linear \ac{gdn} \cite{balle2016density} is employed as the activation function, performing local divisive normalization, a transformation that has been demonstrated to be especially effective for image density modeling and compression \cite{balle2016density,balle2017endtoend}.
Following the spectral feature extraction, \ac{ours} subsequently leverages the spatial redundancies by stacking a configurable number of spatial stages $S$, each consisting of a \ac{2d} convolutional layer that captures local spatial redundancies using a $k \times k$ convolutional kernel with a stride of \num{2}, combined with a \ac{gdn} activation function.
This progressively decreases the spatial dimensions while simultaneously enriching the feature space with spatially contextual information.
Afterwards, another pixelwise convolution is applied to fuse the attained features maps, yielding the latent representation $\vect{Y}$ that captures spatio-spectral features.

$\vect{Y}$ is subsequently discretized by a quantizer $Q: \mathbb{R}^{\Sigma \times \Omega \times M} \rightarrow \mathbb{Z}^{\Sigma \times \Omega \times M}$, resulting in the quantized latent $\vect{\hat{Y}} = Q ( \vect{Y} )$, where $Q ( \vect{Y} ) = \lfloor \vect{Y} \rceil$ denotes a rounding operation that maps $\vect{Y}$ to its closest integer.
Because quantization yields zero gradients almost everywhere, $Q ( \vect{Y} )$ is substituted with additive uniform noise during training to enable gradient-based optimization \cite{balle2017endtoend}. % $\mathcal{U} (\vect{Y} - \num{0.5}, \vect{Y} + \num{0.5} )$
$\vect{\hat{Y}}$ can be losslessly compressed into a bitstream via entropy coding, which relies on a prior distribution $p_\vect{\hat{Y}} ( \vect{\hat{Y}} )$.
To estimate $p_\vect{\hat{Y}} ( \vect{\hat{Y}} )$, we employ the mean \& scale hyperprior architecture \cite{minnen2018joint}, which introduces a hyperencoder and a hyperdecoder.

\subsection{HyVIC Hyperencoder}
The \ac{ours} hyperencoder $E^\mathcal{H}_\Psi: \mathbb{R}^{\Sigma \times \Omega \times M} \rightarrow \mathbb{R}^{\Pi \times \Xi \times N}$ transforms the latent $\vect{Y}$ into the hyperlatent $\vect{Z}$, where $\vect{Z} = E^\mathcal{H}_\Psi ( \vect{Y} )$.
$\vect{Z}$ contains side information that models the spatio-spectral redundancies among the elements of $\vect{Y}$.
First, spectral information is extracted via a pixelwise convolution that reduces the channel dimension from $M$ to $N$.
A \ac{leakyrelu} is applied to introduce non-linearity and enhance the capability to learn complex patterns from data.
Subsequently, two \ac{2d} convolutional layers with kernel size $k$ and stride \num{2} are applied to integrate spatial redundancies into $\vect{Z}$, thereby producing a compact hyperlatent representation that captures both spectral and spatial information.
An additional \ac{leakyrelu} activation function is present between these two layers to enable the model to capture complex non-linear relationships.

Discretization is achieved via the quantizer $Q^\mathcal{H} : \mathbb{R}^{\Pi \times \Xi \times N} \rightarrow \mathbb{Z}^{\Pi \times \Xi \times N}$, where $\vect{\hat{Z}} = Q^\mathcal{H} ( \vect{Z} )$.
As in the \ac{ours} encoder, quantization is replaced by additive uniform noise during training.
The quantized hyperlatent $\vect{\hat{Z}}$ is entropy-coded and transmitted as side information.
This is done via $\ac{ae}^\mathcal{H}$ and $\ac{ad}^\mathcal{H}$ that share a fixed, non-parametric and fully-factorized entropy model \cite{balle2017endtoend}, which is learned during traning.
% The non-parametric, fully factorized entropy model \cite{balle2018variational} is as follows:
% \begin{align}
%     p_\vect{\hat{Z}} ( \vect{\hat{Z}} ) = \prod_i \left( p_{Z_i} ... * \mathcal{U} \left( -\frac{1}{2}, \frac{1}{2} \right) \right) ( \hat{Z}_i )
% \end{align}

\subsection{HyVIC Hyperdecoder}
The hyperdecoder $D^\mathcal{H}_\Gamma: \mathbb{Z}^{\Pi \times \Xi \times N} \rightarrow \mathbb{R}^{\Sigma \times \Omega \times M} \times \mathbb{R}^{\Sigma \times \Omega \times M}$ of \ac{ours} uses the quantized hyperlatent $\vect{\hat{Z}}$ to estimate the spatial distribution of means $\boldsymbol{\hat{\mu}}$ and the spatial distribution of standard deviations $\boldsymbol{\hat{\sigma}}$, i.e., $( \boldsymbol{\hat{\mu}}, \boldsymbol{\hat{\sigma}} ) = D^\mathcal{H}_\Gamma ( \vect{\hat{Z}} )$.
$\boldsymbol{\hat{\mu}}$ and $\boldsymbol{\hat{\sigma}}$ are used inside the conditional \ac{gmm} to perform entropy coding.
$D^\mathcal{H}_\Gamma$ provides an image-dependent entropy model, which effectively reduces the expected bitrate.
The transmission of additional side information for each \ac{hsi} has been shown to require fewer bits than the encoding of every \ac{hsi} based on a fixed entropy model \cite{minnen2018joint}.
To effectively model $\boldsymbol{\hat{\mu}}$ and $\boldsymbol{\hat{\sigma}}$, $D^\mathcal{H}_\Gamma$ first restores the spatial dimension using two \ac{2d} convolutional layers with kernel size $k$ and a stride of \num{2}, thereby performing learned upsampling.
To enhance the representational capacity, the number of feature channels is increased to $M$ and $\frac{3}{2} M$ in the first and second convolutional layer, respectively, following \cite{minnen2018joint}.
Each convolution is followed by a \ac{leakyrelu} activation to introduce non-linearity and improve the modeling of complex spatial dependencies.
Afterwards, a pixelwise convolution is employed to fuse the spatial feature maps across all channels, resulting in a joint spatio-spectral feature representation.
In this process, the channel dimension is increased from $\frac{3}{2} M$ to $2M$, as both $\boldsymbol{\hat{\mu}}$ and $\boldsymbol{\hat{\sigma}}$ require $M$ channels each.

Each element $\hat{Y}_i$ is modeled as a Gaussian with mean $\mu_i$ and standard deviation $\sigma_i$, convolved with a standard uniform distribution \cite{minnen2018joint}:
\begin{equation}
    \begin{aligned}
        p_\vect{\hat{Y} \vert \vect{\hat{Z}}} ( \vect{\hat{Y}} \vert \vect{\hat{Z}} , \vect{\Gamma}) = &\prod_i \left( \mathcal{N} \left( \mu_i, \sigma_i^2 \right) * \mathcal{U} \left( -\frac{1}{2}, \frac{1}{2} \right) \right) ( \hat{Y}_i ) , \\
        &\text{with } ( \boldsymbol{\hat{\mu}}, \boldsymbol{\hat{\sigma}} ) = D^\mathcal{H}_\Gamma ( \vect{\hat{Z}} ) .
        \label{eq:gmm}
    \end{aligned}
\end{equation}
Based on this conditional \ac{gmm}, arithmetic coding \cite{rissanen1979arithmetic} is employed, which compresses and reconstructs $\vect{\hat{Y}}$ using \ac{ae} and \ac{ad}, respectively.

\subsection{HyVIC Decoder}
Finally, the \ac{ours} decoder $D_{\Phi'}: \mathbb{Z}^{\Sigma \times \Omega \times M} \rightarrow \mathbb{R}^{H \times W \times C}$ is responsible for reconstructing $\vect{\hat{X}}$ based on the quantized latent $\vect{\hat{Y}}$, where $\vect{\hat{X}} = D_{\Phi'} ( \vect{\hat{Y}} )$.
This is achieved by first extracting spectral features with a pixelwise convolution that reduces the channels from $M$ to $N$.
An \ac{igdn} \cite{balle2016density} is subsequently employed as the activation function, as studies have demonstrated its effectiveness in the context of image compression \cite{balle2016density,balle2017endtoend}.
Afterwards, a \ac{2d} convolutional layer with kernel size $k$ and a stride of \num{2} in combination with an \ac{igdn} activation is stacked $S$ times, to progressively recover spatial information and restore the original image resolution.
Finally, the spatio-spectral feature maps are further fused along the spectral dimension by a pixelwise convolution, which simultaneously adjusts the number of output channels to match the number of spectral bands $C$.

\subsection{Rate-Distortion Loss}
To train \ac{ours}, we formulate the training objective as a \ac{rd} optimization problem \cite{balle2017endtoend} with a Lagrangian multiplier $\lambda$.
The corresponding \ac{rd} loss $\mathcal{L_{RD}}$ is given as:
\begin{equation}
    \begin{aligned}
        \mathcal{L_{RD}} &= \mathcal{L_R} + \lambda \cdot \mathcal{L_D}\\
        % &= \mathcal{R}_\text{bpppc} ( \vect{\hat{Y}} ) + \mathcal{R}_\text{bpppc} ( \vect{\hat{Z}} ) + \lambda \cdot \mathbb{E}_{\vect{X} \sim p_\vect{X}} [ \mathcal{D} ( \vect{X}, \vect{\hat{X}} ) ] \\
        &= \frac{1}{H \cdot W \cdot C} \cdot \mathbb{E}_{\vect{X} \sim p_\vect{X}} [ - \log_2 p_\vect{\hat{Y}} ( \vect{\hat{Y}} ) - \log_2 p_\vect{\hat{Z}} ( \vect{\hat{Z}} ) ] \\
        &\quad + \lambda \cdot \mathbb{E}_{\vect{X} \sim p_\vect{X}} [ \mathcal{D} ( \vect{X}, \vect{\hat{X}} ) ] ,
    \end{aligned}
    \label{eq:rdloss}
\end{equation}
where $\mathcal{L_R}$ represents the normalized bitrate required to encode the latent and hyperlatent.
$\mathcal{L_D}$ denotes the distortion introduced by the compression model, evaluated over the sensor-specific dynamic range, and $\lambda$ is a Lagrange multiplier determining the desired trade-off between compression efficiency and reconstruction quality.
$p_\vect{X}$ denotes the unknown probability distribution of the original \acp{hsi}.
$p_\vect{\hat{Y}}$ and $p_\vect{\hat{Z}}$ are the learned entropy models of the quantized latent and hyperlatent, respectively.

As the distortion measure, we employ the \ac{mse} defined as:
\begin{align}
    \text{MSE}(\vect{X}, \vect{\hat{X}}) = \frac{1}{H \cdot W \cdot C} \sum_{h,w,c} \left( \vect{X}(h,w,c) - \vect{\hat{X}}(h,w,c) \right)^2 .
    \label{eq:mse}
\end{align}
Thus, since $\mathcal{D} ( \vect{X}, \vect{\hat{X}} )$ corresponds to a Gaussian likelihood with closed-form expression, \ac{ours} can be interpreted as a \ac{vae} \cite{kingma2013auto}.
% The rate terms $\mathcal{R}$ correspond to the cross entropy between the marginal distributions of the latent and hyperlatent and the learned entropy model, which is minimized when the distributions are identical. 

\section{Dataset Description And Experimental Setup}
\label{sec:dataset-setup}

\subsection{Dataset Description}
To assess the performance of the proposed model, we carried out our experiments on two benchmark datasets described in the following.

\subsubsection{HySpecNet-11k}
\begin{figure*}
    % first row
    \begin{minipage}[b]{.118\linewidth}
        \centering
        \includegraphics[width=\linewidth]{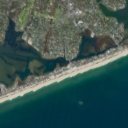}
    \end{minipage}
    \hfill
    \begin{minipage}[b]{.118\linewidth}
        \centering
        \includegraphics[width=\linewidth]{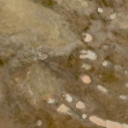}
    \end{minipage}
    \hfill
    \begin{minipage}[b]{.118\linewidth}
        \centering
        \includegraphics[width=\linewidth]{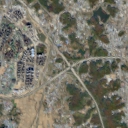}
    \end{minipage}
    \hfill
    \begin{minipage}[b]{.118\linewidth}
        \centering
        \includegraphics[width=\linewidth]{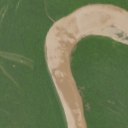}
    \end{minipage}
    \hfill
    \begin{minipage}[b]{.118\linewidth}
        \centering
        \includegraphics[width=\linewidth]{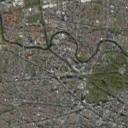}
    \end{minipage}
    \hfill
    \begin{minipage}[b]{.118\linewidth}
        \centering
        \includegraphics[width=\linewidth]{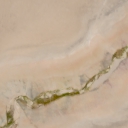}
    \end{minipage}
    \hfill
    \begin{minipage}[b]{.118\linewidth}
        \centering
        \includegraphics[width=\linewidth]{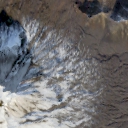}
    \end{minipage}
    \hfill
    \begin{minipage}[b]{.118\linewidth}
        \centering
        \includegraphics[width=\linewidth]{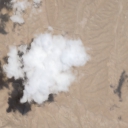}
    \end{minipage}
    % second row
    \begin{minipage}[b]{.118\linewidth}
        \vspace{.055\linewidth}
        \centering
        \includegraphics[width=\linewidth]{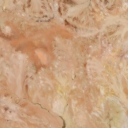}
    \end{minipage}
    \hfill
    \begin{minipage}[b]{.118\linewidth}
        \centering
        \includegraphics[width=\linewidth]{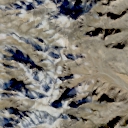}
    \end{minipage}
    \hfill
    \begin{minipage}[b]{.118\linewidth}
        \centering
        \includegraphics[width=\linewidth]{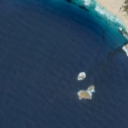}
    \end{minipage}
    \hfill
    \begin{minipage}[b]{.118\linewidth}
        \centering
        \includegraphics[width=\linewidth]{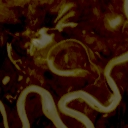}
    \end{minipage}
    \hfill
    \begin{minipage}[b]{.118\linewidth}
        \centering
        \includegraphics[width=\linewidth]{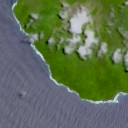}
    \end{minipage}
    \hfill
    \begin{minipage}[b]{.118\linewidth}
        \centering
        \includegraphics[width=\linewidth]{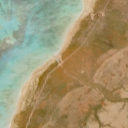}
    \end{minipage}
    \hfill
    \begin{minipage}[b]{.118\linewidth}
        \centering
        \includegraphics[width=\linewidth]{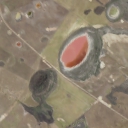}
    \end{minipage}
    \hfill
    \begin{minipage}[b]{.118\linewidth}
        \centering
        \includegraphics[width=\linewidth]{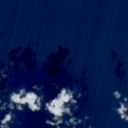}
    \end{minipage}
    % \vspace*{-1em}
    \caption{True color representations of exemplary \acp{hsi} present in the HySpecNet-11k dataset \cite{fuchs2023hyspecnet}. Contains modified EnMAP data ©DLR [2022].}
    \label{fig:img-hyspecnet-11k}
\end{figure*}

HySpecNet-11k \cite{fuchs2023hyspecnet} is a large-scale hyperspectral benchmark dataset constructed from \num{250} tiles acquired by the \ac{enmap} satellite \cite{guanter2015enmap}.
The dataset comprises \num{11483} nonoverlapping \acp{hsi} with a \ac{gsd} of \SI{30}{\meter} and a spectral range of {\SIrange{420}{2450}{\nano\meter}}.
Each \ac{hsi} consist of \qtyproduct{128x128}{} \si{\pixels} and \SI{202}{\sband}.
The data is radiometrically, geometrically, and atmospherically corrected (i.e., the \acs{l2a} water \& land product).
We used the recommended splits from \cite{fuchs2023hyspecnet} for training, validation, and test sets covering \SI{70}{\percent}, \SI{20}{\percent}, and \SI{10}{\percent} of the \acp{hsi}, respectively.
\autoref{fig:img-hyspecnet-11k} illustrates example images from this dataset.

\subsubsection{MLRetSet}
\begin{figure*}
    % first row
    \begin{minipage}[b]{.093\linewidth}
        \centering
        \includegraphics[width=\linewidth]{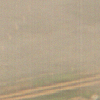}
    \end{minipage}
    \hfill
    \begin{minipage}[b]{.093\linewidth}
        \centering
        \includegraphics[width=\linewidth]{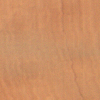}
    \end{minipage}
    \hfill
    \begin{minipage}[b]{.093\linewidth}
        \centering
        \includegraphics[width=\linewidth]{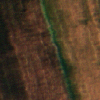}
    \end{minipage}
    \hfill
    \begin{minipage}[b]{.093\linewidth}
        \centering
        \includegraphics[width=\linewidth]{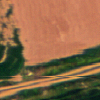}
    \end{minipage}
    \hfill
    \begin{minipage}[b]{.093\linewidth}
        \centering
        \includegraphics[width=\linewidth]{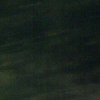}
    \end{minipage}
    \hfill
    \begin{minipage}[b]{.093\linewidth}
        \centering
        \includegraphics[width=\linewidth]{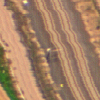}
    \end{minipage}
    \hfill
    \begin{minipage}[b]{.093\linewidth}
        \centering
        \includegraphics[width=\linewidth]{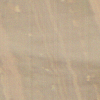}
    \end{minipage}
    \hfill
    \begin{minipage}[b]{.093\linewidth}
        \centering
        \includegraphics[width=\linewidth]{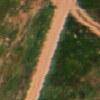}
    \end{minipage}
    \hfill
    \begin{minipage}[b]{.093\linewidth}
        \centering
        \includegraphics[width=\linewidth]{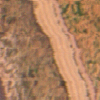}
    \end{minipage}
    \hfill
    \begin{minipage}[b]{.093\linewidth}
        \centering
        \includegraphics[width=\linewidth]{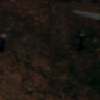}
    \end{minipage}
    % second row
    \begin{minipage}[b]{.093\linewidth}
        \vspace{.07\linewidth}
        \centering
        \includegraphics[width=\linewidth]{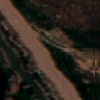}
    \end{minipage}
    \hfill
    \begin{minipage}[b]{.093\linewidth}
        \centering
        \includegraphics[width=\linewidth]{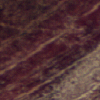}
    \end{minipage}
    \hfill
    \begin{minipage}[b]{.093\linewidth}
        \centering
        \includegraphics[width=\linewidth]{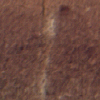}
    \end{minipage}
    \hfill
    \begin{minipage}[b]{.093\linewidth}
        \centering
        \includegraphics[width=\linewidth]{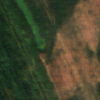}
    \end{minipage}
    \hfill
    \begin{minipage}[b]{.093\linewidth}
        \centering
        \includegraphics[width=\linewidth]{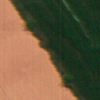}
    \end{minipage}
    \hfill
    \begin{minipage}[b]{.093\linewidth}
        \centering
        \includegraphics[width=\linewidth]{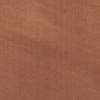}
    \end{minipage}
    \hfill
    \begin{minipage}[b]{.093\linewidth}
        \centering
        \includegraphics[width=\linewidth]{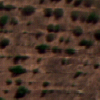}
    \end{minipage}
    \hfill
    \begin{minipage}[b]{.093\linewidth}
        \centering
        \includegraphics[width=\linewidth]{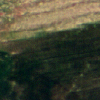}
    \end{minipage}
    \hfill
    \begin{minipage}[b]{.093\linewidth}
        \centering
        \includegraphics[width=\linewidth]{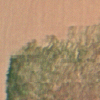}
    \end{minipage}
    \hfill
    \begin{minipage}[b]{.093\linewidth}
        \centering
        \includegraphics[width=\linewidth]{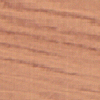}
    \end{minipage}
    % \vspace*{-1em}
    \caption{True color representations of examplary \acp{hsi} present in the MLRetSet dataset \cite{omruuzun2024novel}.}
    \label{fig:img-mlretset}
\end{figure*}

MLRetSet \cite{omruuzun2024novel} is a hyperspectral benchmark dataset with a \ac{gsd} of \SI{27.86}{\centi\meter} derived from twelve tiles acquired during an airborne flight covering the Turkish towns Yenice and Yeşilkaya on 4 May 2019.
The dataset comprises \num{3840} nonoverlapping \acp{hsi} of size \qtyproduct{100x100}{} \si{\pixels} with \SI{369}{\sband} each.
We split the data into
\begin{enumerate*}[i)]
    \item a training set that includes \SI{70}{\percent} of the \acp{hsi}; 
    \item a validation set that includes \SI{20}{\percent} of the \acp{hsi}; and
    \item test set that includes \SI{10}{\percent} of the \acp{hsi}.
\end{enumerate*}
\autoref{fig:img-mlretset} shows exemplary images of this dataset.

\subsection{Experimental Setup}
% Maybe add (i.e. \SIrange{0}{10000}{} in the case of HySpecNet-11k) for rate-discortion loss
Our implementation is based on PyTorch and the CompressAI framework \cite{begaint2020compressai}.
For the HySpecNet-11k dataset, we followed the easy split as introduced in \cite{fuchs2023hyspecnet}.
For the MLRetSet dataset, the \acp{hsi} were center-cropped to \qtyproduct{96x96}{} \si{\pixels} to facilitate successive spatial downsampling by factors of two, consistent with the requirements of spatial and spatio-spectral compression models.
Training runs were carried out on a single NVIDIA A100 \acs{gpu} with \SI{40}{\giga\byte} of \acs{vram}.
Following \cite{minnen2018joint}, we employed two Adam optimizers \cite{kingma2014adam}: a main optimizer for all parameters excluding those of the entropy model, and an \ac{aux} optimizer for the entropy model parameters, with \acp{lr} of \num{1e-4} and \num{1e-3}, respectively.
The \ac{bs} was set to \num{16} and the number of epochs to \num{150} based on empirical observations balancing loss convergence, training time and final model performance.
Kernel sizes $k \in \{ \num{3}, \num{5}, \num{7} \}$ were considered in our experiments to assess the impact of the spatial receptive field on the reconstruction fidelity.
The number of spatial stages $S$ inside \ac{ours} encoder and decoder was varied between \SI{0}{\times} and \SI{4}{\times} to evaluate the effect of different levels of spatial processing.
To account for different transform capacities, we varied $M \in \{ \num{384}, \num{768}, \num[group-separator={}]{1024}, \num[group-separator={}]{1280} \}$.
\hl{As suggested in} \cite{balle2018variational}\hl{, we fixed the hyperlatent channels to $N = \frac{3}{5} M$ to reduce the search space and limit the computational complexity of the hyperprior.}
To obtain different \acp{cr} for each model and create a \ac{rd} curve covering a wide range of \acp{cr}, we varied $\lambda \in \{\num{1e-7}, \num{1e-6}, \num{1e-5}, \num{1e-4}, \num{1e-2}, \num{1}, \num{1e+2}, \num{1e+4}\}$ throughout our experiments.
For entropy coding, we employed the \ac{ans} \cite{duda2015use} within \ac{ae}, \ac{ad}, $\ac{ae}^\mathcal{H}$ and $\ac{ad}^\mathcal{H}$.
An overview of all hyperparameters used in this study is provided in \autoref{tab:hyperparameters}.
\begin{table}
    \centering
    \caption{Selected hyperparameter values used for training.}
    \label{tab:hyperparameters}
    \begin{tabular}{ll}
        \hline
        Hyperparameter & Value \\
        \hline
        Epochs & \num{150} \\
        \acs{bs} & \num{16} \\
        \acs{lr} (main, aux) & (\num{1e-4}, \num{1e-3}) \\
        $\lambda$ & \{\num{1e-7}, \num{1e-6}, \num{1e-5}, \num{1e-4}, \num{1e-2}, \num{1}, \num{1e+2}, \num{1e+4}\} \\
        $S$ & \{\SI{0}{\times}, \SI{1}{\times}, \SI{2}{\times}, \SI{3}{\times}, \SI{4}{\times}\} \\
        ($M$, $N$) & \{(\num{384}, \num{230}), (\num{768}, \num{460}), (\num[group-separator={}]{1024}, \num{614}), (\num[group-separator={}]{1280}, \num{768})\} \\
        $k$ & \{\num{3}, \num{5}, \num{7}\} \\
        \hline
    \end{tabular}
\end{table}

We compare \ac{ours} with two traditional compression approaches, namely JPEG2000 and \ac{pca}.
For JPEG2000, the normalized hyperspectral data is rescaled to the full range of the respective sensor and processed as {\num{16}}-bit unsigned integer.
For {\ac{pca}}, the projected coefficient matrix, the corresponding principal component basis vectors and the mean spectral signature of the {\ac{hsi}} are taken into account to compute the {\ac{cr}}, as they are required for reconstruction.
We would like to note that we do not consider any additional lossless coding strategies, as we employ {\ac{pca}} as a pure dimensionality reduction baseline.
Furthermore, we compare with the following state-of-the-art learning-based {\ac{hsi}} compression models:
\begin{enumerate*}[1)]
    \item \ac{1dcae} \cite{kuester20211d}, a spectral compression model;
    \item \ac{sscnet} \cite{la2022hyperspectral}, a \ac{cae} performing spatial compression;
    \item \ac{3dcae} \cite{chong2021end}, a spatio-spectral compression model;
    \item \ac{hycot} \cite{fuchs2024hycot}, a transformer-based autoencoder exploiting long-range spectral redundancies for pixelwise compression;
    \item \ac{hific}\textsubscript{\acs{se}} \cite{fuchs2024generative}, a \ac{gan}-based spatio-spectral compression model;
    \item \ac{hycass} \cite{fuchs2025adjustable}, an adjustable compression model in both spectral and spatial dimensions;
    \item Verdú et al.  \cite{mijares2023reduced}, a \ac{vae} based on channel clustering;
    \item {\acs{linerwkv}}-L \cite{valsesia2024onboard}, a predictive neural network for lossless and near-lossless {\ac{hsi}} compression;
    \item hyperspectral-VAE \cite{park2025hyperspectral}, a \ac{vae} for joint data compression and component extraction of \acp{hsi};
    \item \ac{msahific} \cite{wan2025msahific}, a super-prior driven high-fidelity spectral attention network;
    \item {\acs{btcnet}} V2 \cite{zhou2026btc}, a model that compress hyperspectral tensor data at the bit-level; and
    \item mean \& scale hyperprior \cite{minnen2018joint}, a variational image compression model that employs spatially adaptive Gaussian entropy modeling.
\end{enumerate*}
All the models were trained from scratch.
Where available, we directly report the evaluation results from the corresponding publication to ensure a fair and consistent comparison.
For the remaining models that do not report results on the considered datasets, we chose different training epochs, \acp{lr}, and \acp{bs} for each model and dataset to balance training time and \acs{gpu} memory usage while ensuring loss convergence in each training run.
The corresponding hyperparameters were determined empirically through preliminary grid search and subsequent analysis of the results.

\subsection{Evaluation Metrics}
For the assessment of the \ac{hsi} compression methods, we employ two kinds of metrics:
\begin{itemize*}[i)]
    \item metrics that quantify compression efficiency (rate); and
    \item metrics that quantify reconstruction fidelity (distortion).
\end{itemize*}
In our experiments we use the \ac{cr} to measure the data reduction, while \ac{psnr}, \ac{bdpsnr}, \ac{sa}, and \ac{ssim} are used to measure the reconstruction quality.
Given an original \ac{hsi} $\vect{X} \in \mathbb{R}^{H \times W \times C}$, its reconstruction $\vect{\hat{X}} \in \mathbb{R}^{H \times W \times C}$, and the compressed bitstream $\vect{b} \in \{ \num{0}, \num{1} \}^L$, the metrics are defined as follows.

\subsubsection{Compression Ratio (CR)}
The compression efficiency is evaluated using the \ac{cr} as a metric.
The \ac{cr} between $\vect{X}$ with bit depth $N_b$ and $\vect{b}$, which includes both latent and hyperlatent, of length $L$ can be expressed as follows:
\begin{align}
    \text{\acs{cr}} \left( \vect{X}, \vect{b} \right) = \frac{N_b \cdot H \cdot W \cdot C}{L} .
\end{align}
A higher \ac{cr} corresponds to stronger compression of the hyperspectral data, which typically results in lower reconstruction fidelity due to the loss of information.

\subsubsection{Peak Signal-to-Noise Ratio (PSNR)}
To evaluate the reconstruction quality, we employ the \ac{psnr} between $\vect{X}$ and $\vect{\hat{X}}$, defined as:
\begin{align}
    \text{\acs{psnr}} \left( \vect{X}, \vect{\hat{X}} \right) = 10 \cdot \log_{10} \left( \frac{\text{MAX}^2}{\text{MSE} \left( \vect{X}, \vect{\hat{X}} \right)} \right),
\end{align}
where $\text{MAX}$ denotes the maximum possible pixel value, and the \ac{mse} is defined as in \autoref{eq:mse}.
The \ac{psnr} is a widely used metric to quantify the quality of a reconstructed image.
A higher \ac{psnr} value indicates improved reconstruction quality, which corresponds to a reduced distortion.

\subsubsection{Bjøntegaard Delta PSNR (BD-PSNR)}
The \ac{bdpsnr} \cite{bjontegaard2001calculation} is a metric commonly employed in the evaluation of video codecs.
In the context of \ac{hsi} compression, it quantifies the average difference in reconstruction quality (as \ac{psnr} in \ac{decibel}) between two \ac{rd} curves at the same \ac{cr}.
The \ac{bdpsnr} between two compression models $A$ and $B$, can be defined as follows:
\begin{equation}
    \begin{aligned}
        &\text{\ac{bdpsnr}} \left( A, B \right) = \\
        &\frac{1}{\text{\ac{cr}}_\text{max}-\text{\ac{cr}}_\text{min}} \int_{\text{\ac{cr}}_\text{min}}^{\text{\ac{cr}}_\text{max}} \text{\ac{psnr}}_A \left( \text{\ac{cr}} \right) - \text{\ac{psnr}}_B \left( \text{\ac{cr}} \right) d\text{\ac{cr}},
    \end{aligned}
    \label{eq:bdpsnr}
\end{equation}
where $\text{\ac{cr}}_\text{min}$ and $\text{\ac{cr}}_\text{max}$ denote the range of integration and the continuous \ac{rd} curves are calculated using the Akima interpolation \cite{herglotz2022beyond}.
Positive \ac{bdpsnr} values indicate that compression model $A$ provides on average a better reconstruction quality than the reference model $B$.

\subsubsection{Spectral Angle (SA)}
To quantify the average spectral distortion between $\vect{X}$ and $\vect{\hat{X}}$, we report the \ac{sa} defined as:
\begin{equation}
\begin{aligned}
    &\text{\acs{sa}} \left( \vect{X}, \vect{\hat{X}} \right) \\
    &= \frac{1}{H \cdot W} \sum_{h, w} \frac{180}{\pi} \arccos \left( \frac{\langle \vect{X} \left( h, w \right), \vect{\hat{X}} \left( h, w \right) \rangle}{\parallel \vect{X} \left( h, w \right)\parallel_2 \cdot \parallel \vect{\hat{X}} \left( h, w \right)\parallel_2} \right),
\end{aligned}
\end{equation}
where $\langle \cdot , \cdot \rangle$ and $\parallel \cdot \parallel_2$ denote the dot product and the Euclidean norm, respectively.
A lower \ac{sa} corresponds to higher spectral similarity and is inherently scale-invariant.

\subsubsection{Structural Similarity Index Measure (SSIM)}
To assess the perceived spatial quality of a reconstruction, we use the \ac{ssim} defined as:
\begin{align}
    \text{\acs{ssim}} \left( \vect{X}, \vect{\hat{X}} \right) = 
    \frac{(2 \cdot \mu_X \mu_{\hat{X}} + c_1) \cdot (2 \cdot \sigma_{X\hat{X}} + c_2)}{(\mu_X^2 + \mu_{\hat{X}}^2 + c_1) \cdot (\sigma_X^2 + \sigma_{\hat{X}}^2 + c_2)},
\end{align}
where $\mu_X$ and $\mu_{\hat{X}}$ are the means, $\sigma_X^2$ and $\sigma_{\hat{X}}^2$ are the variances and $\sigma_{X\hat{X}}$ is the covariance of $\vect{X}$ and $\vect{\hat{X}}$.
$c_1$ and $c_2$ are constants to provide stability against weak denominators.
The \ac{ssim} metric ranges from \num{-1} to \num{1}, with higher values indicating better perceptual reconstruction quality.

\section{Experimental Results}
\label{sec:experiments}
To assess the performance of the proposed \ac{ours} model, we carried out different kinds of experiments in order to:
\begin{enumerate*}[1)]
    \item conduct an ablation study of \ac{ours} while jointly performing metric-driven hyperparameter selection;
    \item compare \ac{ours} with the state of the art in terms of {\ac{rd}} performance and computational complexity; and
    \item analyze the reconstruction results with respect to spatial and spectral reconstruction fidelity, as well as their impact on downstream task performance.
\end{enumerate*}

\subsection{Ablation Study and Metric-Driven Hyperparameter Selection}
\label{subsec:metric-driven}
In this subsection, we analyze the impact of \ac{ours}'s kernel size $k$, spatial stages $S$, latent channels $M$, and hyperlatent channels $N$ on the reconstruction quality across multiple \acp{cr} using the HySpecNet-11k benchmark dataset.
Based on the results, we employ a metric-driven hyperparameter selection strategy using the \ac{bdpsnr} metric to systematically adapt the hyperparameters of \ac{ours} to the characteristics of \acp{hsi}.

\subsubsection{Kernel Size}
\begin{figure}
    \begin{subfigure}{\linewidth}
        \centering
        \begin{tikzpicture}
            \begin{axis}[
                ybar,
                bar width=18pt,
                width=\linewidth,
                height=0.5\linewidth,
                ylabel={\acs{psnr} [\si{\decibel}]},
                legend pos=north west,
                xtick={1,2,3},
                xticklabels={$\lambda=10^{-6}$,$\lambda=10^{-2}$,$\lambda=10^{4}$},
                enlarge x limits=0.3,
                xmajorgrids=false,
                xminorgrids=false,
                xtick style={draw=none},
            ]
                % k = 3
                \addplot[draw=blue,fill=blue!20] coordinates {
                    (1, 41.03213067503341)
                    (2, 50.47092842972653)
                    (3, 50.70979348292334)
                };
                \addlegendentry{$k=\num{3}$}
                % k = 5
                \addplot[draw=red,fill=red!20] coordinates {
                    (1, 40.01595007334852)
                    (2, 49.85057559329043)
                    (3, 49.26801912792884)
                };
                \addlegendentry{$k=\num{5}$}
                % k = 7
                \addplot[draw=teal,fill=teal!20] coordinates {
                    (1, 39.828346563133216)
                    (2, 49.927323793284984)
                    (3, 47.92042775270416)
                };
                \addlegendentry{$k=\num{7}$}
            \end{axis}
        \end{tikzpicture}
        \caption{Reconstruction quality}
        \label{fig:kernelsize-psnr}
    \end{subfigure}
    \hfill
    \begin{subfigure}{0.49\linewidth}
        \centering
        \begin{tikzpicture}
            \begin{axis}[
                ybar,
                bar width=18pt,
                width=\linewidth,
                height=\linewidth,
                ylabel={Parameters [\si{\mega\nothing}]},
                xtick={1},
                xticklabels={},
                enlarge x limits=0.3,
                xmajorgrids=false,
                xminorgrids=false,
                xtick style={draw=none},
                xmin=0.5, xmax=1.5,
            ]
                % k = 3
                \addplot[draw=blue,fill=blue!20] coordinates {
                    (1, 26892682/1e6)
                };
                % k = 5
                \addplot[draw=red,fill=red!20] coordinates {
                    (1, 83996554/1e6)
                };
                % k = 7
                \addplot[draw=teal,fill=teal!20] coordinates {
                    (1, 178143370/1e6)
                };
            \end{axis}
        \end{tikzpicture}
        \caption{Memory usage}
        \label{fig:kernelsize-params}
    \end{subfigure}
    \hfill
    \begin{subfigure}{0.49\linewidth}
        \centering
        \begin{tikzpicture}
            \begin{axis}[
                ybar,
                bar width=18pt,
                width=\linewidth,
                height=\linewidth,
                ylabel={\acs{flops} [\hl{{\si{\giga\flops}}}]},
                xtick={1},
                xticklabels={},
                enlarge x limits=0.3,
                xmajorgrids=false,
                xminorgrids=false,
                xtick style={draw=none},
                xmin=0.5, xmax=1.5,
            ]
            % k = 3
            \addplot[draw=blue,fill=blue!20] coordinates {
                (1, 128598394368/1e9)
            };
            % k = 5
            \addplot[draw=red,fill=red!20] coordinates {
                (1, 370772166144/1e9)
            };
            % k = 7
            \addplot[draw=teal,fill=teal!20] coordinates {
                (1, 751422408192/1e9)
            };
            \end{axis}
        \end{tikzpicture}
        \caption{Computational complexity}
        \label{fig:kernelsize-flops}
    \end{subfigure}
    \caption{Ablation study of kernel size $k$ on (\subref{fig:kernelsize-psnr}) reconstruction quality evaluated as \ac{psnr}, (\subref{fig:kernelsize-params}) memory usage in terms of parameter count, and (\subref{fig:kernelsize-flops}) computational complexity measured as \ac{flops}.
    Results are reported on the HySpecNet-11k \cite{fuchs2023hyspecnet} validation set (easy split) for three $\lambda$ values, while fixing $S = \num{2}$, $M = \num{768}$, and $N = \num{460}$.}
    \label{fig:kernelsize-ablation}
\end{figure}
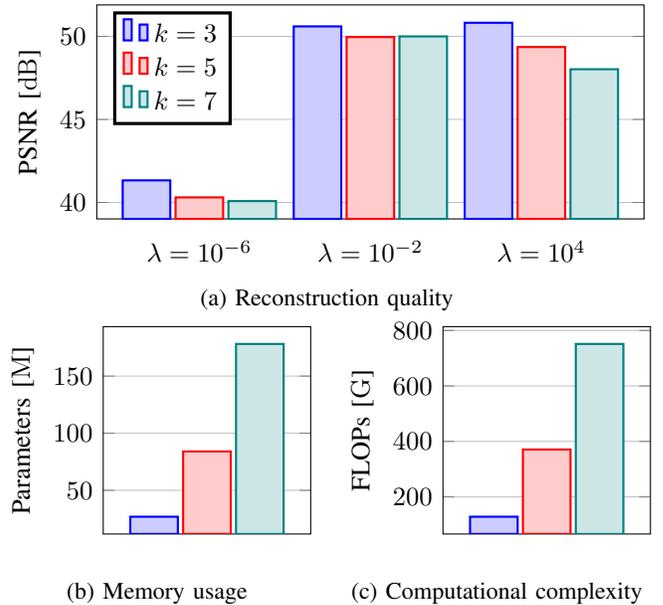

The size $k$ of the convolutional kernels determines the extent of the spatial receptive field within the four components of \ac{ours}.
The effect of different values of $k \in \{ \num{3}, \num{5}, \num{7} \}$ on the reconstruction quality, memory usage and computational complexity is shown in \autoref{fig:kernelsize-ablation}.
To this end, the remaining hyperparameters of \ac{ours} are fixed to $S = \num{2}$, $M = \num{768}$, and $N = \num{460}$, as this configuration provides a reasonable trade-off between spatial and spectral feature learning.
As shown in \autoref{fig:kernelsize-ablation} (\subref{fig:kernelsize-psnr}), a small receptive field of $k = \num{3}$ consistently achieves the best reconstruction quality across all three $\lambda$ values, where a higher $\lambda$ indicates a lower \ac{cr}.
In particular, $k = \num{3}$ yields \ac{psnr} improvements of \SI[round-mode=places,round-precision=2]{1.016180602}{\decibel} and \SI[round-mode=places,round-precision=2]{1.203784112}{\decibel} for $\lambda = \num{1e-6}$, \SI[round-mode=places,round-precision=2]{0.620352836}{\decibel} and \SI[round-mode=places,round-precision=2]{0.543604636}{\decibel} for $\lambda = \num{1e-2}$, and \SI[round-mode=places,round-precision=2]{1.441774355}{\decibel} and \SI[round-mode=places,round-precision=2]{2.78936573}{\decibel} for $\lambda = \num{1e+4}$, compared to $k = \num{5}$ and $k = \num{7}$, respectively.
On average, these gains correspond to relative \ac{psnr} improvements of \SI[round-mode=places,round-precision=2]{2.1736178}{\percent} and \SI[round-mode=places,round-precision=2]{3.3055511}{\percent}, respectively.
These results suggest that the most relevant spatial redundancies are predominantly concentrated in the immediate neighborhood, thereby limiting the benefit of modeling broader spatial context.
This behavior can largely be attributed to the low spatial resolution of \ac{rs} \acp{hsi}.

Moreover, \autoref{fig:kernelsize-ablation} (\subref{fig:kernelsize-params}) and (\subref{fig:kernelsize-flops}) demonstrate that using $k = \num{3}$ substantially reduces both model size and computational complexity, as smaller kernels inherently require fewer parameters and \ac{flops}.
Specifically, using $k = \num{3}$ reduces the number of parameters by \SI{67.96}{\percent} and \SI{84.91}{\percent} and the \ac{flops} by \SI{65.3}{\percent} and \SI{82.9}{\percent} compared to $k = \num{5}$ and $k = \num{7}$, respectively.
We would like to note that the number of parameters and \ac{flops} is determined solely by the model architecture and are therefore independent of $\lambda$.

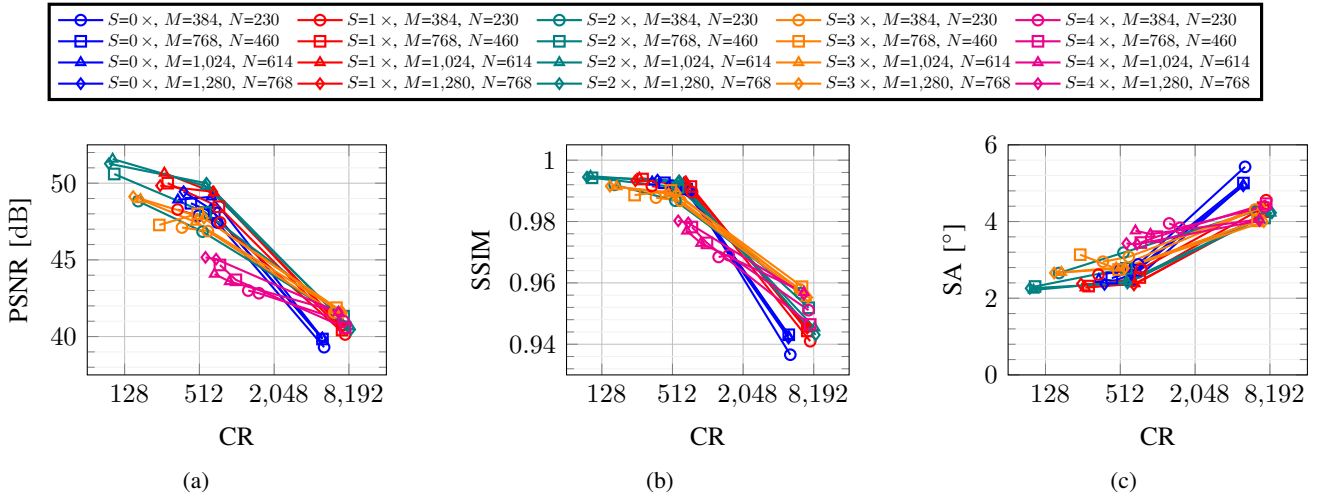
\begin{figure}
    \centering
    \begin{subfigure}{\linewidth}
        \centering
        \ref{sharedlegend-ablation}
        \vspace{1em}
    \end{subfigure}
    \vspace{0.8em}
    \begin{subfigure}{\linewidth}
        \centering
        \begin{tikzpicture}
            \begin{axis}[
                width=\linewidth-6.9573pt,
                height=0.65\linewidth,
                % transpose legend,
                legend cell align={left},
                xmode=log,
                log basis x={2},
                log ticks with fixed point,
                minor tick num=4,
                ymin=37.5, ymax=52.5,
                xmin=64, xmax=16384,
                legend style={nodes={scale=0.8, transform shape}},
                xticklabel={
                    \pgfkeys{/pgf/fpu=true}
                    \pgfmathparse{int(2^\tick)}
                    \pgfmathprintnumber[fixed]{\pgfmathresult}
                },
                tick label style={font=\footnotesize},
                xlabel={\acs{cr}},
                ylabel={\acs{psnr} [\si{\decibel}]},
                % legend cell align={left},
                legend pos=outer north east,
                legend columns=2,
                % transpose legend,
                legend to name=sharedlegend-ablation,
                ]
                % S=0, M=384, N=230
                \addplot[blue,mark=o,dashed,mark options={solid}] coordinates {
                    (5069.0626,39.0698)
                    (693.6365,47.3042)
                    (487.2982,47.7577)
                };
                \addlegendentry{$S$=\SI{0}{\times}, $M$=\num{384}, $N$=\num{230}}
                
                % S=0, M=768, N=460
                \addplot[blue,mark=square,dotted,mark options={solid}] coordinates {
                    (4920.2519,39.5666)
                    (656.4351,47.9247)
                    (426.7519,48.5600)
                };
                \addlegendentry{$S$=\SI{0}{\times}, $M$=\num{768}, $N$=\num{460}}
                
                % S=0, M=1024, N=614
                \addplot[blue,mark=triangle,dashdotted,mark options={solid}] coordinates {
                    (4908.9520,39.6436)
                    (623.1809,48.9933)
                    (338.1303,48.8028)
                };
                \addlegendentry{$S$=\SI{0}{\times}, $M$=\num{1024}, $N$=\num{614}}
                
                % S=0, M=1280, N=768
                \addplot[blue,mark=diamond,dashdotdotted,mark options={solid}] coordinates {
                    (4901.2927,39.6391)
                    (617.3513,48.1688)
                    (372.0025,49.3406)
                };
                \addlegendentry{$S$=\SI{0}{\times}, $M$=\num{1280}, $N$=\num{768}}

                % S=1, M=384, N=230
                \addplot[red,mark=o,dashed,mark options={solid}] coordinates {
                    (7480.0449,39.8439)
                    (731.0297,47.2628)
                    (328.1401,48.1338)
                };
                \addlegendentry{$S$=\SI{1}{\times}, $M$=\num{384}, $N$=\num{230}}
                
                % S=1, M=768, N=460
                \addplot[red,mark=square,dotted,mark options={solid}] coordinates {
                    (7072.3254,40.1363)
                    (719.1089,48.2952)
                    (276.0017,49.8727)
                };
                \addlegendentry{$S$=\SI{1}{\times}, $M$=\num{768}, $N$=\num{460}}
                
                % S=1, M=1024, N=614
                \addplot[red,mark=triangle,dashdotted,mark options={solid}] coordinates {
                    (7249.2181,40.3128)
                    (657.0737,49.2427)
                    (259.8498,50.5611)
                };
                \addlegendentry{$S$=\SI{1}{\times}, $M$=\num{1024}, $N$=\num{614}}
                
                % S=1, M=1280, N=768
                \addplot[red,mark=diamond,dashdotdotted,mark options={solid}] coordinates {
                    (7061.1662,40.3128)
                    (645.8364,49.2906)
                    (240.3391,49.7471)
                };
                \addlegendentry{$S$=\SI{1}{\times}, $M$=\num{1280}, $N$=\num{768}}

                % S=2, M=384, N=230
                \addplot[teal,mark=o,dashed,mark options={solid}] coordinates {
                    (6137.4479,41.2811)
                    (523.7247,46.7370)
                    (160.0461,48.6650)
                };
                \addlegendentry{$S$=\SI{2}{\times}, $M$=\num{384}, $N$=\num{230}}
                
                % S=2, M=768, N=460
                \addplot[teal,mark=square,dotted,mark options={solid}] coordinates {
                    (7265.3052,41.0321)
                    (557.9463,47.9644)
                    (102.5452,50.4709)
                };
                \addlegendentry{$S$=\SI{2}{\times}, $M$=\num{768}, $N$=\num{460}}
                
                % S=2, M=1024, N=614
                \addplot[teal,mark=triangle,dashdotted,mark options={solid}] coordinates {
                    (8173.1088,40.3436)
                    (561.7357,49.6773)
                    (99.6044,51.5032)
                };
                \addlegendentry{$S$=\SI{2}{\times}, $M$=\num{1024}, $N$=\num{614}}
                
                % S=2, M=1280, N=768
                \addplot[teal,mark=diamond,dashdotdotted,mark options={solid}] coordinates {
                    (8339.2978,40.1744)
                    (571.1264,49.8788)
                    (92.7424,51.2071)
                };
                \addlegendentry{$S$=\SI{2}{\times}, $M$=\num{1280}, $N$=\num{768}}

                % S=3, M=384, N=230
                \addplot[orange,mark=o,dashed,mark options={solid}] coordinates {
                    (6079.5057,41.3202)
                    (585.2929,46.6542)
                    (362.6691,46.8717)
                };
                \addlegendentry{$S$=\SI{3}{\times}, $M$=\num{384}, $N$=\num{230}}
                
                % S=3, M=768, N=460
                \addplot[orange,mark=square,dotted,mark options={solid}] coordinates {
                    (6354.3192,41.6006)
                    (482.3139,47.8169)
                    (238.1362,47.1060)
                };
                \addlegendentry{$S$=\SI{3}{\times}, $M$=\num{768}, $N$=\num{460}}
                
                % S=3, M=1024, N=614
                \addplot[orange,mark=triangle,dashdotted,mark options={solid}] coordinates {
                    (6897.6496,41.3724)
                    (460.5879,47.2773)
                    (168.0590,48.7684)
                };
                \addlegendentry{$S$=\SI{3}{\times}, $M$=\num{1024}, $N$=\num{614}}
                
                % S=3, M=1280, N=768
                \addplot[orange,mark=diamond,dashdotdotted,mark options={solid}] coordinates {
                    (7163.9588,41.2311)
                    (539.5963,47.5700)
                    (146.9179,48.9653)
                };
                \addlegendentry{$S$=\SI{3}{\times}, $M$=\num{1280}, $N$=\num{768}}
                
                % S=4, M=384, N=230
                \addplot[magenta,mark=o,dashed,mark options={solid}] coordinates {
                    (7212.6326,40.7323)
                    (1514.7326,42.5239)
                    (1244.8809,42.7042)
                };
                \addlegendentry{$S$=\SI{4}{\times}, $M$=\num{384}, $N$=\num{230}}
                
                % S=4, M=768, N=460
                \addplot[magenta,mark=square,dotted,mark options={solid}] coordinates {
                    (7461.4989,40.2579)
                    (986.1798,43.3844)
                    (734.7848,44.3922)
                };
                \addlegendentry{$S$=\SI{4}{\times}, $M$=\num{768}, $N$=\num{460}}
                
                % S=4, M=1024, N=614
                \addplot[magenta,mark=triangle,dashdotted,mark options={solid}] coordinates {
                    (6578.2170,41.2801)
                    (876.5989,43.2863)
                    (664.1996,43.8014)
                };
                \addlegendentry{$S$=\SI{4}{\times}, $M$=\num{1024}, $N$=\num{614}}
                
                % S=4, M=1280, N=768
                \addplot[magenta,mark=diamond,dashdotdotted,mark options={solid}] coordinates {
                    (6574.0738,41.2728)
                    (683.2421,44.7985)
                    (562.3618,44.9242)
                };
                \addlegendentry{$S$=\SI{4}{\times}, $M$=\num{1280}, $N$=\num{768}}
            \end{axis}
        \end{tikzpicture}
        \caption{}
        \label{fig:ablation-smn-psnr}
    \end{subfigure}
    \vspace{0.8em}
    \begin{subfigure}{\linewidth}
        \centering
        \begin{tikzpicture}
            \begin{axis}[
                width=\linewidth-13.19052pt ,
                height=0.65\linewidth,
                legend columns=2,
                % transpose legend,
                legend cell align={left},
                xmode=log,
                log basis x={2},
                log ticks with fixed point,
                minor tick num=3,
                ymin=0.93, ymax=1.004,
                xmin=64, xmax=16384,
                legend style={nodes={scale=0.65, transform shape}},
                xticklabel={
                    \pgfkeys{/pgf/fpu=true}
                    \pgfmathparse{int(2^\tick)}
                    \pgfmathprintnumber[fixed]{\pgfmathresult}
                },
                tick label style={font=\footnotesize},
                xlabel={\acs{cr}},
                ylabel={\acs{ssim}},
                % legend cell align={left},
                legend pos=outer north east,
                ]
                % S=0, M=384, N=230
                \addplot[blue,mark=o,dashed,mark options={solid}] coordinates {
                    (5069.0626,0.9357874)
                    (693.6365,0.9901056)
                    (487.2982,0.9916907)
                };
                
                % S=0, M=768, N=460
                \addplot[blue,mark=square,dotted,mark options={solid}] coordinates {
                    (4920.2519,0.9418717)
                    (656.4351,0.9906740)
                    (426.7519,0.9926618)
                };
                
                % S=0, M=1024, N=614
                \addplot[blue,mark=triangle,dashdotted,mark options={solid}] coordinates {
                    (4908.9520,0.9415455)
                    (623.1809,0.9924710)
                    (338.1303,0.9930108)
                };
                
                % S=0, M=1280, N=768
                \addplot[blue,mark=diamond,dashdotdotted,mark options={solid}] coordinates {
                    (4901.2927,0.9408700)
                    (617.3513,0.9913704)
                    (372.0025,0.9935535)
                };

                % S=1, M=384, N=230
                \addplot[red,mark=o,dashed,mark options={solid}] coordinates {
                    (7480.0449,0.9396253)
                    (731.0297,0.9895359)
                    (328.1401,0.9915661)
                };
                
                % S=1, M=768, N=460
                \addplot[red,mark=square,dotted,mark options={solid}] coordinates {
                    (7072.3254,0.9431318)
                    (719.1089,0.9914268)
                    (276.0017,0.9938769)
                };
                
                % S=1, M=1024, N=614
                \addplot[red,mark=triangle,dashdotted,mark options={solid}] coordinates {
                    (7249.2181,0.9440687)
                    (657.0737,0.9928362)
                    (259.8498,0.9943135)
                };
                
                % S=1, M=1280, N=768
                \addplot[red,mark=diamond,dashdotdotted,mark options={solid}] coordinates {
                    (7061.1662,0.9441970)
                    (645.8364,0.9929688)
                    (240.3391,0.9935169)
                };

                % S=2, M=384, N=230
                \addplot[teal,mark=o,dashed,mark options={solid}] coordinates {
                    (6137.4479,0.9559949)
                    (523.7247,0.9871780)
                    (160.0461,0.9919621)
                };
                
                % S=2, M=768, N=460
                \addplot[teal,mark=square,dotted,mark options={solid}] coordinates {
                    (7265.3052,0.9509103)
                    (557.9463,0.9912691)
                    (102.5452,0.9943028)
                };
                
                % S=2, M=1024, N=614
                \addplot[teal,mark=triangle,dashdotted,mark options={solid}] coordinates {
                    (8173.1088,0.9439737)
                    (561.7357,0.9930555)
                    (99.6044,0.9950053)
                };
                
                % S=2, M=1280, N=768
                \addplot[teal,mark=diamond,dashdotdotted,mark options={solid}] coordinates {
                    (8339.2978,0.9418099)
                    (571.1264,0.9934489)
                    (92.7424,0.9946725)
                };

                % S=3, M=384, N=230
                \addplot[orange,mark=o,dashed,mark options={solid}] coordinates {
                    (6079.5057,0.9575126)
                    (585.2929,0.9866962)
                    (362.6691,0.9876476)
                };
                
                % S=3, M=768, N=460
                \addplot[orange,mark=square,dotted,mark options={solid}] coordinates {
                    (6354.3192,0.9579710)
                    (482.3139,0.9901226)
                    (238.1362,0.9889348)
                };
                
                % S=3, M=1024, N=614
                \addplot[orange,mark=triangle,dashdotted,mark options={solid}] coordinates {
                    (6897.6496,0.9549649)
                    (460.5879,0.9889749)
                    (168.0590,0.9914934)
                };
                
                % S=3, M=1280, N=768
                \addplot[orange,mark=diamond,dashdotdotted,mark options={solid}] coordinates {
                    (7163.9588,0.9540130)
                    (539.5963,0.9893712)
                    (146.9179,0.9915745)
                };
                
                % S=4, M=384, N=230
                \addplot[magenta,mark=o,dashed,mark options={solid}] coordinates {
                    (7212.6326,0.9502539)
                    (1514.7326,0.9681765)
                    (1244.8809,0.9679185)
                };
                
                % S=4, M=768, N=460
                \addplot[magenta,mark=square,dotted,mark options={solid}] coordinates {
                    (7461.4989,0.9448807)
                    (986.1798,0.9723718)
                    (734.7848,0.9778269)
                };
                
                % S=4, M=1024, N=614
                \addplot[magenta,mark=triangle,dashdotted,mark options={solid}] coordinates {
                    (6578.2170,0.9559135)
                    (876.5989,0.9722810)
                    (664.1996,0.9766591)
                };
                
                % S=4, M=1280, N=768
                \addplot[magenta,mark=diamond,dashdotdotted,mark options={solid}] coordinates {
                    (6574.0738,0.9553741)
                    (683.2421,0.9793961)
                    (562.3618,0.9802225)
                };
            \end{axis}
        \end{tikzpicture}
        \caption{}
        \label{fig:ablation-smn-ssim}
    \end{subfigure}
    \begin{subfigure}{\linewidth}
        \centering
        \begin{tikzpicture}
            \begin{axis}[
                width=\linewidth-1.89871pt,
                height=0.65\linewidth,
                legend columns=2,
                % transpose legend,
                legend cell align={left},
                xmode=log,
                log basis x={2},
                log ticks with fixed point,
                minor tick num=3,
                ymin=0.6, ymax=5.8,
                xmin=64, xmax=16384,
                legend style={nodes={scale=0.65, transform shape}},
                xticklabel={
                    \pgfkeys{/pgf/fpu=true}
                    \pgfmathparse{int(2^\tick)}
                    \pgfmathprintnumber[fixed]{\pgfmathresult}
                },
                tick label style={font=\footnotesize},
                xlabel={\acs{cr}},
                ylabel={\acs{sa} [\si{\degree}]},
                ytick={0,1,2,3,4,5,6},
                yticklabels={0,1,2,3,4,5,6},
                % legend cell align={left},
                legend pos=outer north east,
                ]
                % S=0, M=384, N=230
                \addplot[blue,mark=o,dashed,mark options={solid}] coordinates {
                    (5069.0626,5.2546)
                    (693.6365,2.7877)
                    (487.2982,2.6444)
                };
                
                % S=0, M=768, N=460
                \addplot[blue,mark=square,dotted,mark options={solid}] coordinates {
                    (4920.2519,4.9243)
                    (656.4351,2.6312)
                    (426.7519,2.4410)
                };
                
                % S=0, M=1024, N=614
                \addplot[blue,mark=triangle,dashdotted,mark options={solid}] coordinates {
                    (4908.9520,4.8624)
                    (623.1809,2.3378)
                    (338.1303,2.3969)
                };
                
                % S=0, M=1280, N=768
                \addplot[blue,mark=diamond,dashdotdotted,mark options={solid}] coordinates {
                    (4901.2927,4.8469)
                    (617.3513,2.5277)
                    (372.0025,2.2707)
                };
                
                % S=1, M=384, N=230
                \addplot[red,mark=o,dashed,mark options={solid}] coordinates {
                    (7480.0449,4.4695)
                    (731.0297,2.6726)
                    (328.1401,2.5225)
                };
                
                % S=1, M=768, N=460
                \addplot[red,mark=square,dotted,mark options={solid}] coordinates {
                    (7072.3254,4.2722)
                    (719.1089,2.4515)
                    (276.0017,2.2338)
                };
                
                % S=1, M=1024, N=614
                \addplot[red,mark=triangle,dashdotted,mark options={solid}] coordinates {
                    (7249.2181,4.1263)
                    (657.0737,2.2857)
                    (259.8498,2.1693)
                };
                
                % S=1, M=1280, N=768
                \addplot[red,mark=diamond,dashdotdotted,mark options={solid}] coordinates {
                    (7061.1662,4.1289)
                    (645.8364,2.2604)
                    (240.3391,2.2592)
                };
                
                % S=2, M=384, N=230
                \addplot[teal,mark=o,dashed,mark options={solid}] coordinates {
                    (6137.4479,4.1799)
                    (523.7247,3.0400)
                    (160.0461,2.5683)
                };
                
                % S=2, M=768, N=460
                \addplot[teal,mark=square,dotted,mark options={solid}] coordinates {
                    (7265.3052,4.0113)
                    (557.9463,2.6704)
                    (102.5452,2.2156)
                };
                
                % S=2, M=1024, N=614
                \addplot[teal,mark=triangle,dashdotted,mark options={solid}] coordinates {
                    (8173.1088,4.1490)
                    (561.7357,2.3333)
                    (99.6044,2.0975)
                };
                
                % S=2, M=1280, N=768
                \addplot[teal,mark=diamond,dashdotdotted,mark options={solid}] coordinates {
                    (8339.2978,4.1407)
                    (571.1264,2.2918)
                    (92.7424,2.1484)
                };
                
                % S=3, M=384, N=230
                \addplot[orange,mark=o,dashed,mark options={solid}] coordinates {
                    (6079.5057,4.1769)
                    (585.2929,2.9897)
                    (362.6691,2.8679)
                };
                
                % S=3, M=768, N=460
                \addplot[orange,mark=square,dotted,mark options={solid}] coordinates {
                    (6354.3192,3.9196)
                    (482.3139,2.6630)
                    (238.1362,3.0016)
                };
                
                % S=3, M=1024, N=614
                \addplot[orange,mark=triangle,dashdotted,mark options={solid}] coordinates {
                    (6897.6496,4.2011)
                    (460.5879,2.7065)
                    (168.0590,2.5725)
                };
                
                % S=3, M=1280, N=768
                \addplot[orange,mark=diamond,dashdotdotted,mark options={solid}] coordinates {
                    (7163.9588,3.9297)
                    (539.5963,2.6815)
                    (146.9179,2.5563)
                };
                
                % S=4, M=384, N=230
                \addplot[magenta,mark=o,dashed,mark options={solid}] coordinates {
                    (7212.6326,4.2666)
                    (1514.7326,3.7567)
                    (1244.8809,3.8422)
                };
                
                % S=4, M=768, N=460
                \addplot[magenta,mark=square,dotted,mark options={solid}] coordinates {
                    (7461.4989,4.3831)
                    (986.1798,3.5324)
                    (734.7848,3.3610)
                };
                
                % S=4, M=1024, N=614
                \addplot[magenta,mark=triangle,dashdotted,mark options={solid}] coordinates {
                    (6578.2170,3.9795)
                    (876.5989,3.5964)
                    (664.1996,3.6780)
                };
                
                % S=4, M=1280, N=768
                \addplot[magenta,mark=diamond,dashdotdotted,mark options={solid}] coordinates {
                    (6574.0738,3.9514)
                    (683.2421,3.2955)
                    (562.3618,3.3091)
                };
            \end{axis}
        \end{tikzpicture}
        \caption{}
        \label{fig:ablation-smn-sa}
    \end{subfigure}
    \caption{Ablation study for spatial stages $S$, latent channels $M$, and hyperlatent channels $N$ on the \ac{rd} performance for the HySpecNet-11k \cite{fuchs2023hyspecnet} validataion set (easy split). Rate is visualized as \ac{cr} in a logarithmic scale and distortion is given as (\subref{fig:ablation-smn-psnr}) \ac{psnr}, (\subref{fig:ablation-smn-ssim}) \ac{ssim}, and (\subref{fig:ablation-smn-sa}) \ac{sa}.}
    \label{fig:ablation-smn}
\end{figure}

Overall, the results demonstrate that, despite the increased representational capacity of larger convolutional kernels, a kernel size of $k=\num{3}$ consistently achieves higher reconstruction fidelity while substantially reducing both model size and computational complexity.
Therefore, $k=\num{3}$ is selected for all subsequent experiments.
It is noted that the findings regarding the kernel size were also confirmed through experiments on the MLRetSet dataset (not reported for space constraints).

\subsubsection{Spatial Stages, Latent and Hyperlatent Channels}
\autoref{fig:ablation-smn} presents the ablation study of \ac{ours} with respect to the spatial stages $S$, latent channels $M$, and hyperlatent channels $N$.
To evaluate both spatial and spectral reconstruction fidelity, the \ac{rd} curves are visualized using (\subref{fig:ablation-smn-psnr}) \ac{psnr}, (\subref{fig:ablation-smn-ssim}) \ac{ssim}, and (\subref{fig:ablation-smn-sa}) \ac{sa} as distortion metrics, with the rate expressed as \ac{cr} on a logarithmic scale.

By analyzing the \ac{rd} curves associated to different values of $S$ in \autoref{fig:ablation-smn} (\subref{fig:ablation-smn-psnr}), one can observe that the choice of $S$ has a significant impact on the reconstruction fidelity with its effect varying depending on the \ac{cr}.
As an example, in the case of zero spatial stages ($S = \SI{0}{\times}$), where \ac{ours} relies solely on spectral feature learning for the latent representation, the reconstruction quality decreases notably at higher \acp{cr}.
This behavior can be attributed to the fact that spatial redundancies are not exploited with \SI{0}{\times} spatial stages, although they become increasingly important under stronger compression constraints \cite{fuchs2025adjustable}.
In addition, the hyperencoder benefits from spatial structure in the latent representation to more effectively model the underlying probability distribution.
In contrast, configurations with \SI{3}{\times} and \SI{4}{\times} spatial stages exhibit reduced reconstruction fidelity at lower \acp{cr}, where preserving spectral information is more crucial \cite{fuchs2025adjustable}.
In this \ac{cr} regime, excessive spatial downsampling limits the model’s ability to accurately reconstruct spectral content.
% Among the evaluated configurations, $S = \SI{2}{\times}$ consistently achieves the highest reconstruction quality across the entire \ac{rd} curve, indicating a balanced exploitation of spatial redundancies.

As mentioned in \cite{balle2018variational}, lower \acp{cr} require higher $M$ and $N$, but beyond a certain point, further increasing the transform capacity results in only marginal improvements.
This behavior is also evident in \autoref{fig:ablation-smn} (\subref{fig:ablation-smn-psnr}), where increasing $M$ and $N$ improves the reconstruction quality. % as indicated by different markers of the same color.
Given the high number of spectral bands in \acp{hsi}, higher $M$ and $N$ are required compared to natural images.
However, the performance saturates relatively early, suggesting that the latent and hyperlatent dimensions do not increase proportionally with the number of input channels.
% Based on these observations, $M=\num{1260}$ and $N=\num{768}$ yield the best reconstruction fidelity.

The trends observed for \ac{psnr} in \autoref{fig:ablation-smn} (\subref{fig:ablation-smn-psnr}) are consistent across other distortion metrics, as shown in (\subref{fig:ablation-smn-ssim}) for \ac{ssim} and (\subref{fig:ablation-smn-sa}) for \ac{sa}.
Overall, the results demonstrate that the proposed model achieves a balanced spatio-spectral compression, effectively preserving both spatial structure and spectral coherence.

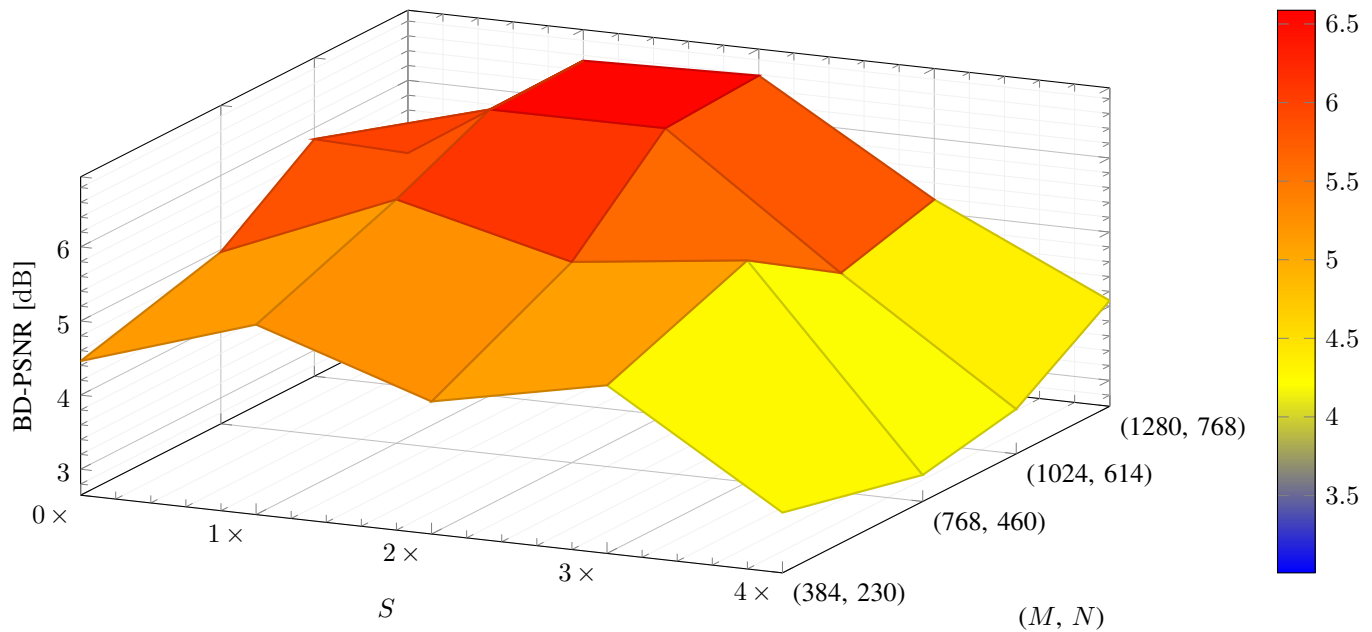
\begin{figure*}
    \centering
    \begin{tikzpicture}
        \begin{axis}[
            colorbar,
            width=0.85\linewidth-4.30716pt,
            height=0.50\linewidth,
            xlabel={$S$},
            ylabel={($M$, $N$)},
            zlabel={\acs{bdpsnr} [\si{\decibel}]},
            legend columns=2,
            % transpose legend,
            legend cell align={left},
            legend pos=north east,
            minor tick num=4,
            xtick={0,1,2,3,4},
            xticklabels={\SI{0}{\times}, \SI{1}{\times}, \SI{2}{\times}, \SI{3}{\times}, \SI{4}{\times}},
            xticklabel style={anchor=north east},
            ytick={384,768,1024,1280},
            yticklabels={(384, 230),(768, 460),(1024, 614),(1280, 768)},
            yticklabel style={anchor=north west}
            % legend style={nodes={scale=0.5, transform shape}},
            % view={45}{80},
        ]
            \addplot3[surf] coordinates {
                (0, 384, 4.236471150234325)        % ours_s0_n230_m384_k3-lmbd
                (0, 768, 4.739255450025691)        % ours_s0_n460_m768_k3-lmbd
                (0, 1024, 5.637504214260625)       % ours_s0_n614_m1024_k3-lmbd
                (0, 1280, 4.793927044361848)       % ours_s0_n768_m1280_k3-lmbd
                
                (1, 384, 4.947310917213766)        % ours_s1_n230_m384_k3-lmbd
                (1, 768, 5.694716406458971)        % ours_s1_n460_m768_k3-lmbd
                (1, 1024, 6.281215211342802)       % ours_s1_n614_m1024_k3-lmbd
                (1, 1280, 6.296302985067629)       % ours_s1_n768_m1280_k3-lmbd
                
                (2, 384, 4.220406813965368)        % ours_s2_n230_m384_k3-lmbd
                (2, 768, 5.139241190105107)        % ours_s2_n460_m768_k3-lmbd
                (2, 1024, 6.301283647039186)       % ours_s2_n614_m1024_k3-lmbd
                (2, 1280, 6.384173185770688)       % ours_s2_n768_m1280_k3-lmbd
                
                (3, 384, 4.642174095183255)        % ours_s3_n230_m384_k3-lmbd
                (3, 768, 5.3663252197074724)       % ours_s3_n460_m768_k3-lmbd
                (3, 1024, 4.560190597773558)       % ours_s3_n614_m1024_k3-lmbd
                (3, 1280, 4.916029846519774)       % ours_s3_n768_m1280_k3-lmbd
                
                (4, 384, 3.1195513188595414)       % ours_s4_n230_m384_k3-lmbd
                (4, 768, 2.6261639742337515)       % ours_s4_n460_m768_k3-lmbd
                (4, 1024, 2.8866707184682063)      % ours_s4_n614_m1024_k3-lmbd
                (4, 1280, 3.746235746696868)       % ours_s4_n768_m1280_k3-lmbd
            };
        \end{axis}
    \end{tikzpicture}
    % \caption{Ablation study for $S$, $M$ \& $N$ evaluated via \ac{bdpsnr} with \ac{hycass} \cite{fuchs2025adjustable} as reference model on the test set of HySpecNet-11k \cite{fuchs2023hyspecnet} (easy split).}
    \caption{\ac{bdpsnr} results, with \ac{hycass} \cite{fuchs2025adjustable} as reference model, for different combinations of $S$, $M$, and $N$ on the validation set of HySpecNet-11k \cite{fuchs2023hyspecnet} (easy split).}
    \label{fig:bd-psnr-ablation}
\end{figure*}

To enable a justified selection of \ac{ours}'s hyperparameters, we employ the \ac{bdpsnr} metric to reduce each \ac{rd} curve from \autoref{fig:ablation-smn} (\subref{fig:ablation-smn-psnr}) into a single scalar value that summarizes the overall compression performance.
As state-of-the-art reference model (see \autoref{eq:bdpsnr}),  we use the \ac{hycass} model as it represents the best-performing model prior to this work.
\autoref{fig:bd-psnr-ablation} shows the resulting \ac{bdpsnr} values for the \num{20} evaluated hyperparameter combinations, with each point in the \ac{3d} plot representing a specific combination of $S$, $M$, and $N$.
Saturation is observed within the considered values of $M$ and $N$, indicating that further increases in latent or hyperlatent channels would provide negligible gains \cite{balle2018variational}.
Considering $(\num[group-separator={}]{1280}, \num{768})$ as the best configuration for $(M, N)$, \ac{bdpsnr} plateaus for \SI{2}{\times} spatial stages, while \SI{1}{\times} achieves comparable reconstruction quality but incurs higher computational complexity due to larger spatial dimensions remaining in the latent representation.
Based on this analysis, the hyperparameters are fixed to $S = \SI{2}{\times}$, $M = \num{1280}$, and $N = \num{768}$ for the subsequent experiments.
We would like to note that the same settings were also applied on the MLRetSet dataset, demonstrating the generalization capability of the proposed model and the robustness of the selected hyperparameters, as shown in the following subsection.

\subsection{Comparison with Other Approaches}
In this subsection, we analyze the effectiveness of \ac{ours} by comparing its \ac{rd} curve with several traditional baselines and state-of-the-art learning-based \ac{hsi} compression models on the HySpecNet-11k and the MLRetSet benchmark datasets.
Furthermore, we compare the computational complexity of the proposed model with the state of the art.
The models considered for comparison include:
\begin{enumerate*}[1)]
    \item JPEG2000;
    \item \ac{pca};
    \item \ac{1dcae} \cite{kuester20211d};
    \item \ac{sscnet} \cite{la2022hyperspectral};
    \item \ac{3dcae} \cite{chong2021end};
    \item \ac{hycot} \cite{fuchs2024hycot};
    \item \ac{hific}\textsubscript{\acs{se}} \cite{fuchs2024generative};
    \item \ac{hycass} \cite{fuchs2025adjustable};
    \item Verdú et al. \cite{mijares2023reduced};
    \item {\acs{linerwkv}}-L \cite{valsesia2024onboard};
    \item hyperspectral-\ac{vae} \cite{park2025hyperspectral};
    \item \ac{msahific} \cite{wan2025msahific};
    \item {\acs{btcnet} V2} \cite{zhou2026btc}; and
    \item mean \& scale hyperprior \cite{minnen2018joint}.
\end{enumerate*}

\begin{figure*}
    \centering
    \begin{subfigure}{\linewidth}
        \centering
        \ref{sharedlegend-sota}
        \vspace{1em}
    \end{subfigure}
    \begin{subfigure}{0.48\linewidth}
        \centering
        \begin{tikzpicture}
            \begin{axis}[
                    width=\linewidth-6.97697pt,
                    height=0.8\linewidth,
                    xlabel={\acs{cr}},
                    ylabel={\acs{psnr} [\si{\decibel}]},
                    legend columns=5,
                    % transpose legend,
                    legend cell align={left},
                    legend pos=north east,
                    xmode=log,
                    log basis x={2},
                    log ticks with fixed point,
                    minor tick num=4,
                    ymin=32.5,
                    ymax=62.5,
                    xmin=2,
                    xmax=16384,
                    legend style={nodes={scale=0.65, transform shape}},
                    xticklabel={
                        \pgfkeys{/pgf/fpu=true}
                        \pgfmathparse{int(2^\tick)}
                        \pgfmathprintnumber[fixed]{\pgfmathresult}
                    },
                    legend to name=sharedlegend-sota,
                ]
    
                \addplot[cyan,dashed,mark=pentagon*,mark options={solid}] coordinates {
                    (4.496926408056514, 84.78510284423828)
                    (8.055907758301597, 65.63658905029297)
                    (16.002064534761335, 54.256221771240234)
                    (28.004665796950693, 49.11704635620117)
                    (32.0055978731953, 48.17148208618164)
                    (50.00820630746861, 45.49775695800781)
                    (64.00993033199593, 44.26495361328125)
                    (100.01443666502945, 42.33738327026367)
                    (128.01734169296608, 41.40114212036133)
                    (202.0133409860396, 39.83381271362305)
                    (255.99386549451086, 39.09691619873047)
                    (511.89092896530104, 37.16923522949219)
                    (1023.2993725386474, 35.469783782958984)
                    (2044.8653967626858, 33.85462188720703)
                };
                \addlegendentry{JPEG2000};
                
                \addplot[violet,dotted,mark=diamond*,mark options={solid}] coordinates {
                    (3.9116120308998594, 60.84214401245117)
                    (7.671016461229654, 57.73904037475586)
                    (15.334853118339357, 55.34886169433594)
                    (28.456183794194526, 53.02031326293945)
                    (49.73353770324287, 49.92660903930664)
                    (99.16605740995986, 44.76227951049805)
                    (197.13890874434122, 39.86271286010742)
                };
                \addlegendentry{\acs{pca}};
    
                \addplot[red,mark=*] coordinates {
                    (3.960784314, 54.84735990895165)        % cae1d
                    (7.769230769, 53.902654146940)          % cae1d_4bpppc
                    (15.538461538, 52.38174220872962)       % cae1d_2bpppc
                    (28.857142857, 48.952538482825375)      % cae1d_1bpppc
                };
                \addlegendentry{\acs{1dcae} \cite{kuester20211d}};
    
                \addplot[teal,mark=square*] coordinates {
                    (3.96078431373, 43.29476347234514)      % sscnet_like_cae1d
                    (5.33333333333, 43.52298931280772)      % sscnet_6bpppc
                    (8, 43.69138220945994)                  % sscnet_4bpppc
                    (12.625, 43.64347089661492)             % sscnet
                    (15.8431372549, 43.60347966353098)      % sscnet_like_cae3d
                    (32, 43.2428080505795)                  % sscnet_1bpppc
                    (50.5, 43.847)
                    (101, 43.597)
                    (202, 43.149)
                    (269.33, 42.918)
                    (517.12, 41.855)
                    (1077.3, 40.107)
                    (2154.7, 38.365)
                    (4309.3, 36.875)
                    (12928, 33.053)
                };
                \addlegendentry{\acs{sscnet} \cite{la2022hyperspectral}};
    
                \addplot[blue,mark=triangle*] coordinates {
                    (3.96078431373, 39.94142961502075)      % cae3d_like_cae1d
                    (7.92156862745, 39.6920463376575)       % cae3d_4bpppc
                    (15.8431372549, 39.5384429163403)       % cae3d
                    (31.6862745098, 39.06108974227706)      % cae3d_1bpppc
                    (50.698, 38.052)
                    (126.75, 37.414)
                    (253.49, 36.675)
                };
                \addlegendentry{\acs{3dcae} \cite{chong2021end}};
    
                \addplot[magenta,mark=asterisk] coordinates {
                    (3.9608, 56.294)                        % 2000 epochs, LR=1e-3, CR=4
                    (7.7692, 55.377)                        % 2000 epochs, LR=1e-3, CR=8
                    (15.538, 53.202)                        % 2000 epochs, LR=1e-3, CR=16
                    (28.857, 50.257)                        % 2000 epochs, LR=1e-3, CR=32
                    (50.5, 46.273)                          % 2000 epochs, LR=1e-3, CR=64
                    (101, 40.354)                           % 2000 epochs, LR=1e-3, CR=128
                    (202, 33.912)                           % 2000 epochs, LR=1e-3, CR=256
                };
                \addlegendentry{\acs{hycot} \cite{fuchs2024hycot}};
    
                \addplot[yellow,mark=+] coordinates {
                    (8.0256684492, 38.26842837273992)
                    (11.7755982738, 38.212795549374135)
                    (20.2264150943, 38.184979137691236)
                    (32.6260869565, 34.06815020862309)
                };
                \addlegendentry{\acs{hific}\textsubscript{\acs{se}} \cite{fuchs2024generative}}
    
                \addplot[brown,mark=Mercedes star flipped] coordinates {
                    (3.9608, 56.444)     % 0x
                    (7.7692, 55.155)     % 0x
                    (15.538, 52.828)    % 0x
                    (28.86, 49.719)    % 0x
                    (50.50, 48.61)    % 1x
                    (101.0, 46.84)    % 2x
                    (202.0, 45.09)    % 2x
                    (404.0, 42.76)    % 3x
                    (808.0, 41.46)    % 3x
                    % (12928.0, 34.307) % 3x
                    (12928.0, 35.371) % 4x
                };
                \addlegendentry{\acs{hycass} \cite{fuchs2025adjustable}};
                % Add nodes (labels)
                % \node[font=\footnotesize,brown] at (axis cs:3.96, 56.444) [anchor=south] {\SI{0}{\times}};
                % \node[font=\footnotesize,brown] at (axis cs:7.77, 55.155) [anchor=south] {\SI{0}{\times}};
                % \node[font=\footnotesize,brown] at (axis cs:15.54, 52.828) [anchor=south] {\SI{0}{\times}};
                % \node[font=\footnotesize,brown] at (axis cs:28.86, 49.719) [anchor=south] {\SI{0}{\times}};
                % \node[font=\footnotesize,brown] at (axis cs:50.50, 48.61) [anchor=south ] {\SI{1}{\times}};
                % \node[font=\footnotesize,brown] at (axis cs:101.0, 46.84) [anchor=south] {\SI{2}{\times}};
                % \node[font=\footnotesize,brown] at (axis cs:202.0, 45.09) [anchor=south] {\SI{2}{\times}};
                % \node[font=\footnotesize,brown] at (axis cs:404.0, 42.76) [anchor=south] {\SI{3}{\times}};
                % \node[font=\footnotesize,brown] at (axis cs:808.0, 41.46) [anchor=south] {\SI{3}{\times}};
                % \node[font=\footnotesize,brown] at (axis cs:12928.0, 35.371) [anchor=south] {\SI{4}{\times}};
    
                \addplot[orange,mark=10-pointed star] coordinates {
                    (10.255691769, 59.86301369863014)
                    (11.133079848, 59.726027397260275)
                    (16.589235127, 53.35616438356164)
                    (35.926380368, 48.150684931506845)
                };
                \addlegendentry{Verdú et al. \cite{mijares2023reduced}};

                \addplot[lime,mark=oplus] coordinates {
                    (29.495652173901863, 52.810615199034984)
                    (28.320772946843476, 58.61881785283474)
                    (26.187439613500622, 64.31845597104946)
                    (22.570048309136645, 69.8552472858866)
                    (21.147826086908076, 71.70084439083233)
                    (19.107246376754034, 73.87213510253318)
                    (16.262801932296895, 76.74909529553679)
                    (11.563285024063354, 81.36308805790108)
                };
                \addlegendentry{{\acs{linerwkv}}-L \cite{valsesia2024onboard}};
    
                \addplot[olive,mark=halfsquare left*] coordinates {
                    (4.208333333333333, 43.30593818950072)      % embed_dim_768
                    (16.833333333333332, 43.806451400743555)    % embed_dim_192
                    (67.33333333333333, 43.37962673310096)      % embed_dim_48
                    (538.6666666666666, 41.57015017782532)      % embed_dim_6
                    (808.0, 40.67018600211546)                  % embed_dim_4
                    (1077.3333333333333, 39.944282558298404)    % embed_dim_3
                    (1616.0, 38.91358190126477)                 % embed_dim_2
                    (3232.0, 37.28220299371333)                 % embed_dim_1
                };
                \addlegendentry{Hyperspectral-VAE \cite{park2025hyperspectral}};
    
                \addplot[green,mark=halfsquare right*] coordinates {
                    (7.942842022, 45.083449235048676)
                    (15.52016546, 44.49930458970793)
                    (30.258064516, 43.94297635605007)
                };
                \addlegendentry{\ac{msahific} \cite{wan2025msahific}};

                \addplot[gray,mark=otimes] coordinates {
                    (119.70370370370371, 38.53309023950907)
                    (523.4100090057816, 38.837033536560746)
                };
                \addlegendentry{\acs{btcnet} V2 \cite{zhou2026btc}};
    
                \addplot[pink,mark=halfcircle*] coordinates {
                    (294.8588411044161, 43.5334135276323)       % meanscalehyperprior_hsi_n192_m320-lmbd_1e+02-ans
                    (1335.7257379223518, 44.02959816636367)     % meanscalehyperprior_hsi_n192_m320-lmbd_1e+00-ans
                    (2195.629192269011, 42.760835911524204)     % meanscalehyperprior_hsi_n128_m192-lmbd_1e-02-ans
                    (2875.956812550863, 42.11744812781962)      % meanscalehyperprior_hsi_n128_m192-lmbd_1e-04-ans
                    (6174.667424422025, 41.59496062687939)      % meanscalehyperprior_hsi_n128_m192-lmbd_1e-06-ans
                    % (167600.7594869763, 34.599445050856254)     % meanscalehyperprior_hsi_n192_m320-lmbd_1e-08-ans
                };
                \addlegendentry{Mean \& scale hyperprior \cite{minnen2018joint}};

                \addplot[purple,mark=x] coordinates {
                    (5.658355844829346, 52.791458524964604)     % ours_s2_n768_m1280_k3-lmbd_1e+04
                    (25.353835559962334, 52.85995545150717)     % ours_s2_n768_m1280_k3-lmbd_1e+02
                    % (28.451260780823468, 52.84579559737854)     % ours_s2_n768_m1280_k3-lmbd_1e+00
                    (78.66339385689716, 52.44036617561254)      % ours_s2_n768_m1280_k3-lmbd_1e-02
                    (569.5933710346746, 50.311646341136274)     % ours_s2_n768_m1280_k3-lmbd_1e-04
                    (1581.1412127328085, 47.34807188932122)     % ours_s2_n768_m1280_k3-lmbd_1e-05
                    (8042.156799796971, 40.67069776251173)      % ours_s2_n768_m1280_k3-lmbd_1e-06
                    (34363.759551780204, 36.502651234519696)    % ours_s2_n768_m1280_k3-lmbd_1e-07
                };  
                \addlegendentry{\ac{ours}}; % \textbar{} S=2 \textbar{} M=1280 \textbar{} N=768 \textbar{} k=3 \textbar{} $\lambda \downarrow \uparrow$
                % Add nodes (labels) at exact coordinates
                \node[font=\scriptsize,purple,anchor=north] at (axis cs:5.658355844829346, 52.791458524964604) {\num{1e+4}};
                \node[font=\scriptsize,purple,anchor=south] at (axis cs:25.353835559962334, 52.85995545150717) {\num{1e+2}};
                % \node[font=\scriptsize,purple,anchor=south] at (axis cs:28.451260780823468, 52.84579559737854) {\num{1e+0}};
                \node[font=\scriptsize,purple,anchor=south] at (axis cs:78.66339385689716, 52.44036617561254) {\num{1e-2}};
                \node[font=\scriptsize,purple,anchor=south] at (axis cs:569.5933710346746, 50.311646341136274) {\num{1e-4}};
                \node[font=\scriptsize,purple,anchor=south] at (axis cs:1581.1412127328085, 47.34807188932122) {\num{1e-5}};
                \node[font=\scriptsize,purple,anchor=south] at (axis cs:8042.156799796971, 40.67069776251173) {\num{1e-6}};
                \node[font=\scriptsize,purple,anchor=south] at (axis cs:34363.759551780204, 36.502651234519696) {\num{1e-7}};

            \end{axis}
        \end{tikzpicture}
        \caption{HySpecNet-11k (easy split)}
        \label{fig:rd-plot-sota-hyspecnet11k}
    \end{subfigure}
    \begin{subfigure}{0.48\linewidth}
        \centering
        \begin{tikzpicture}
            \begin{axis}[
                width=\linewidth-6.97697pt,
                height=0.8\linewidth,
                legend columns=2,
                % transpose legend,
                legend cell align={left},
                xmode=log,
                log basis x={2},
                log ticks with fixed point,
                minor tick num=4,
                ymin=32.5,
                ymax=62.5,
                xmin=2,
                xmax=16384,
                legend style={nodes={scale=0.65, transform shape}},
                xticklabel={
                    \pgfkeys{/pgf/fpu=true}
                    \pgfmathparse{int(2^\tick)}
                    \pgfmathprintnumber[fixed]{\pgfmathresult}
                },
                xlabel={\acs{cr}},
                ylabel={\acs{psnr} [\si{\decibel}]},
                ]
                \addplot[cyan,dashed,mark=pentagon*,mark options={solid}] coordinates {
                    (4.0026843771946075, 80.45751953125)
                    (8.000290205879358, 62.84672546386719)
                    (16.00153130503918, 51.70823287963867)
                    (32.00380074072066, 46.77833557128906)
                    (64.00925486384209, 43.94217300415039)
                    (128.00834985490735, 41.71170425415039)
                    (255.96729842104898, 39.466590881347656)
                    (511.8083462658521, 37.18948745727539)
                    (1023.1360914819182, 34.7945671081543)
                    (2043.8843954051897, 29.743745803833008)
                };
                % \addlegendentry{JPEG2000};
                
                \addplot[violet,dotted,mark=diamond*,mark options={solid}] coordinates {
                    (4.201238956027217, 46.9564323425293)
                    (8.223734659241973, 46.0838737487793)
                    (16.43240237332737, 45.62356185913086)
                    (32.80470229068165, 45.18788146972656)
                    (65.37030891996525, 44.1751594543457)
                    (97.69942435424355, 43.125091552734375)
                    (193.29330685667563, 37.20418167114258)
                };
                % \addlegendentry{\ac{pca}};
                
                \addplot[red,mark=*] coordinates {
                    (3.9677, 45.757)        % cae1d_cr004
                    (7.8511, 45.318)        % cae1d_cr008
                    (15.375, 44.620)        % cae1d_cr016
                    (30.750, 44.671)        % cae1d_cr032
                };
                % \addlegendentry{\ac{1dcae} \cite{kuester20211d}};
                
                \addplot[teal,mark=square*] coordinates {
                    (4.0, 40.749)
                    (8.0, 40.603)
                    (16.0, 40.384)
                    (32.0, 40.182)
                    (184.5, 39.729)
                    (1026.8, 38.979)
                    (4723.2, 38.375)
                    (11808, 36.600)
                };
                % \addlegendentry{\ac{sscnet} \cite{la2022hyperspectral}};
    
                \addplot[blue,mark=triangle*] coordinates {
                    (3.9677, 40.245)
                    (15.871, 40.102)
                    (63.484, 38.651)
                    (253.94, 37.654)
                };
                % \addlegendentry{\ac{3dcae} \cite{chong2021end}};
    
                \addplot[magenta,mark=asterisk] coordinates {
                    (3.9677, 44.502)
                    (30.750, 44.464)
                    (123.0, 42.975)
                    (369.0, 34.657)
                };
                % \addlegendentry{\ac{hycot} \cite{fuchs2024hycot}};
    
                \addplot[brown,mark=Mercedes star flipped] coordinates {
                    (4.01, 44.86)
                    (7.85, 44.89)
                    (16.04, 44.86)
                    (30.75, 44.79)
                    (61.5, 44.23)
                    (123, 42.93)
                    (184.5, 42.40)
                    (369, 42.17)
                    (738, 41.57)
                    (1476, 41.04)
                };
                % \addlegendentry{\ac{hycass} \cite{fuchs2025adjustable}};

                \addplot[pink,mark=halfcircle*] coordinates {
                    (660.718431118577, 40.883151511351265)  % meanscalehyperprior_hsi_n192_m320-lmbd_1e+02-ans
                    (2258.4277222722812, 40.90618607401848) % meanscalehyperprior_hsi_n192_m320-lmbd_1e+00-ans
                    (4586.781087659769, 40.3718613187472)   % meanscalehyperprior_hsi_n128_m192-lmbd_1e-02-ans
                    (7369.970517190776, 39.78493210673332)  % meanscalehyperprior_hsi_n128_m192-lmbd_1e-04-ans
                    (13727.940457293033, 39.61252527435621) % meanscalehyperprior_hsi_n128_m192-lmbd_1e-06-ans
                };
                % \addlegendentry{Mean \& scale hyperprior \cite{minnen2018joint}};
    
                \addplot[purple,mark=x] coordinates {
                    % (18.434940046121003, 43.95902168750763) % ours_s2_n768_m1280_k3-lmbd_1e+04
                    (19.21961973459278, 44.018004993597664) % ours_s2_n768_m1280_k3-lmbd_1e+02
                    (86.33514970289097, 43.766751646995544) % ours_s2_n768_m1280_k3-lmbd_1e+00
                    (518.4216091324312, 43.2636869152387)   % ours_s2_n768_m1280_k3-lmbd_1e-02
                    (2383.301247433498, 42.492975364128746) % ours_s2_n768_m1280_k3-lmbd_1e-04
                    (5787.842267853895, 42.38384420673052) % ours_s2_n768_m1280_k3-lmbd_1e-05
                };  
                % \addlegendentry{\ac{ours}}; % \textbar{} S=2 \textbar{} M=1280 \textbar{} N=768 \textbar{} k=3 \textbar{} $\lambda \downarrow \uparrow$

                % \node[font=\scriptsize,purple,anchor=north] at (axis cs:18.434940046121003,43.95902168750763) {\num{1e+4}};
                \node[font=\scriptsize,purple,anchor=north] at (axis cs:19.21961973459278,44.018004993597664) {\num{1e+2}};
                \node[font=\scriptsize,purple,anchor=south] at (axis cs:86.33514970289097,43.766751646995544) {\num{1e+0}};
                \node[font=\scriptsize,purple,anchor=south] at (axis cs:518.4216091324312,43.2636869152387) {\num{1e-2}};
                \node[font=\scriptsize,purple,anchor=south] at (axis cs:2383.301247433498,42.492975364128746) {\num{1e-4}};
                \node[font=\scriptsize,purple,anchor=south] at (axis cs:5787.842267853895,42.38384420673052) {\num{1e-5}};
            \end{axis}
        \end{tikzpicture}
        \caption{MLRetSet}
        \label{fig:rd-plot-sota-mlretset}
    \end{subfigure}
    \caption{\Acl{rd} performance on the test set of (\subref{fig:rd-plot-sota-hyspecnet11k}) HySpecNet-11k \cite{fuchs2023hyspecnet} (easy split) and (\subref{fig:rd-plot-sota-mlretset}) MLRetSet \cite{omruuzun2024novel}. The annotated numbers indicate the corresponding values of $\lambda$ during training.}
    \label{fig:rd-plot-sota}
\end{figure*}

\subsubsection{HySpecNet-11k} % Generalization Capability
In the first set of experiments, we compare \ac{ours} with the state of the art on the HySpecNet-11k dataset.
\autoref{fig:rd-plot-sota} (\subref{fig:rd-plot-sota-hyspecnet11k}) shows the \ac{rd} curves of the considered models, where the rate is expressed as the \ac{cr} in a logarithmic scale and the distortion is measured as \ac{psnr} in \ac{decibel}.
Due to its metric-driven design adapted to the \ac{hsi} domain, \ac{ours} consistently achieves higher reconstruction fidelity compared to all state-of-the-art methods for $\acp{cr} > \num{32}$.
At lower $\acp{cr} < \num{32}$, {\acs{linerwkv}}-L achieves the highest reconstruction fidelity, as its specific design for lossless and near-lossless compression is particularly beneficial in this {\ac{cr}} range. However, this also poses a limitation in the achievable {\acp{cr}}, restricting {\acs{linerwkv}} to a relatively narrow {\ac{cr}} range.
It is worth noting that at lower $\acp{cr} < \num{32}$, traditional approaches (i.e., JPEG2000 and \ac{pca}) also achieve better reconstruction quality than \ac{ours}, highlighting the effectiveness of predefined mathematical transformations in low-compression regimes.
However, their reconstruction performance degrades rapidly as the \ac{cr} increases beyond this range, indicating a limited ability to maintain reconstruction fidelity under stronger compression constraints.
This behavior highlights the need for learning-based compression that can better model the complex spatio-spectral redundancies of \acp{hsi} in scenarios with strict bandwidth or storage space constraints.
For medium $\acp{cr} \in [ \num{32}, \num{512} ]$, \ac{ours} is highly effective as its metric-driven design allows for a transform capacity inside the latent and hyperlatent that is adjusted to the characteristics of \acp{hsi}.
Due to balanced spatial and spectral feature learning, \ac{ours} reduces the entropy of the latent representation while simultaneously increasing reconstruction fidelity.
For high $\acp{cr} > \num{512}$, \ac{ours} maintains strong reconstruction fidelity.
However, its advantage diminishes compared to models with high spatial compression (e.g., \ac{sscnet}, \ac{hycass}, and mean \& scale hyperprior), as spatial redundancy becomes increasingly important in this \ac{cr} regime \cite{fuchs2025adjustable}.
In summary, \ac{ours} improves upon \ac{hycass}, the formerly best-performing model, by \SI{4.66}{\decibel} in \ac{bdpsnr} across the entire \ac{rd} curve.
Compared to the mean \& scale hyperprior, the metric-driven selection of our configurable architecture yields substantial improvements, further highlighting the effectiveness of the proposed model.

\subsubsection{MLRetSet} % Hyperparameter Robustness
In the second set of experiments, we assess the generalization capability of the proposed model and the robustness of the selected hyperparameters by comparing its \ac{rd} performance with those obtained by state-of-the-art models on the MLRetSet \cite{omruuzun2024novel} dataset, as shown in \autoref{fig:rd-plot-sota} (\subref{fig:rd-plot-sota-mlretset}).
The results demonstrate that the proposed model achieves the highest reconstruction fidelity among the evaluated compression models for $\acp{cr} > \num{64}$.
For lower $\acp{cr} < \num{64}$, spectral learning-based models (i.e., \ac{1dcae}, \ac{hycot}, and \ac{hycass} with zero spatial stages) are particularly effective, as the large number of \SI{369}{\sband} introduces strong spectral redundancies.
However, in this \ac{cr} regime, the traditional baselines JPEG2000 and \ac{pca} further improve upon this, suggesting that the redundancies can be effectively captured by linear transformations under low compression constraints.
Compared to the HySpecNet-11k dataset, traditional methods remain competitive over a wider range of \acp{cr} (up to $\ac{cr} \approx \num{64}$ instead of \num{32}).
This can be attributed to the higher number of spectral bands and the higher spatial resolution of MLRetSet, which introduces stronger local spatial and spectral redundancies due to the narrower distances of adjacent bands and pixels, which can be effectively captured by linear transformations.
Once the \ac{cr} exceeds \num{64}, the proposed model demonstrates its success by effectively capturing spatial and spectral redundancies via the metric-driven selection of the configurable architecture.
Compared to HySpecNet-11k, the absolute reconstruction fidelity (in terms of \ac{psnr}) is lower for this dataset across all models.
This result indicates that the fine spatial and spectral details introduced by the higher spatial resolution and number of spectral bands pose a challenge for the compression models in achieving fine-grained reconstructions.
However, for very high $\acp{cr} > \num{4096}$, the reconstruction fidelity surpasses that observed for HySpecNet-11k.
This behavior can be attributed to the different characteristics of the datasets, particularly the higher spatial resolution and increased number of spectral bands, which increase the spatial and spectral redundancies and therefore allow for more efficient compression under very strong compression constraints.
Overall, the results demonstrate the strong generalization capability of the proposed model and the robustness of the selected hyperparameters across \acp{hsi} with different sensor characteristics.

\subsubsection{Computational Complexity}
In the third set of experiments, we compare the computational complexity of {\ac{ours}} with those of other state-of-the-art learning-based {\ac{hsi}} compression models.
In {\autoref{fig:complexity-sota}}, the {\ac{flops}}, the number of model parameters, and the corresponding {\acp{psnr}} on the HySpecNet-11k test set (easy split) are reported at a fixed $\ac{cr} \approx \num{32}$.
As shown in the figure, {\ac{ours}} \hl{(which is designed for ground-based compression)} achieves the highest reconstruction quality while also requiring a relatively high amount of \ac{flops} (\SI[round-mode=places,round-precision=2]{351.37693286}{\giga\flops}) and model parameters ($\sim$\SI{74.60}{\mega\nothing}) compared to e.g. \acs{linerwkv}-L (\SI[round-mode=places,round-precision=2]{52.428800000}{\giga\flops} and $\sim$\SI{0.9}{\mega\nothing}), which is specifically designed for onboard compression.
We would like to mention that, while this \ac{cr} operating point facilitates a comparison of the computational
complexity across many models, selecting a higher \ac{cr} would result in a more significant \ac{psnr} improvement for the proposed model at the expense of limiting the number of models that could be included in the comparison.
Nevertheless, when compared with other {\ac{vae}}-based models, such as hyperspectral-{\ac{vae}}, mean \& scale hyperprior, and Verdú et al., the increase in model size by a factor of {\SIrange{3}{5}{}} yields a \SIrange{4}{9}{\decibel} gain in {\ac{psnr}}.
In summary, \ac{ours} achieves improved \ac{rd} performance over the state of the art at medium and high \acp{cr} at the cost of a higher computational complexity.
We consider {\ac{ours}} as a first step towards large-scale deep learning paradigms for \hl{ground-based} {\ac{hsi}} compression, enabling more scalable and data-driven compression.
We would like to note that the training of a single \ac{ours} model typically requires only a few hours on a single \acs{gpu} due to its fast convergence within relatively few epochs.
In contrast, other models require several days (e.g., \ac{1dcae}, and \ac{3dcae}) and multiple \acsp{gpu} (e.g., \acs{linerwkv}-L) to train.
However, it needs to be mentioned that \ac{ours}'s metric-driven hyperparameter selection strategy (see \autoref{subsec:metric-driven}) necessitates multiple training runs to evaluate different model configurations.
Importantly, this procedure is conducted only once in an offline manner.
After the hyperparameters are determined, they are fixed and consistently used for all subsequent experiments.
\begin{figure}
    \centering
    \begin{subfigure}{\linewidth}
        \centering
        \ref{sharedlegend-cc}
        \vspace{1em}
    \end{subfigure}
    % \centering
    \begin{tikzpicture}
        \begin{axis}[
                width=\linewidth,
                height=0.8\linewidth,
                xlabel={\acs{flops} [\hl{{\si{\giga\flops}}}]},
                ylabel={\acs{psnr} [\si{\decibel}]},
                legend columns=2,
                % transpose legend,
                legend cell align={left},
                minor tick num=4,
                xmin=0.1,xmax=40000,
                xmode=log,
                log ticks with fixed point,
                legend style={nodes={scale=0.65, transform shape}},
                legend to name=sharedlegend-cc,
                ]
                \addplot[
                    scatter,
                    only marks,
                    scatter src=explicit symbolic,
                    scatter/classes={
                        % pca={mark=*,violet,mark size=1pt},
                        cae1d={mark=*,red},
                        sscnet={mark=square*,teal},
                        cae3d={mark=triangle*,blue},
                        hycot={mark=asterisk,magenta},
                        hycass={mark=Mercedes star flipped,brown},
                        verdu={mark=10-pointed star,orange},
                        linerwkv={mark=oplus,lime},
                        hyvae={mark=halfsquare left*,olive},
                        btcnet={mark=otimes,gray},
                        meanscale={mark=halfcircle*,green},
                        ours={mark=x,purple,mark size=2pt}
                    },
                ]
                table[
                    x expr=\thisrow{flops}/1e9,
                    y=psnr,
                    col sep=space,
                    meta=model,
                ] {data/computational_complexity.txt};
                % \addlegendentry{\ac{pca}};
                \addlegendentry{\ac{1dcae} \cite{kuester20211d} ($\sim$\SI[round-mode=places,round-precision=2]{3.852482}{\mega\nothing})};
                \addlegendentry{\ac{sscnet} \cite{la2022hyperspectral} ($\sim$\SI[round-mode=places,round-precision=2]{8.068082}{\mega\nothing})};
                \addlegendentry{\ac{3dcae} \cite{chong2021end} ($\sim$\SI[round-mode=places,round-precision=2]{0.485371}{\mega\nothing})};
                \addlegendentry{\ac{hycot} \cite{fuchs2024hycot} ($\sim$\SI[round-mode=places,round-precision=2]{0.398069}{\mega\nothing})};
                \addlegendentry{\ac{hycass} \cite{fuchs2025adjustable} ($\sim$\SI[round-mode=places,round-precision=2]{0.430289}{\mega\nothing})};
                \addlegendentry{Verdú et al. \cite{mijares2023reduced} ($\sim$\SI[round-mode=places,round-precision=2]{13.960000}{\mega\nothing})};
                \addlegendentry{\acs{linerwkv}-L \cite{valsesia2024onboard} ($\sim$\SI[round-mode=places,round-precision=2]{0.900000}{\mega\nothing})};
                \addlegendentry{Hyperspectral-VAE \cite{park2025hyperspectral} ($\sim$\SI[round-mode=places,round-precision=2]{25.167947}{\mega\nothing})};
                \addlegendentry{\acs{btcnet} V2 (\num{32}-bit) \cite{zhou2026btc} ($\sim$\SI[round-mode=places,round-precision=2]{12.601681}{\mega\nothing})};
                \addlegendentry{Mean \& scale hyperprior \cite{minnen2018joint} ($\sim$\SI[round-mode=places,round-precision=2]{19.472298}{\mega\nothing})}; 
                \addlegendentry{\ac{ours} ($\sim$\SI[round-mode=places,round-precision=2]{74.595402}{\mega\nothing})}; 
                \addplot[
                    scatter,
                    only marks,
                    scatter src=explicit symbolic,
                    scatter/classes={
                        cae1d={mark=*,red,opacity=0.2},
                        sscnet={mark=*,teal,opacity=0.2},
                        cae3d={mark=*,blue,opacity=0.2},
                        hycot={mark=*,magenta,opacity=0.2},
                        hycass={mark=*,brown,opacity=0.2},
                        verdu={mark=*,orange,opacity=0.2},
                        linerwkv={mark=*,lime,opacity=0.2},
                        hyvae={mark=*,olive,opacity=0.2},
                        btcnet={mark=*,gray,opacity=0.2},
                        meanscale={mark=*,green,opacity=0.2},
                        ours={mark=*,purple,opacity=0.2}
                    },
                    visualization depends on={ln(\thisrow{params})/ln(10) \as \lp},
                    scatter/@pre marker code/.append style={
                        /tikz/mark size={\lp}
                    }
                ]
                table[
                    x expr=\thisrow{flops}/1e9,
                    y=psnr,
                    col sep=space,
                    meta=model
                ] {data/computational_complexity.txt};
                
                % Mark inacurrate PSNR
                \addplot[
                    only marks,
                    mark=star,
                    mark options={
                        xshift=3.5pt,
                        yshift=3.5pt
                    }
                ] coordinates {
                    (53.507520000, 44.02959816636367)
                };
                \addplot[
                    only marks,
                    mark=star,
                    mark options={
                        xshift=3.5pt,
                        yshift=3.5pt
                    }
                ] coordinates {
                    (141.862735106, 38.53309023950907)
                };
            \end{axis}
    \end{tikzpicture}
    \caption{Computational complexity analysis for $\ac{cr} \approx \num{32}$. {\ac{flops}} are calculated using an {\ac{hsi}} of size $\num{128} \times \num{128} \times 202$. The size of the circles indicates the number of model parameters (logarithmic scale), which is also explicitly noted in the legend. For the models marked with a $\star$, the {\ac{psnr}} values should be interpreted with caution, as these models are trained and evaluated at a different {\ac{cr}}. However, the model size and {\ac{flops}} remain largely unaffected.}
    \label{fig:complexity-sota}
\end{figure}

\subsection{Reconstruction Results Analysis}
\subsubsection{True Color Reconstruction Results}
\begin{figure*}
    \centering
    \begin{subfigure}[t]{0.13\linewidth}
        \centering
        \begin{tikzpicture}[spy using outlines={green,circle,connect spies,magnification=4,size=15mm}]
            \node[inner sep=0pt,outer sep=0pt] (main) {\includegraphics[width=0.91\linewidth,height=0.91\linewidth]{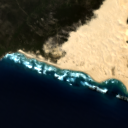}};
            \spy on (0.8,-0.2) in node [fill=white,anchor=south west,inner sep=0pt,outer sep=0pt] at (-0.8,-0.8);
        \end{tikzpicture}
        \caption*{\centering \ac{cr} \\ \scriptsize \ac{psnr}[\si{\decibel}]/\ac{ssim}/\ac{sa}}
        \caption{Original \ac{hsi}}
        \label{fig:rec-spat-0-in}
    \end{subfigure}
    \hfill
    \begin{subfigure}[t]{0.13\linewidth}
        \centering
        \begin{tikzpicture}[spy using outlines={green,circle,connect spies,magnification=4,size=15mm}]
            \node[inner sep=0pt,outer sep=0pt] (main) {\includegraphics[width=0.91\linewidth,height=0.91\linewidth]{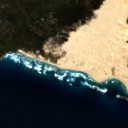}};
            \spy on (0.8,-0.2) in node [fill=white,anchor=south west,inner sep=0pt,outer sep=0pt] at (-0.8,-0.8);
        \end{tikzpicture}
        \caption*{\centering \num{28.86} \\ \scriptsize \num[round-mode=places,round-precision=1]{43.58}/\num[round-mode=places,round-precision=3]{0.9931}/\ang[round-mode=places,round-precision=1]{3.45}}
        \caption{\ac{hycot} \cite{fuchs2024hycot}}
        \label{fig:rec-spat-0-hycot}
    \end{subfigure}
    \hfill
    \begin{subfigure}[t]{0.13\linewidth}
        \centering
        \begin{tikzpicture}[spy using outlines={green,circle,connect spies,magnification=4,size=15mm}]
            \node[inner sep=0pt,outer sep=0pt] (main) {\includegraphics[width=0.91\linewidth,height=0.91\linewidth]{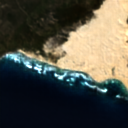}};
            \spy on (0.8,-0.2) in node [fill=white,anchor=south west,inner sep=0pt,outer sep=0pt] at (-0.8,-0.8);
        \end{tikzpicture}
        \caption*{\centering \num{32.00} \\ \scriptsize \num[round-mode=places,round-precision=1]{38.64}/\num[round-mode=places,round-precision=3]{0.9662}/\ang[round-mode=places,round-precision=1]{3.99}}
        \caption{\ac{sscnet} \cite{la2022hyperspectral}}
        \label{fig:rec-spat-0-sscnet}
    \end{subfigure}
    \hfill
    \begin{subfigure}[t]{0.13\linewidth}
        \centering
        \begin{tikzpicture}[spy using outlines={green,circle,connect spies,magnification=4,size=15mm}]
            \node[inner sep=0pt,outer sep=0pt] (main) {\includegraphics[width=0.91\linewidth,height=0.91\linewidth]{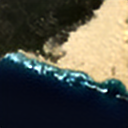}};
            \spy on (0.8,-0.2) in node [fill=white,anchor=south west,inner sep=0pt,outer sep=0pt] at (-0.8,-0.8);
        \end{tikzpicture}
        \caption*{\centering \num{31.69} \\ \scriptsize \num[round-mode=places,round-precision=1]{35.57}/\num[round-mode=places,round-precision=3]{0.9079}/\ang[round-mode=places,round-precision=1]{7.93}}
        \caption{\ac{3dcae} \cite{chong2021end}}
        \label{fig:rec-spat-0-3dcae}
    \end{subfigure}
    \hfill
    \begin{subfigure}[t]{0.13\linewidth}
        \centering
        \begin{tikzpicture}[spy using outlines={green,circle,connect spies,magnification=4,size=15mm}]
            \node[inner sep=0pt,outer sep=0pt] (main) {\includegraphics[width=0.91\linewidth,height=0.91\linewidth]{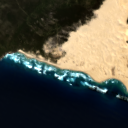}};
            \spy on (0.8,-0.2) in node [fill=white,anchor=south west,inner sep=0pt,outer sep=0pt] at (-0.8,-0.8);
        \end{tikzpicture}
        \caption*{\centering \num{28.86} \\ \scriptsize \num[round-mode=places,round-precision=1]{43.07}/\num[round-mode=places,round-precision=3]{0.9937}/\ang[round-mode=places,round-precision=1]{3.35}}
        \caption{\ac{hycass} \cite{fuchs2025adjustable}}
        \label{fig:rec-spat-0-hycass}
    \end{subfigure}
    \hfill
    \begin{subfigure}[t]{0.13\linewidth}
        \begin{tikzpicture}[spy using outlines={green,circle,connect spies,magnification=4,size=15mm}]
            \node[inner sep=0pt,outer sep=0pt] (main) {\includegraphics[width=0.91\linewidth,height=0.91\linewidth]{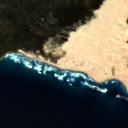}};
            \spy on (0.8,-0.2) in node [fill=white,anchor=south west,inner sep=0pt,outer sep=0pt] at (-0.8,-0.8);
        \end{tikzpicture}
        \caption*{\centering \num{33.19} \\ \scriptsize \num[round-mode=places,round-precision=1]{49.25}/\num[round-mode=places,round-precision=3]{0.9950}/\ang[round-mode=places,round-precision=1]{2.99}}
        \caption{\ac{pca}}
        \label{fig:rec-spat-0-pca}
    \end{subfigure}
    \hfill
    \begin{subfigure}[t]{0.13\linewidth}
        \centering
        \begin{tikzpicture}[spy using outlines={green,circle,connect spies,magnification=4,size=15mm}]
            \node[inner sep=0pt,outer sep=0pt] (main) {\includegraphics[width=0.91\linewidth,height=0.91\linewidth]{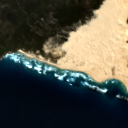}};
            \spy on (0.8,-0.2) in node [fill=white,anchor=south west,inner sep=0pt,outer sep=0pt] at (-0.8,-0.8);
        \end{tikzpicture}
        \caption*{\centering \num{29.82} \\ \scriptsize \textbf{\num[round-mode=places,round-precision=1]{50.82}}/\textbf{\num[round-mode=places,round-precision=3]{0.9956}}/\textbf{\ang[round-mode=places,round-precision=1]{2.67}}}
        \caption{\ac{ours}}
        \label{fig:rec-spat-0-ours}
    \end{subfigure}
    \vfill
    \vspace{0.5em}
    %%%
    \begin{subfigure}[t]{0.13\linewidth}
        \centering
        \includegraphics[width=0.91\linewidth,height=0.91\linewidth]{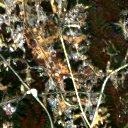}
        \caption*{\centering \ac{cr} \\ \scriptsize \ac{psnr}[\si{\decibel}]/\ac{ssim}/\ac{sa}}
        \caption{Original \ac{hsi}}
        \label{fig:rec-spat-1-in}
    \end{subfigure}
    \hfill
    \begin{subfigure}[t]{0.13\linewidth}
        \centering
        \includegraphics[width=0.91\linewidth,height=0.91\linewidth]{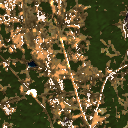}
        \caption*{\centering \num{202.00} \\ \scriptsize \num[round-mode=places,round-precision=1]{31.26}/\num[round-mode=places,round-precision=3]{0.8379}/\ang[round-mode=places,round-precision=1]{7.81}}
        \caption{\ac{hycot} \cite{fuchs2024hycot}}
        \label{fig:rec-spat-1-hycot}
    \end{subfigure}
    \hfill
    \begin{subfigure}[t]{0.13\linewidth}
        \centering
        \includegraphics[width=0.91\linewidth,height=0.91\linewidth]{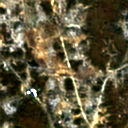}
        \caption*{\centering \num{202.00} \\ \scriptsize \num[round-mode=places,round-precision=1]{34.28}/\num[round-mode=places,round-precision=3]{0.8838}/\ang[round-mode=places,round-precision=1]{4.99}}
        \caption{\ac{sscnet} \cite{la2022hyperspectral}}
        \label{fig:rec-spat-1-sscnet}
    \end{subfigure}
    \hfill
    \begin{subfigure}[t]{0.13\linewidth}
        \centering
        \includegraphics[width=0.91\linewidth,height=0.91\linewidth]{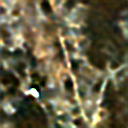}
        \caption*{\centering \num{253.49} \\ \scriptsize \num[round-mode=places,round-precision=1]{30.38}/\num[round-mode=places,round-precision=3]{0.7509}/\ang[round-mode=places,round-precision=1]{7.28}}
        \caption{\ac{3dcae} \cite{chong2021end}}
        \label{fig:rec-spat-1-3dcae}
    \end{subfigure}
    \hfill
    \begin{subfigure}[t]{0.13\linewidth}
        \centering
        \includegraphics[width=0.91\linewidth,height=0.91\linewidth]{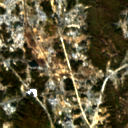}
        \caption*{\centering \num{202.00} \\ \scriptsize \num[round-mode=places,round-precision=1]{36.96}/\num[round-mode=places,round-precision=3]{0.9449}/\ang[round-mode=places,round-precision=1]{3.78}}
        \caption{\ac{hycass} \cite{fuchs2025adjustable}}
        \label{fig:rec-spat-1-hycass}
    \end{subfigure}
    \hfill
    \begin{subfigure}[t]{0.13\linewidth}
        \centering
        \includegraphics[width=0.91\linewidth,height=0.91\linewidth]{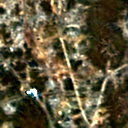}
        \caption*{\centering \num{243.94} \\ \scriptsize \num[round-mode=places,round-precision=1]{34.84}/\num[round-mode=places,round-precision=3]{0.9066}/\ang[round-mode=places,round-precision=1]{4.89}}
        \caption{Mean \& scale hyperprior \cite{minnen2018joint}}
        \label{fig:rec-spat-1-msh}
    \end{subfigure}
    \hfill
    \begin{subfigure}[t]{0.13\linewidth}
        \centering
        \includegraphics[width=0.91\linewidth,height=0.91\linewidth]{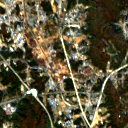}
        \caption*{\centering \num{295.92} \\ \scriptsize \textbf{\num[round-mode=places,round-precision=1]{46.42}}/\textbf{\num[round-mode=places,round-precision=3]{0.9939}}/\textbf{\ang[round-mode=places,round-precision=1]{1.45}}}
        \caption{\ac{ours}}
        \label{fig:rec-spat-1-ours}
    \end{subfigure}
    \caption{True color reconstruction results of \ac{ours} and state-of-the-art \ac{hsi} compression models for two different HySpecNet-11k \cite{fuchs2023hyspecnet} images. Contains modified EnMAP data ©DLR [2022].}
    \label{fig:rec-spat}
\end{figure*}

To visually assess the spatial reconstruction fidelity of the proposed model, the true color reconstruction results of \ac{ours} and several state-of-the-art learning-based \ac{hsi} compression models are presented in \autoref{fig:rec-spat}.
In order to analyze the compression artifacts under varying scene characteristics and compression constraints, two representative HySpecNet-11k images are evaluated at two different \acp{cr}.
For each reconstruction, the corresponding \ac{cr} as well as the \ac{psnr}, \ac{ssim}, and \ac{sa} are reported to enable a comprehensive comparison.
For the coastal scene in \autoref{fig:rec-spat} (\subref{fig:rec-spat-0-in}), a low \ac{cr} of approximately \num{32} was targeted, while targeting a medium \ac{cr} of \num{256} for the urban area scene in (\subref{fig:rec-spat-1-in}).
We would like to note that the \acp{cr} of \ac{hycot}, \ac{sscnet}, \ac{3dcae}, \ac{hycass} and \ac{pca} are fixed by their architectural design, while the \ac{cr} of variational \ac{hsi} compression models (i.e., mean \& scale hyperior and \ac{ours}) may vary across different \acp{hsi} depending on the entropy of the latent and hyperlatent representations.
Close inspection of (\subref{fig:rec-spat-0-hycot}) - (\subref{fig:rec-spat-0-ours}) and (\subref{fig:rec-spat-1-hycot}) - (\subref{fig:rec-spat-1-ours}) reveals noticeable differences in reconstruction fidelity both across the respective compression models and \acp{cr}.
Notably, \ac{hycot}, \ac{hycass} with \SI{0}{\times} spatial stages and \ac{pca} do not introduce spatial artifacts due to their pixelwise compression strategy in (\subref{fig:rec-spat-0-hycot}), (\subref{fig:rec-spat-0-hycass}) and (\subref{fig:rec-spat-0-pca}), respectively.
In fact, their reconstruction errors manifest as intensity offsets with increasing deviations for higher \acp{cr} as observed by comparing (\subref{fig:rec-spat-0-hycot}) and (\subref{fig:rec-spat-1-hycot}) with the respective original \ac{hsi}.
Spatial compression applied by \ac{sscnet} in (\subref{fig:rec-spat-0-sscnet}) leads to a loss of fine spatial details and a blurring along the boundaries between different land cover types, which becomes more pronounced for the increased \acp{cr} in (\subref{fig:rec-spat-1-sscnet}).
Spatial blurring can also be observed for \ac{3dcae} in (\subref{fig:rec-spat-0-3dcae}) due to the spatio-spectral compression using large convolutional kernels.
Increasing the \ac{cr} in (\subref{fig:rec-spat-0-3dcae}) introduces higher distortion, which substantially affects the perceptual quality.
For the mean \& scale hyperprior in (\subref{fig:rec-spat-1-msh}), one can observe loss of spatial structure due to the excessive spatial downsampling that is not adjusted to the low spatial resolution of \acp{hsi}.
In contrast, \ac{ours} achieves the best perceptual quality in both (\subref{fig:rec-spat-0-ours}) and (\subref{fig:rec-spat-1-ours}), preserving spatial structure and intensity values due to the metric-driven architectural design that balances spatial and spectral feature learning.
This is also reflected by the reported distortion metrics, indicating higher spatial and spectral reconstruction fidelity, surpassing the state of the art by \SI{1.5}{\decibel}/\num{0.001}/\ang{0.3} and \SI{9.4}{\decibel}/\num{0.049}/\ang{2.3} for \ac{psnr}/\ac{ssim}/\ac{sa} on the two different \acp{hsi}, respectively.
Overall, the visual comparisons demonstrate that \ac{ours} consistently preserves both spatial structures and spectral coherence across different land cover types and compression constraints, confirming its effectiveness for high-fidelity \ac{hsi} compression.
Similar behavior of the true color reconstruction results has been observed for other \acp{hsi} at various \acp{cr}, as well as on the MLRetSet dataset (not reported for space constraints).

\subsubsection{Reconstructed Spectral Signatures}
\begin{figure}
    \centering
    \begin{subfigure}{\linewidth}
        \centering
        \ref{sharedlegend-rec-spec}
        \vspace{1em}
    \end{subfigure}
    \begin{subfigure}{\linewidth}
        \begin{tikzpicture}
            \begin{axis}[
                width=\linewidth-1.09673pt,
                height=0.55\linewidth,
                legend columns=4,
                % transpose legend,
                legend cell align={left},
                minor tick num=4,
                xmin=0, xmax=201,
                ymin=0, ymax=0.8,
                legend style={nodes={scale=0.65, transform shape}},
                xlabel={Spectral Band Index},
                ylabel={Normalized Reflectance},
                no markers,
                legend to name=sharedlegend-rec-spec,
            ]
                % ORG / INPUT
                \addplot[green,densely dashed] table[x expr=\coordindex,col sep=comma,y=org] {data/rec-spec-0.txt};
                \addlegendentry{Original};
                % SOTA CR=202
                \addplot[magenta] table[x expr=\coordindex,col sep=comma,y=hycot_cr256] {data/rec-spec-0.txt};
                \addlegendentry{\ac{hycot} \cite{fuchs2024hycot}};
                \addplot[teal] table[x expr=\coordindex,col sep=comma,y=sscnet_cr202] {data/rec-spec-0.txt};
                \addlegendentry{\ac{sscnet} \cite{la2022hyperspectral}};
                \addplot[blue] table[x expr=\coordindex,col sep=comma,y=cae3d_cr253] {data/rec-spec-0.txt};
                \addlegendentry{\ac{3dcae} \cite{chong2021end}};
                \addplot[brown] table[x expr=\coordindex,col sep=comma,y=hycass_cr202_spatial2x_n128] {data/rec-spec-0.txt};
                \addlegendentry{\ac{hycass} \cite{fuchs2025adjustable}};
                \addplot[pink] table[x expr=\coordindex,col sep=comma,y=meanscalehyperprior_hsi_n192_m320-lmbd_1e+02] {data/rec-spec-0.txt};
                \addlegendentry{\makebox[0pt][l]{Mean \& scale hyperprior \cite{minnen2018joint}}};

                \addlegendimage{empty legend};
                \addlegendentry{};
                
                \addplot[purple] table[x expr=\coordindex,col sep=comma,y=ours_s2_n768_m1280_k3-lmbd_1e-04] {data/rec-spec-0.txt};
                \addlegendentry{\makebox[0pt][l]{\ac{ours}}};
                \addplot[green,densely dashed] table[x expr=\coordindex,col sep=comma,y=org] {data/rec-spec-0.txt};
            \end{axis}
        \end{tikzpicture}
        \caption{Dry Bare Soil}
        \label{fig:rec-spec-soil}
    \end{subfigure}
    \begin{subfigure}{\linewidth}
        \begin{tikzpicture}
            \begin{axis}[
                width=\linewidth-1.09673pt,
                height=0.55\linewidth,
                minor tick num=4,
                xmin=0, xmax=201,
                ymin=0, ymax=0.3,
                legend style={nodes={scale=0.65, transform shape}},
                xlabel={Spectral Band Index},
                ylabel={Normalized Reflectance},
                no markers,
            ]
                % ORG / INPUT
                \addplot[green,densely dashed] table[x expr=\coordindex,col sep=comma,y=org] {data/rec-spec-1.txt};
                % \addlegendentry{Original};
                % SOTA CR=202
                \addplot[magenta] table[x expr=\coordindex,col sep=comma,y=hycot_cr256] {data/rec-spec-1.txt};
                % \addlegendentry{\ac{hycot} \cite{fuchs2024hycot}};
                \addplot[teal] table[x expr=\coordindex,col sep=comma,y=sscnet_cr202] {data/rec-spec-1.txt};
                % \addlegendentry{\ac{sscnet} \cite{la2022hyperspectral}};
                \addplot[blue] table[x expr=\coordindex,col sep=comma,y=cae3d_cr253] {data/rec-spec-1.txt};
                % \addlegendentry{\ac{3dcae} \cite{chong2021end}};
                \addplot[brown] table[x expr=\coordindex,col sep=comma,y=hycass_cr202_spatial2x_n128] {data/rec-spec-1.txt};
                % \addlegendentry{\ac{hycass} \cite{fuchs2025adjustable}};
                \addplot[pink] table[x expr=\coordindex,col sep=comma,y=meanscalehyperprior_hsi_n192_m320-lmbd_1e+02] {data/rec-spec-1.txt};
                % \addlegendentry{Mean \& scale hyperprior \cite{minnen2018joint}};
                \addplot[purple] table[x expr=\coordindex,col sep=comma,y=ours_s2_n768_m1280_k3-lmbd_1e-04] {data/rec-spec-1.txt};
                % \addlegendentry{\ac{ours}};
                \addplot[green,densely dashed] table[x expr=\coordindex,col sep=comma,y=org] {data/rec-spec-1.txt};
            \end{axis}
        \end{tikzpicture}
        \caption{Vegetation}
        \label{fig:rec-spec-veg}
    \end{subfigure}
    \caption{Exemplary reconstructed spectral signatures of \ac{ours} and state-of-the-art learning-based \ac{hsi} compression models for $\ac{cr} \approx \num{256}$. Contains modified EnMAP data ©DLR [2022].}
    \label{fig:rec-spec}
\end{figure}
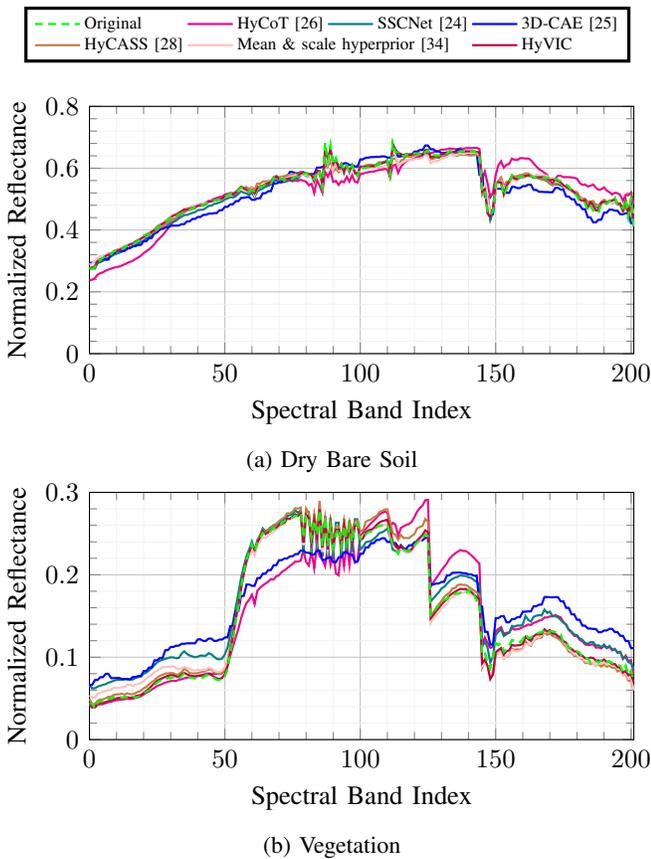

\autoref{fig:rec-spec} illustrates the reconstructed spectral signatures of two representative land cover types (i.e., dry bare soil and vegetation), derived from randomly selected pixels, for \ac{ours} and multiple state-of-the-art learning-based \ac{hsi} compression models at a \ac{cr} of approximately \num{256}.
For dry bare soil in (\subref{fig:rec-spec-soil}), all the reconstructions are able to follow the general shape of the spectral signature.
However, reconstruction errors for \ac{hycot} and \ac{3dcae} become more noticeable for the spectral bands \SIrange{150}{202}{}, likely due to the higher \ac{snr} of the sensor in the \ac{swir} bands.
Compared to \ac{sscnet}, \ac{hycass} and mean \& scale hyperprior, which all closely reconstruct the original spectral signature, \ac{ours} achieves the most accurate reconstruction of the detailed spikes in the spectral signature.

The spectral signature of a vegetation pixel surrounded by urban area, in \autoref{fig:rec-spec} (\subref{fig:rec-spec-veg}) presents a more challenging reconstruction problem.
Models that predominantly rely on spatial information, such as \ac{sscnet} and \ac{3dcae}, exhibit larger errors due to the spatial dimensionality reduction, which limits their ability to capture the detailed spectral variations of individual pixels.
This can be attributed to the vegetation pixel being surrounded by urban area, making it more difficult for these models to generalize effectively.
\ac{hycot} exhibits an increase in the reconstruction error, as the pronounced variation in the spectral signature is difficult to capture for a purely spectral compression model at such a high \ac{cr}.
\ac{ours} accurately reconstructs the original spectral signature, demonstrating only minor smoothing while largely preserving the overall spectral shape without introducing a significant offset.
We observed a similar behavior of the reconstructed spectral signatures for different land cover types and on the MLRetSet dataset (not reported for space constraints).

We observe that the reconstructed spectral signatures are generally smoother compared to the original data, suggesting that learning-based {\ac{hsi}} compression models possess an intrinsic denoising capability.
We would like to note that none of the considered {\ac{hsi}} compression models explicitly accounts for noisy spectral bands, which leads to non-uniform reconstruction errors across spectral bands, with some bands exhibiting higher errors than others.
Addressing this issue could further improve the consistency of spectral reconstruction fidelity as well as overall {\ac{rd}} performance.

\subsubsection{Distribution of SAs, PSNRs and CRs}
\begin{figure}
    \centering
    \begin{subfigure}{\linewidth}
        \begin{tikzpicture}
            \begin{axis}[
                name=main,
                ybar,
                bar width=3.9pt,
                width=\linewidth,
                height=0.55\linewidth,
                xlabel={Pixelwise \acs{sa} [\si{\degree}]},
                ylabel={Relative Frequency [\%]},
                ymin=0, ymax=12,
                xmin=0, xmax=5.1,
                xtick={0,1,2,3,4,5},
                xticklabels={0,1,2,3,4,$\geq 5$},
                grid=both,
                clip=true,
                minor tick num=4,
                xtick pos=bottom,
            ]
                \addplot table[
                    x index=0,
                    y index=1
                ] {data/ours-pixelwise_sas.txt};
                % mean
                \addplot[
                    sharp plot,
                    thick,
                    no markers,
                    draw=red,
                    % forget plot,
                ] coordinates {(2.258945066011709,0) (2.258945066011709,12)};
            \end{axis}
            % =========================
            % INSET PLOT (OUTSIDE AXIS)
            % =========================
            \begin{scope}[overlay]
                \node (plot) at (main.east) [
                    anchor=east,
                    xshift=1.5mm,
                    yshift=2.3mm
                ] {
                \begin{tikzpicture}
                    \begin{axis}[
                        width=4.5cm,
                        height=3.6cm,
                        xmin=0, xmax=201,
                        no markers,
                        axis background/.style={fill=white},
                        legend style={nodes={scale=0.65, transform shape}},
                        legend cell align={left},
                        legend pos=north west,
                    ]
                        \addplot[green,densely dashed] table[
                            x expr=\coordindex,
                            y index=0,
                            col sep=comma,
                        ] {data/ours-sa_outlier.txt};
                        \addlegendentry{Original};
                        \addplot[purple] table[
                            x expr=\coordindex,
                            y index=1,
                            col sep=comma,
                        ] {data/ours-sa_outlier.txt};
                        \addlegendentry{Reconstructed};
                        \addplot[green,densely dashed] table[
                            x expr=\coordindex,
                            y index=0,
                            col sep=comma,
                        ] {data/ours-sa_outlier.txt};
                    \end{axis}
                \end{tikzpicture}
                };
                \draw[arrow,dashed] ($(plot.east)+(-0.6,0)$) -- ++(0.3,0);
            \end{scope}
        \end{tikzpicture}
        \caption{}
        \label{fig:hist-sas}
    \end{subfigure}
    \hfill
    \begin{subfigure}{\linewidth}
        \begin{tikzpicture}
            \begin{axis}[
                ybar,
                width=\linewidth,
                height=0.55\linewidth,
                xlabel={Imagewise \acs{psnr} [\si{\decibel}]},
                ylabel={Relative Frequency [\%]},
                ymin=0,ymax=11,
                xmin=40,xmax=62,
                grid=both,
                clip=true,
                minor tick num=4,
                xtick pos=bottom,
            ]
                \addplot+[
                    hist={
                        bins=36,
                        data min=43,
                        data max=61,
                    }, 
                    y filter/.code={
                        \pgfmathparse{#1/1149*100}
                        \let\pgfmathresult=\pgfmathresult
                    }
                ] table [y index=0] {data/ours-imagewise_psnrs.txt};
                % mean
                \addplot[
                    sharp plot,
                    thick,
                    no markers,
                    draw=red,
                    % forget plot,
                ] coordinates {(50.312395157245476,0) (50.312395157245476,11)};
                % \addlegendentry{Average \ac{psnr}};
                % standard deviation
                % \addplot[
                %     sharp plot,
                %     thick,
                %     no markers,
                %     draw=red,
                %     dotted,
                %     % forget plot,
                % ] coordinates {(48.232783857,0) (48.232783857,11)};
                % \addplot[
                %     sharp plot,
                %     thick,
                %     no markers,
                %     draw=red,
                %     dotted,
                %     % forget plot,
                % ] coordinates {(52.392006457,0) (52.392006457,11)};
                % images
                \node[inner sep=0pt, outer sep=0pt] (img1) at (axis cs:43.75,7) {\includegraphics[width=.236\linewidth,height=.236\linewidth]{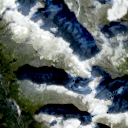}};
                \draw[arrow,dashed] (img1.south) -- (axis cs:43.41,0);
                \node[inner sep=0pt, outer sep=0pt] (img2) at (axis cs:58.25,7) {\includegraphics[width=.236\linewidth,height=.236\linewidth]{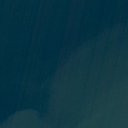}};
                \draw[arrow,dashed] (img2.south) -- (axis cs:60.29,0);
            \end{axis}
        \end{tikzpicture}
        \caption{}
        \label{fig:hist-psnrs}
    \end{subfigure}
    \hfill
    \begin{subfigure}{\linewidth}
        \begin{tikzpicture}
            \begin{axis}[
                ybar,
                width=\linewidth-4.91821pt,
                height=0.55\linewidth,
                xlabel={Imagewise \acs{cr}},
                ylabel={Relative Frequency [\%]},
                ymin=0,ymax=26,
                xmin=64,xmax=4096,
                grid=both,
                clip=true,
                minor tick num=4,
                xtick pos=bottom,
                xmode=log,
                log basis x={2},
                log ticks with fixed point,
                xticklabel={
                    \pgfkeys{/pgf/fpu=true}
                    \pgfmathparse{int(2^\tick)}
                    \pgfmathprintnumber[fixed]{\pgfmathresult}
                },
            ]
                % mean
                \addplot+[
                    hist={
                        bins=40,
                        data min=263,
                        data max=3876,
                    },
                    y filter/.code={
                        \pgfmathparse{#1/1149*100}
                        \let\pgfmathresult=\pgfmathresult
                    }
                ] table [y index=0] {data/ours-imagewise_crs.txt};
                \addplot[
                    sharp plot,
                    thick,
                    no markers,
                    draw=red,
                    % forget plot,
                ] coordinates {(569.6071928176078,0) (569.6071928176078,100)};
                % \addlegendentry{Mean CR};
                % standard deviation
                % \addplot[
                %     sharp plot,
                %     thick,
                %     no markers,
                %     draw=red,
                %     dashed,
                %     % forget plot,
                % ] coordinates {(1846.416394135,0) (1846.416394135,11)};
                % images
                \node[inner sep=0pt, outer sep=0pt] (img1) at (axis cs:135,16.5) {\includegraphics[width=.236\linewidth,height=.236\linewidth]{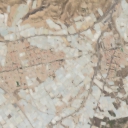}};
                \draw[arrow,dashed] (img1.south) -- (axis cs:263.40,5);
                \node[inner sep=0pt, outer sep=0pt] (img2) at (axis cs:1920,16.5) {\includegraphics[width=.236\linewidth,height=.236\linewidth]{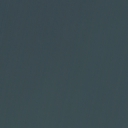}};
                \draw[arrow,dashed] (img2.south) -- (axis cs:3875.37,0.85);
            \end{axis}
        \end{tikzpicture}
        \caption{}
        \label{fig:hist-crs}
    \end{subfigure}
    \caption{Distribution of ({\subref{fig:hist-sas}}) pixelwise {\acp{sa}}, (\subref{fig:hist-psnrs}) imagewise \acp{psnr}, and (\subref{fig:hist-crs}) imagewise \acp{cr} across the HySpecNet-11k \cite{fuchs2023hyspecnet} test set (easy split) for \ac{ours} with $\lambda = \num{1e-4}$. The mean values are indicated in red along spectra and images with the lowest and highest metric values. Contains modified EnMAP data ©DLR [2022].}
    \label{fig:hist}
\end{figure}

In order to emphasize the strengths and limitations of the proposed model in terms of reconstruction fidelity and compression efficiency, the distributions of pixelwise {\acp{sa}} as well as imagewise \acp{psnr} and \acp{cr} across the HySpecNet-11k test set (easy split) are visualized in \autoref{fig:hist}.
For this experiment, the Lagrange multiplier $\lambda$ is set to \num{1e-4}, as the corresponding operating point on the \ac{rd} curve provides a balanced trade-off between reconstruction quality and compression efficiency, enabling a representative assessment of the model's overall behavior.
The distribution of the pixelwise {\acp{sa}} is illustrated in ({\subref{fig:hist-sas}}).
As shown in the figure, the vast majority of pixels are reconstructed with an $\ac{sa} < \SI{5}{\degree}$, indicating high spectral reconstruction fidelity \cite{licciardi2018spectral}.
The distribution is concentrated around {\SI{0.6}{\degree}}, whereas the the mean pixelwise {\ac{sa}} is {\SI{2.26}{\degree}}.
This discrepancy can be attributed to the approximately {\SI{12}{\percent}} of the samples that exhibit $\ac{sa} \geq \SI{5}{\degree}$.
As can be seen in the illustrated exemplary spectral signature, these outliers are predominantly associated with noisy spectral bands of water pixels that result in sharp spikes in the original spectra, which are smoothed in the reconstructions.
This behavior indicates an inherent denoising effect of the proposed model, which makes it noise-robust.
At the same time, it highlights a limitation of the {\ac{sa}} metric, which is particularly sensitive to noisy spectral bands and can therefore overestimate reconstruction errors.
\hl{Employing a noise-aware loss function (such as band-weighted {\ac{mse}}) or including the {\ac{sa}} into the training objective could enable the model to better preserve the spectral characteristics of noisy bands.
However, this may also increase the risk of overfitting to the noise, potentially degrading the overall reconstruction fidelity.}

The distribution of the imagewise \acp{psnr} in \autoref{fig:hist} (\subref{fig:hist-psnrs}) is concentrated around the mean of \SI{50.31}{\decibel} with a standard deviation of \SI{2.08}{\decibel}, closely following a Gaussian distribution.
This result underscores the robust generalization capabilities of the proposed model across diverse land cover types, enabling consistent high-fidelity reconstructions.
As shown in the figure, the \ac{hsi} with the lowest \ac{psnr} of \SI{43.41}{\decibel} predominantly contains clouds and exhibits high spectral and spatial variations due to the presence of multiple land cover types.
This complex scene poses significant challenges for the proposed model, as clouds are also underrepresented in the dataset.
In contrast to that, the highest \ac{psnr} of \SI{60.29}{\decibel} is obtained for a homogeneous scene containing only water, which exhibits low spectral and spatial variations, making it considerably less challenging to compress and accurately reconstruct.

The distribution of the imagewise \acp{cr} in \autoref{fig:hist} (\subref{fig:hist-crs}) indicates that most of the \acp{hsi} can be compressed with a \ac{cr} between \num{256} and \num{1024}.
However, a significant amount of \ac{hsi} achieves much higher \acp{cr}.
In particular, homogeneous \acp{hsi} are compressible to the greatest extent, as illustrated by the water scene achieving the highest \ac{cr} of \num{3875.37}.
In contrast, \acp{hsi} containing complex texture, multiple land cover types, or underrepresented classes are associated with the lowest \acp{cr}, as shown by the example achieving only a \ac{cr} of \num{265.40}.
Overall, our results highlight a bias towards homogeneous \ac{hsi}, indicating that learning from complex scenes to enhance both compression efficiency and reconstruction fidelity of these \acp{hsi} should be considered in future works.
Similar behavior of the distribution of {\acp{sa}}, \acp{psnr} and \acp{cr} has been observed on the MLRetSet dataset (not reported for space constraints).

\subsection{Downstream Task Experiment}
To assess the impact of lossy {\ac{hsi}} compression on downstream task performance, we employ \ac{lulc} segmentation as evaluation task.
Five randomly selected HySpecNet-11k {\acp{hsi}} are manually annotated by an independent expert, using multiple reference maps, and assigning one of the {\num{11}} \ac{gt} {\ac{lulc}} classes from the ESA WorldCover product \cite{zanaga2022esa} to each pixel.
For segmentation, we employ a basic {\ac{2d}} {\ac{cnn}}.
For each {\ac{hsi}}, the model is trained on the original data using a randomly selected subset of {\SI{20}{\percent}} of the annotated pixels as the training set.
Training is performed for {\num{750}} epochs using the Adam optimizer \cite{kingma2014adam}, with a learning rate of {\num{1e-3}} and the cross-entropy loss function.
Evaluation is conducted on the remaining {\SI{80}{\percent}} of pixels for each training run, comparing segmentation performance on both the original data and reconstructions of multiple {\ac{hsi}} compression methods.

\begin{table*}
    \centering
    \caption{Segmentation results for five different {\acp{hsi}} on the original data and reconstructions of state-of-the-art {\ac{hsi}} compression models at $\ac{cr} \approx \num{32}$. Segmentation performance is reported as weighted-average $F_1$-score and {\ac{miou}}, presented as the mean $\pm$ standard deviation over {\num{10}} independent training runs. The best metric value for each {\ac{hsi}} is highlighted in {\textbf{bold}}, and the second best in {\underline{underlined}}. Contains modified EnMAP data ©DLR [2022].}
    \label{tab:downstream-task}
    % \scriptsize
    \begin{tabular}{rcccccc}
        \hline
        \multicolumn{7}{c}{%
            \centering
            \begin{tikzpicture}
                \begin{axis}[
                    hide axis,
                    legend columns=6,
                    legend cell align={left},
                    legend style={
                        draw=none,
                        column sep=0.9em,
                        % font=\footnotesize,
                        % /tikz/every even column/.append style={column sep=0.5cm}
                    },
                    xmin=0, xmax=1,
                    ymin=-0.5, ymax=0.5,
                ]
                
                % ---- class 0 ----
                \addlegendimage{mark=square*, mark size=2pt, only marks,
                    fill={rgb,255:red,0;green,0;blue,0},
                    draw={rgb,255:red,0;green,0;blue,0}}
                \addlegendentry{None}
                
                % ---- 10 ----
                \addlegendimage{mark=square*, mark size=2pt, only marks,
                    fill={rgb,255:red,0;green,100;blue,0},
                    draw={rgb,255:red,0;green,100;blue,0}}
                \addlegendentry{Tree cover}
                
                % ---- 20 ----
                \addlegendimage{mark=square*, mark size=2pt, only marks,
                    fill={rgb,255:red,255;green,187;blue,34},
                    draw={rgb,255:red,255;green,187;blue,34}}
                \addlegendentry{Shrubland}
                
                % ---- 30 ----
                \addlegendimage{mark=square*, mark size=2pt, only marks,
                    fill={rgb,255:red,255;green,255;blue,76},
                    draw={rgb,255:red,255;green,255;blue,76}}
                \addlegendentry{Grassland}
                
                % ---- 40 ----
                \addlegendimage{mark=square*, mark size=2pt, only marks,
                    fill={rgb,255:red,240;green,150;blue,255},
                    draw={rgb,255:red,240;green,150;blue,255}}
                \addlegendentry{Cropland}
                
                % ---- 50 ----
                \addlegendimage{mark=square*, mark size=2pt, only marks,
                    fill={rgb,255:red,250;green,0;blue,0},
                    draw={rgb,255:red,250;green,0;blue,0}}
                \addlegendentry{Built-up}
                
                % ---- 60 ----
                \addlegendimage{mark=square*, mark size=2pt, only marks,
                    fill={rgb,255:red,180;green,180;blue,180},
                    draw={rgb,255:red,180;green,180;blue,180}}
                \addlegendentry{Bare / sparse vegetation}
                
                % ---- 70 ----
                \addlegendimage{mark=square*, mark size=2pt, only marks,
                    fill={rgb,255:red,240;green,240;blue,240},
                    draw={rgb,255:red,240;green,240;blue,240}}
                \addlegendentry{Snow and ice}
                
                % ---- 80 ----
                \addlegendimage{mark=square*, mark size=2pt, only marks,
                    fill={rgb,255:red,0;green,100;blue,200},
                    draw={rgb,255:red,0;green,100;blue,200}}
                \addlegendentry{Permanent water bodies}
                
                % ---- 90 ----
                \addlegendimage{mark=square*, mark size=2pt, only marks,
                    fill={rgb,255:red,0;green,150;blue,160},
                    draw={rgb,255:red,0;green,150;blue,160}}
                \addlegendentry{Herbaceous wetland}
                
                % ---- 95 ----
                \addlegendimage{mark=square*, mark size=2pt, only marks,
                    fill={rgb,255:red,0;green,207;blue,117},
                    draw={rgb,255:red,0;green,207;blue,117}}
                \addlegendentry{Mangroves}
                
                % ---- 100 ----
                \addlegendimage{mark=square*, mark size=2pt, only marks,
                    fill={rgb,255:red,250;green,230;blue,160},
                    draw={rgb,255:red,250;green,230;blue,160}}
                \addlegendentry{Moss and lichen}

                \addplot[draw=none] coordinates {(0,0) (1,1)};
                \end{axis}
            \end{tikzpicture}
        } \\[-6.25em]
        \hline
        \multicolumn{1}{c}{\acs{gt}} & \multicolumn{1}{c}{Original \ac{hsi}} & \multicolumn{1}{c}{\ac{hycot} \cite{fuchs2024hycot}} & \multicolumn{1}{c}{\ac{sscnet} \cite{la2022hyperspectral}} & \multicolumn{1}{c}{\ac{3dcae} \cite{chong2021end}} & \multicolumn{1}{c}{\ac{hycass} \cite{fuchs2025adjustable}} & \multicolumn{1}{c}{\ac{ours}} \\
        \hline
        \\[-1.5mm]
        \includegraphics[width=.118\linewidth,height=.118\linewidth]{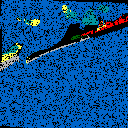} &
        \includegraphics[width=.118\linewidth,height=.118\linewidth]{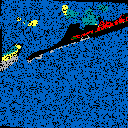} &
        \includegraphics[width=.118\linewidth,height=.118\linewidth]{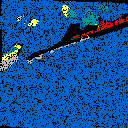} &
        \includegraphics[width=.118\linewidth,height=.118\linewidth]{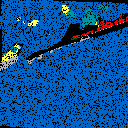} &
        \includegraphics[width=.118\linewidth,height=.118\linewidth]{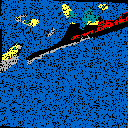} &
        \includegraphics[width=.118\linewidth,height=.118\linewidth]{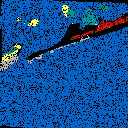} &
        \includegraphics[width=.118\linewidth,height=.118\linewidth]{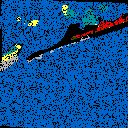}
        \\
        \cmidrule(lr){1-1}
        \cmidrule(lr){2-2}
        \cmidrule(lr){3-3}
        \cmidrule(lr){4-4}
        \cmidrule(lr){5-5}
        \cmidrule(lr){6-6}
        \cmidrule(lr){7-7}
        % \ac{cr} / \ac{psnr} & \num{1.00} / $\infty$ \si{\decibel} & \num{28.86} / \SI{49.70}{\decibel} & \num{32.00} / \SI{42.48}{\decibel} & \num{31.69}/ \SI{37.83}{\decibel}  & \num{28.86} / \SI{49.53}{\decibel} & \num{39.69} / \SI{52.31}{\decibel} \\
        $F_1$ [\si{\percent}] $\uparrow$ & \textbf{\num[round-mode=places,round-precision=2]{97.5921264323911} $\pm$ \num[round-mode=places,round-precision=2]{0.160538237218986}} & \num[round-mode=places,round-precision=2]{97.4267531455534} $\pm$ \num[round-mode=places,round-precision=2]{0.150094811598312} & \num[round-mode=places,round-precision=2]{97.1633941820577} $\pm$ \num[round-mode=places,round-precision=2]{0.177049980780286} & \num[round-mode=places,round-precision=2]{96.1861487525981} $\pm$ \num[round-mode=places,round-precision=2]{0.236617897172884} & \num[round-mode=places,round-precision=2]{97.4568915174118} $\pm$ \num[round-mode=places,round-precision=2]{0.130096846980387} & \underline{\num[round-mode=places,round-precision=2]{97.5516736772868} $\pm$ \num[round-mode=places,round-precision=2]{0.179453609187942}}\\
        \acs{miou} [\si{\percent}] $\uparrow$ & \textbf{\num[round-mode=places,round-precision=2]{58.1369727767281} $\pm$ \num[round-mode=places,round-precision=2]{3.02236166256106}} & \num[round-mode=places,round-precision=2]{56.503351858476} $\pm$ \num[round-mode=places,round-precision=2]{2.84032654871458} & \num[round-mode=places,round-precision=2]{54.3670889012237} $\pm$ \num[round-mode=places,round-precision=2]{2.51589026174033} & \num[round-mode=places,round-precision=2]{48.3357150489751} $\pm$ \num[round-mode=places,round-precision=2]{2.98129404765829} & \num[round-mode=places,round-precision=2]{56.6756345364446} $\pm$ \num[round-mode=places,round-precision=2]{3.02291494698213} & \underline{\num[round-mode=places,round-precision=2]{57.8399857582885} $\pm$ \num[round-mode=places,round-precision=2]{3.0810253036626}} \\[0.3em]
        \hline
        \\[-1.5mm]
        \includegraphics[width=.118\linewidth,height=.118\linewidth]{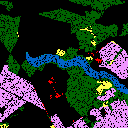} &
        \includegraphics[width=.118\linewidth,height=.118\linewidth]{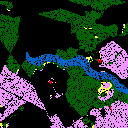} &
        \includegraphics[width=.118\linewidth,height=.118\linewidth]{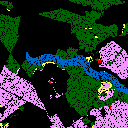} &
        \includegraphics[width=.118\linewidth,height=.118\linewidth]{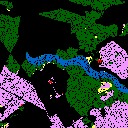} &
        \includegraphics[width=.118\linewidth,height=.118\linewidth]{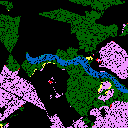} &
        \includegraphics[width=.118\linewidth,height=.118\linewidth]{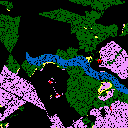} &
        \includegraphics[width=.118\linewidth,height=.118\linewidth]{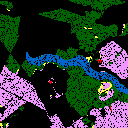}
        \\
        \cmidrule(lr){1-1}
        \cmidrule(lr){2-2}
        \cmidrule(lr){3-3}
        \cmidrule(lr){4-4}
        \cmidrule(lr){5-5}
        \cmidrule(lr){6-6}
        \cmidrule(lr){7-7}
        $F_1$ [\si{\percent}] $\uparrow$ & \textbf{\num[round-mode=places,round-precision=2]{92.3142850498166} $\pm$ \num[round-mode=places,round-precision=2]{0.592473534173584}} & \num[round-mode=places,round-precision=2]{91.7933308244766} $\pm$ \num[round-mode=places,round-precision=2]{0.544236552184775} & \num[round-mode=places,round-precision=2]{91.0455472360793} $\pm$ \num[round-mode=places,round-precision=2]{0.802342750429158} & \num[round-mode=places,round-precision=2]{88.0951252962768} $\pm$ \num[round-mode=places,round-precision=2]{2.75797220695145} & \num[round-mode=places,round-precision=2]{91.8502482813682} $\pm$ \num[round-mode=places,round-precision=2]{0.542705311957364} & \underline{\num[round-mode=places,round-precision=2]{92.1637797553351} $\pm$ \num[round-mode=places,round-precision=2]{0.548193956202256}} \\
        \acs{miou} [\si{\percent}] $\uparrow$ & \textbf{\num[round-mode=places,round-precision=2]{56.6370588941377} $\pm$ \num[round-mode=places,round-precision=2]{3.4266526241031}} & \num[round-mode=places,round-precision=2]{55.5386912418067} $\pm$ \num[round-mode=places,round-precision=2]{3.10137530045416} & \num[round-mode=places,round-precision=2]{54.1309417927139} $\pm$ \num[round-mode=places,round-precision=2]{4.25227093187108} & \num[round-mode=places,round-precision=2]{47.9045144673979} $\pm$ \num[round-mode=places,round-precision=2]{5.80241730911286} & \num[round-mode=places,round-precision=2]{55.9930111303549} $\pm$ \num[round-mode=places,round-precision=2]{3.36492177953391} & \underline{\num[round-mode=places,round-precision=2]{56.2923302955396} $\pm$ \num[round-mode=places,round-precision=2]{3.37462244678884}} \\[0.3em]
        \hline
        \\[-1.5mm]
        \includegraphics[width=.118\linewidth,height=.118\linewidth]{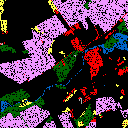} &
        \includegraphics[width=.118\linewidth,height=.118\linewidth]{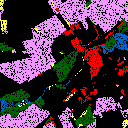} &
        \includegraphics[width=.118\linewidth,height=.118\linewidth]{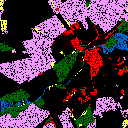} &
        \includegraphics[width=.118\linewidth,height=.118\linewidth]{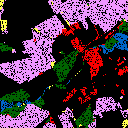} &
        \includegraphics[width=.118\linewidth,height=.118\linewidth]{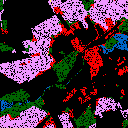} &
        \includegraphics[width=.118\linewidth,height=.118\linewidth]{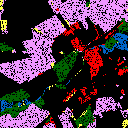} &
        \includegraphics[width=.118\linewidth,height=.118\linewidth]{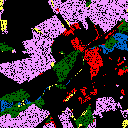}
        \\
        \cmidrule(lr){1-1}
        \cmidrule(lr){2-2}
        \cmidrule(lr){3-3}
        \cmidrule(lr){4-4}
        \cmidrule(lr){5-5}
        \cmidrule(lr){6-6}
        \cmidrule(lr){7-7}
        % \ac{cr} / \ac{psnr} & \num{1.00} / $\infty$ \si{\decibel} & \num{28.86} / \SI{49.70}{\decibel} & \num{32.00} / \SI{42.48}{\decibel} & \num{31.69}/ \SI{37.83}{\decibel}  & \num{28.86} / \SI{49.53}{\decibel} & \num{39.69} / \SI{52.31}{\decibel} \\
        $F_1$ [\si{\percent}] $\uparrow$ & \textbf{\num[round-mode=places,round-precision=2]{89.8948561296184} $\pm$ \num[round-mode=places,round-precision=2]{0.309589804551928}} & \num[round-mode=places,round-precision=2]{89.0913786842273} $\pm$ \num[round-mode=places,round-precision=2]{0.32301486122257} & \num[round-mode=places,round-precision=2]{87.8612330809135} $\pm$ \num[round-mode=places,round-precision=2]{0.305097032774303} & \num[round-mode=places,round-precision=2]{79.9718550712005} $\pm$ \num[round-mode=places,round-precision=2]{4.25166230526363} & \num[round-mode=places,round-precision=2]{89.0394302225927} $\pm$ \num[round-mode=places,round-precision=2]{0.333234018595453} & \underline{\num[round-mode=places,round-precision=2]{89.7501828337917} $\pm$ \num[round-mode=places,round-precision=2]{0.290825012808799}} \\
        \acs{miou} [\si{\percent}] $\uparrow$ & \textbf{\num[round-mode=places,round-precision=2]{59.772155768488} $\pm$ \num[round-mode=places,round-precision=2]{0.970430784651257}} & \num[round-mode=places,round-precision=2]{58.1128392611846} $\pm$ \num[round-mode=places,round-precision=2]{0.907750775554861} & \num[round-mode=places,round-precision=2]{56.1877226430996} $\pm$ \num[round-mode=places,round-precision=2]{0.780429223992965} & \num[round-mode=places,round-precision=2]{46.0273470326259} $\pm$ \num[round-mode=places,round-precision=2]{3.65514227816818} & \num[round-mode=places,round-precision=2]{58.0301024287015} $\pm$ \num[round-mode=places,round-precision=2]{0.878879981857698} & \underline{\num[round-mode=places,round-precision=2]{59.5276958982309} $\pm$ \num[round-mode=places,round-precision=2]{0.902740544270734}} \\[0.3em]
        \hline
        \\[-1.5mm]
        \includegraphics[width=.118\linewidth,height=.118\linewidth]{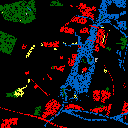} &
        \includegraphics[width=.118\linewidth,height=.118\linewidth]{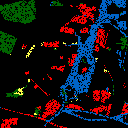} &
        \includegraphics[width=.118\linewidth,height=.118\linewidth]{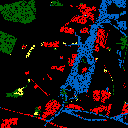} &
        \includegraphics[width=.118\linewidth,height=.118\linewidth]{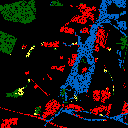} &
        \includegraphics[width=.118\linewidth,height=.118\linewidth]{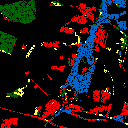} &
        \includegraphics[width=.118\linewidth,height=.118\linewidth]{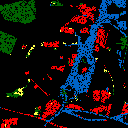} &
        \includegraphics[width=.118\linewidth,height=.118\linewidth]{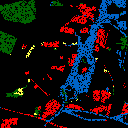}
        \\
        \cmidrule(lr){1-1}
        \cmidrule(lr){2-2}
        \cmidrule(lr){3-3}
        \cmidrule(lr){4-4}
        \cmidrule(lr){5-5}
        \cmidrule(lr){6-6}
        \cmidrule(lr){7-7}
        % \ac{cr} / \ac{psnr} & \num{1.00} / $\infty$ \si{\decibel} & \num{28.86} / \SI{49.70}{\decibel} & \num{32.00} / \SI{42.48}{\decibel} & \num{31.69}/ \SI{37.83}{\decibel}  & \num{28.86} / \SI{49.53}{\decibel} & \num{39.69} / \SI{52.31}{\decibel} \\
        $F_1$ [\si{\percent}] $\uparrow$ & \textbf{\num[round-mode=places,round-precision=2]{92.4179443938328} $\pm$ \num[round-mode=places,round-precision=2]{0.471618225595706}} & \num[round-mode=places,round-precision=2]{91.1530775909873} $\pm$ \num[round-mode=places,round-precision=2]{0.68325994685687} & \num[round-mode=places,round-precision=2]{89.9844385333687} $\pm$ \num[round-mode=places,round-precision=2]{0.815123327823397} & \num[round-mode=places,round-precision=2]{84.8974966026153} $\pm$ \num[round-mode=places,round-precision=2]{1.89438620907353} & \num[round-mode=places,round-precision=2]{90.4448528728254} $\pm$ \num[round-mode=places,round-precision=2]{0.756758924776952} & \underline{\num[round-mode=places,round-precision=2]{91.9590716902637} $\pm$ \num[round-mode=places,round-precision=2]{0.487732344258347}} \\
        \acs{miou} [\si{\percent}] $\uparrow$ & \textbf{\num[round-mode=places,round-precision=2]{69.426941333473} $\pm$ \num[round-mode=places,round-precision=2]{4.25231622355435}} & \num[round-mode=places,round-precision=2]{64.9425976307172} $\pm$ \num[round-mode=places,round-precision=2]{4.29431943970165} & \num[round-mode=places,round-precision=2]{61.4923615153657} $\pm$ \num[round-mode=places,round-precision=2]{3.36767988276387} & \num[round-mode=places,round-precision=2]{57.8983496108752} $\pm$ \num[round-mode=places,round-precision=2]{3.67528512942404} & \num[round-mode=places,round-precision=2]{63.3611327187803} $\pm$ \num[round-mode=places,round-precision=2]{4.42673856176818} & \underline{\num[round-mode=places,round-precision=2]{67.5418464617711} $\pm$ \num[round-mode=places,round-precision=2]{4.04840752258704}} \\[0.3em]
        \hline
        \\[-1.5mm]
        \includegraphics[width=.118\linewidth,height=.118\linewidth]{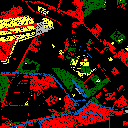} &
        \includegraphics[width=.118\linewidth,height=.118\linewidth]{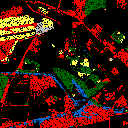} &
        \includegraphics[width=.118\linewidth,height=.118\linewidth]{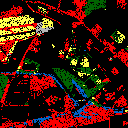} &
        \includegraphics[width=.118\linewidth,height=.118\linewidth]{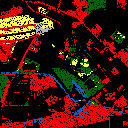} &
        \includegraphics[width=.118\linewidth,height=.118\linewidth]{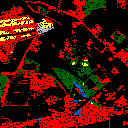} &
        \includegraphics[width=.118\linewidth,height=.118\linewidth]{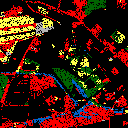} &
        \includegraphics[width=.118\linewidth,height=.118\linewidth]{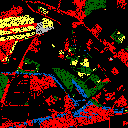}
        \\
        \cmidrule(lr){1-1}
        \cmidrule(lr){2-2}
        \cmidrule(lr){3-3}
        \cmidrule(lr){4-4}
        \cmidrule(lr){5-5}
        \cmidrule(lr){6-6}
        \cmidrule(lr){7-7}
        % \ac{cr} / \ac{psnr} & \num{1.00} / $\infty$ \si{\decibel} & \num{28.86} / \SI{49.70}{\decibel} & \num{32.00} / \SI{42.48}{\decibel} & \num{31.69}/ \SI{37.83}{\decibel}  & \num{28.86} / \SI{49.53}{\decibel} & \num{39.69} / \SI{52.31}{\decibel} \\
        $F_1$ [\si{\percent}] $\uparrow$ & \textbf{\num[round-mode=places,round-precision=2]{85.218047587945} $\pm$ \num[round-mode=places,round-precision=2]{0.46925134821269}} & \num[round-mode=places,round-precision=2]{83.4865431601047} $\pm$ \num[round-mode=places,round-precision=2]{0.456127883861799} & \num[round-mode=places,round-precision=2]{83.2183221877812} $\pm$ \num[round-mode=places,round-precision=2]{0.507002587949912} & \num[round-mode=places,round-precision=2]{75.7758012522783} $\pm$ \num[round-mode=places,round-precision=2]{3.26009496713617} & \num[round-mode=places,round-precision=2]{81.353613635631} $\pm$ \num[round-mode=places,round-precision=2]{1.15759304883577} & \underline{\num[round-mode=places,round-precision=2]{84.8491768725084} $\pm$ \num[round-mode=places,round-precision=2]{0.360602270175662}} \\
        \acs{miou} [\si{\percent}] $\uparrow$ & \textbf{\num[round-mode=places,round-precision=2]{68.8297687822791} $\pm$ \num[round-mode=places,round-precision=2]{0.734211421000938}} & \num[round-mode=places,round-precision=2]{66.1433027841048} $\pm$ \num[round-mode=places,round-precision=2]{0.906520641276125} & \num[round-mode=places,round-precision=2]{65.9423120513906} $\pm$ \num[round-mode=places,round-precision=2]{0.895444410176155} & \num[round-mode=places,round-precision=2]{53.7850603259249} $\pm$ \num[round-mode=places,round-precision=2]{5.47995046571383} & \num[round-mode=places,round-precision=2]{61.7465012794822} $\pm$ \num[round-mode=places,round-precision=2]{1.58893010362021} & \underline{\num[round-mode=places,round-precision=2]{68.2858613266997} $\pm$ \num[round-mode=places,round-precision=2]{0.733992705178399}} \\[0.3em]
        \hline
    \end{tabular}
\end{table*}

Segmentation results are qualitatively and quantitatively reported in \autoref{tab:downstream-task} in terms of segmentation maps, weighted-average $F_1$-scores across all {\ac{lulc}} classes and \acp{miou}, where both quantitative metrics are reported as mean and standard deviation over {\num{10}} independent training runs for each {\ac{hsi}}.
The results indicate that the reconstructions of {\ac{ours}} lead to the best segmentation performance compared to other compression models.
In comparison to the original {\ac{hsi}}, {\ac{ours}} does not significantly degrade segmentation performance.
Overall, these results highlight the robustness of {\ac{ours}}'s reconstructions through the preservation of task-relevant spatial and spectral information.
We would like to note that segmentation can also be applied on much higher compressed data without significant loss of performance (not reported for space constraints).

\color{black}
\section{Conclusion}
\label{sec:conclusion}
In this paper, we have explored the effects of spatio-spectral feature learning on the {\ac{rd}} performance of variational {\ac{hsi}} compression as a first time in {\ac{rs}}.
To this end, we have proposed to use simple but efficient configurable spatial and spectral feature learning blocks for the {\ac{vae}}-based compression of {\acp{hsi}}.
To achieve this, we have presented \ac{ours}, a configurable spatio-spectral {\ac{vae}} for \hl{gound-based} \ac{hsi} compression.
The proposed model builds upon the mean \& scale hyperprior \cite{minnen2018joint} by introducing four main components:
\begin{enumerate*}[1)]
    \item a configurable spatio-spectral encoder;
    \item a spatio-spectral hyperencoder;
    \item a spatio-spectral hyperdecoder; and
    \item a configurable spatio-spectral decoder.
\end{enumerate*}
Unlike existing methods that naively adapt variational image compression models from the \ac{cv} domain without explicitly modeling the spatio-spectral redundancies inherent in \acp{hsi}, \ac{ours} provides flexible control over balancing spatial and spectral feature learning to effectively leverage the unique characteristics of hyperspectral data.
In the experiments, we have evaluated the proposed model on two benchmark datasets in order to conduct:
\begin{enumerate*}[i)]
    \item ablation studies including a metric-driven selection of the proposed model's hyperparameters;
    \item comparisons with state-of-the-art methods; and
    \item analyses of the reconstruction results.
\end{enumerate*}
Our experimental results validate the effectiveness of \ac{ours} across a wide range of \acp{cr} and demonstrate the necessity to balance spatio-spectral feature learning to achieve high-fidelity spatial and spectral reconstructions.
Our findings suggest that, compared to natural image compression, variational \ac{hsi} compression benefits from a reduced spatial receptive field, reduced spatial downsampling and increased transform capacity to more effectively preserve the spatial and spectral information of \acp{hsi}.
Furthermore, we observe that compression efficiency is significantly reduced for \acp{hsi} containing noisy bands, complex textures, multiple land cover types, or underrepresented classes, whereas it remains highly effective for homogeneous \acp{hsi} dominated by water.
This highlights the need to account for such biases during model training to ensure a more balanced and robust \ac{hsi} compression across diverse scenes.

\hl{As a final remark, we would like to note that we fixed the hyperlatent channels as suggested in the literature.
As a future work, we plan to investigate the effect of varying the ratio between the latent and hyperlatent channels.}
Moreover, we also plan to explore entropy models that may more effectively capture the statistical dependencies within the latent representations of \acp{hsi}.
% Moreover, we plan to develop an interactive \ac{llm} agent framework that guides the \ac{hsi} compression process based on user input and contextual information.
%\ac{llm} agents can perform tasks such as data interpretation and analysis by reasoning across information, creating a plan to solve a problem, and executing the plan with the help of a set of tools.
% We believe that leveraging the reasoning capabilities of \ac{llm} agents can serve as a versatile, user-driven, and context-aware framework for \ac{hsi} compression.
%Hence, as a future work, we plan to develop an interactive \ac{llm} agent framework that guides the \ac{hsi} compression process based on user input and contextual information.

\section*{Acknowledgement}
The authors would like to thank Jan Hagen Philipps for providing the {\ac{lulc}} {\ac{gt}} and assisting with the downstream task experiment.

\bibliographystyle{IEEEtran}
\bibliography{bib/refs.bib}

\end{document}